\documentclass{article}

\PassOptionsToPackage{numbers, compress}{natbib}

\usepackage[final]{neurips_2022}

\usepackage[utf8]{inputenc} 
\usepackage[T1]{fontenc}    
\usepackage{hyperref}       
\usepackage{url}            
\usepackage{booktabs}       
\usepackage{amsfonts}       
\usepackage{nicefrac}       
\usepackage{microtype}
\usepackage{graphicx}
\usepackage{booktabs} 
\usepackage{makecell}
\usepackage{caption}
\usepackage{courier}
\usepackage{wrapfig}
\usepackage{bm}
\usepackage{multirow}
\usepackage{color}
\usepackage{subcaption}
\usepackage{amssymb}
\usepackage{amsmath}
\usepackage{gensymb}
\usepackage{amsfonts}       
\usepackage{enumitem}
\usepackage{multirow}
\usepackage{amsmath,nccmath}
\usepackage{amsthm}
\usepackage[dvipsnames]{xcolor}
\usepackage{hyperref}
\usepackage{algorithm}
\usepackage[noend]{algorithmic}

\newtheorem{theorem}{Theorem}
\newtheorem{lemma_theorem}{Theorem}

\newcommand{\E}{\mathbb{E}}
\newcommand{\R}{\mathbb{R}}

\newtheorem{lemma}{Lemma}[]
\renewcommand\footnotemark{}

\title{C-Mixup: Improving Generalization in Regression}
\author{Huaxiu Yao$^{1*}$\thanks{$^*$Equal contribution. This work was done when Yiping Wang was remotely co-mentored by Huaxiu Yao and Linjun Zhang.}, Yiping Wang$^{2*}$, Linjun Zhang$^3$, James Zou$^1$, Chelsea Finn$^1$\\$^1$Stanford University, $^2$Zhejiang University, $^3$Rutgers University\\$^1$\{huaxiu,cbfinn\}@cs.stanford.edu, jamesz@stanford.edu\\ $^2$yipingwang6161@gmail.com, $^3$linjun.zhang@rutgers.edu}

\begin{document}

\maketitle

\begin{abstract}
Improving the generalization of deep networks is an important open challenge, particularly in domains without plentiful data. The mixup algorithm improves generalization by linearly interpolating a pair of examples and their corresponding labels. These interpolated examples augment the original training set. Mixup has shown promising results in various classification tasks, but systematic analysis of mixup in regression remains underexplored. Using mixup directly on regression labels can result in arbitrarily incorrect labels.
In this paper, we propose a simple yet powerful algorithm, C-Mixup, to improve generalization on regression tasks. In contrast with vanilla mixup, which picks training examples for mixing with uniform probability, C-Mixup adjusts the sampling probability based on the similarity of the labels. Our theoretical analysis confirms that C-Mixup with label similarity obtains a smaller mean square error in supervised regression and meta-regression than vanilla mixup and using feature similarity. Another benefit of C-Mixup is that it can improve out-of-distribution robustness, where the test distribution is different from the training distribution. By selectively interpolating examples with similar labels, it mitigates the effects of domain-associated information and yields domain-invariant representations. We evaluate C-Mixup on eleven datasets, ranging from tabular to video data. Compared to the best prior approach, C-Mixup achieves 6.56\%, 4.76\%, 5.82\% improvements in in-distribution generalization, task generalization, and out-of-distribution robustness, respectively. Code is released at \href{https://github.com/huaxiuyao/C-Mixup}{https://github.com/huaxiuyao/C-Mixup}.
\end{abstract}
\section{Introduction}
\label{sec:intro}
Deep learning practitioners commonly face the challenge of overfitting. To improve generalization, prior works have proposed a number of techniques, including data augmentation~\cite{antoniou2017data,cubuk2019autoaugment,devries2017improved,yun2019cutmix,zhang2017mixup} and explicit regularization~\cite{gal2016dropout,krogh1992simple,srivastava2014dropout}. Representatively, mixup~\cite{zhang2017mixup,zhang2020does} densifies the data distribution and implicitly regularizes the model by linearly interpolating the features of randomly sampled pairs of examples and applying the same interpolation on the corresponding labels. Despite mixup having demonstrated promising results in improving generalization in classification problems, it has rarely been studied in the context of regression with continuous labels, on which we focus in this paper.

\begin{figure}[t]
\centering
\includegraphics[width=\textwidth]{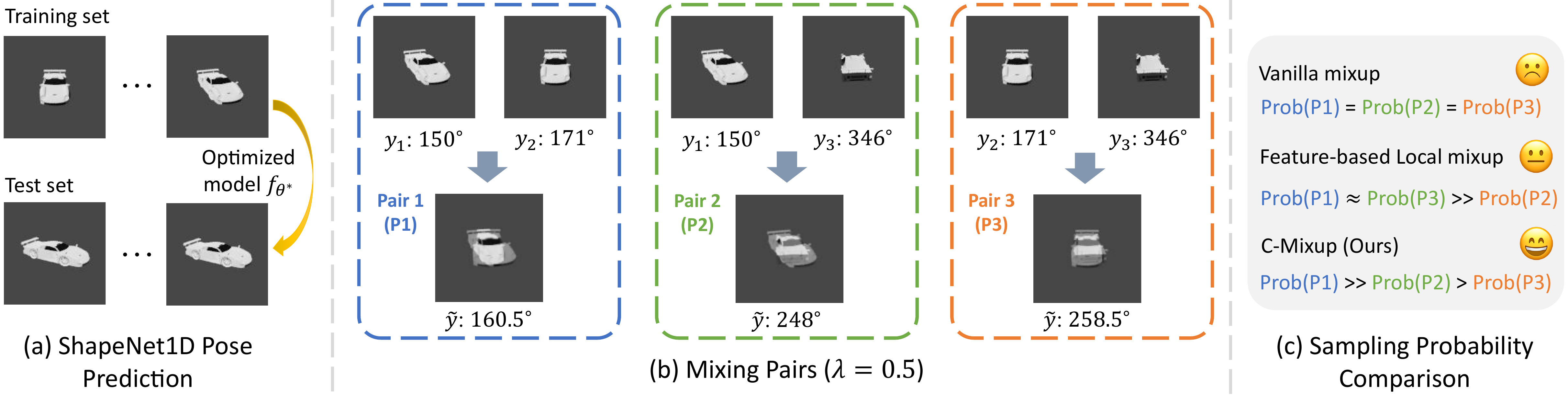}
\caption{Illustration of C-Mixup on ShapeNet1D pose prediction. $\lambda$ represents the interpolation ratio. (a) ShapeNet1D pose prediction task, aiming to predict the current orientation of the object relative to its canonical orientation. (b) Three mixing pairs are randomly picked, where the interpolated images are visualized and $\Tilde{y}$ represents interpolated labels. (c) Illustration of a rough comparison of sampling probabilities among three mixing pairs in (b). The Euclidean distance measures input feature distance and the corresponding results between examples in pairs 1, 2, 3 are \begin{small}$1.51\times 10^5$\end{small}, \begin{small}$1.82\times 10^5$\end{small}, \begin{small}$1.50\times 10^5$\end{small}, respectively. Hence, pairs 1 and 3 have similar results, leading to similar sampling probabilities. C-Mixup is able to assign higher sampling probability for more reasonable mixing pairs.}
\label{fig:iid_mixup}
\vspace{-1em}
\end{figure}

In contrast to classification, which formalizes the label as a one-hot vector, the goal of regression is to predict a continuous label
from each input. Directly applying mixup to input features and labels in regression tasks may yield arbitrarily incorrect labels. For example, as shown in Figure~\ref{fig:iid_mixup}(a), ShapeNet1D pose prediction~\cite{gao2022matters} aims to predict the current orientation of the object relative to its canonical orientation. We randomly select three mixing pairs and show the mixed images and labels in Figure~\ref{fig:iid_mixup}(b), where only pair 1 exhibits reasonable mixing results. We thus see that sampling mixing pairs uniformly from the dataset introduces a number of noisy pairs. 

In this paper, we aim to adjust the sampling probability of mixing pairs according to the similarity of examples, resulting in a simple training technique named \textbf{C-Mixup}. Specifically, we employ a Gaussian kernel to calculate the sampling probability of drawing another example for mixing, where closer examples are more likely to be sampled. Here, the core question is: \emph{how to measure the similarity between two examples?} The most straightforward solution is to compute input feature similarity. Yet, using input similarity has two major downsides when dealing with high-dimensional data such as images or time-series: substantial computational costs and lack of good distance metrics. Specifically, it takes considerable time to compute pairwise similarities across all samples, and directly applying classical distance metrics (e.g., Euclidean distance, cosine distance) does not reflect the high-level relation between input features. In the ShapeNet1D rotation prediction example (Figure~\ref{fig:iid_mixup}(a)), pair 1 and 3 have close input similarities, while only pair 1 can be reasonably interpolated.

To overcome these drawbacks, C-Mixup instead uses label similarity, which is typically much faster to compute since the label space is usually low dimensional. In addition to the computational advantages, C-Mixup benefits three kinds of regression problems. \emph{First}, it empirically improves in-distribution generalization in supervised regression compared to using vanilla mixup or using feature similarity. \emph{Second}, we extend C-Mixup to gradient-based meta-learning by incorporating it into MetaMix, a mixup-based task augmentation method~\cite{yao2021improving}. Compared to vanilla MetaMix, C-Mixup empirically improves task generalization. \emph{Third}, C-Mixup is well-suited for improving out-of-distribution robustness without domain information, particularly to covariate shift (see the corresponding example in Appendix~\ref{app:a_1}). 
By performing mixup on examples with close continuous labels, examples with different domains are mixed. In this way, C-Mixup encourages the model to rely on domain-invariant features to make prediction and ignore unrelated or spurious correlations, making the model more robust to covariate shift. 

The primary contribution of this paper is C-Mixup, a simple and scalable algorithm for improving generalization in regression problems. In linear or monotonic non-linear models, our theoretical analysis shows that C-Mixup improves generalization in multiple settings compared to vanilla mixup or compared to using feature similarities. Moreover, our experiments thoroughly evaluate C-Mixup on eleven datasets, including many large-scale real-world applications like drug-target interaction prediction~\cite{tdc}, ejection fraction estimation with echocardiogram videos~\cite{ouyang2020video}, poverty estimation with satellite imagery~\cite{yeh2020using}. Compared to the best prior method, the results demonstrate the promise of C-Mixup with 6.56\%, 4.76\%, 5.82\% improvements in in-distribution generalization, task generalization, and out-of-distribution robustness, respectively.


\section{Preliminaries}
\newcommand{\task}{\mathcal{T}}
\newcommand{\taski}{\task_i}
\newcommand{\taskt}{\task_t}
\newcommand{\testtask}{\task_t}
\newcommand{\data}{\mathcal{D}}
\newcommand{\datak}{\mathcal{D}_m}
\newcommand{\dataks}{\mathcal{D}_m^s}
\newcommand{\datakq}{\mathcal{D}_m^q}
\newcommand{\datats}{\mathcal{D}_t^s}
\newcommand{\datatq}{\mathcal{D}_t^q}
\newcommand{\featks}{\mathbf{X}_m^s}
\newcommand{\featkq}{\mathbf{X}_m^q}
\newcommand{\labelks}{y_m^s}
\newcommand{\labelkq}{y_m^q}

\newcommand{\datas}{\mathcal{D}^s}
\newcommand{\dataq}{\mathcal{D}^q}
\newcommand{\feat}{\mathbf{X}}
\newcommand{\feats}{\mathbf{X}^s}
\newcommand{\featq}{\mathbf{X}^q}
\newcommand{\labely}{y}
\newcommand{\labels}{y^s}
\newcommand{\labelq}{y^q}
\newcommand{\loss}{\mathcal{L}}

In this section, we define notation and describe the background of ERM and mixup in the supervised learning setting, and MetaMix in the meta-learning setting for task generalization.

\textbf{ERM.} Assume a machine learning model $f$ with parameter space $\Theta$. In this paper, we consider the setting where one predicts the continuous label $y \in \mathcal{Y}$ according to the input feature $x\in \mathcal{X}$. Given a loss function $\ell$, we train a model $f_{\theta}$ under the empirical training distribution $P^{tr}$ with the following objective, and get the optimized parameter $\theta^{*} \in \Theta$:
\begin{equation}
\small
\label{eq:erm}
\theta^{*} \leftarrow \arg\min_{\theta \in \Theta} \mathbb{E}_{(x,y)\sim P^{tr}} [\ell(f_{\theta}(x), y)].
\end{equation}
Typically, we expect the model to perform well on unseen examples drawn from the test distribution \begin{small}$P^{ts}$\end{small}. We are interested in both \emph{in-distribution} (\begin{small}$P^{tr}=P^{ts}$\end{small}) and \emph{out-of-distribution} (\begin{small}$P^{tr}\neq P^{ts}$\end{small}) settings. 

\textbf{Mixup.} The mixup algorithm samples a pair of instances $(x_i, y_i)$ and $(x_j, y_j)$, sampled uniformly at random from the training dataset, and generates new examples by performing linear interpolation on the input features and corresponding labels as:
\begin{equation}
\label{eq:interpolate}
\small
    \Tilde{x}=\lambda\cdot x_i + (1-\lambda)\cdot x_j,\; \Tilde{y}=\lambda\cdot y_i + (1-\lambda)\cdot y_j,
\end{equation}
where the interpolation ratio \begin{small}$\lambda \in [0,1]$\end{small} is drawn from a Beta distribution, i.e., \begin{small}$\lambda\sim \mathrm{Beta}(\alpha, \alpha)$\end{small}. The interpolated examples are then used to optimize the model as follows:
\begin{equation}
\small
\label{eq:mixup}
\theta^{*} \leftarrow \arg\min_{\theta \in \Theta} \mathbb{E}_{(x_i,y_i), (x_j, y_j)\sim P^{tr}} [\ell(f_{\theta}(\Tilde{x}), \Tilde{y})].
\end{equation}

\textbf{Task Generalization and MetaMix.} 
In this paper, we also investigate few-shot \emph{task generalization} under the gradient-based meta-regression setting. Given a task distribution \begin{small}$p(\task)$\end{small}, we assume each task \begin{small}$\task_m$\end{small} is sampled from \begin{small}$p(\task)$\end{small} and is associated with a dataset \begin{small}$\mathcal{D}_m$\end{small}. A support set \begin{small}$\dataks=\{(X_m^s, Y_m^s)\}=\{(x_{m,i}^s, y_{m,i}^s)\}_{i=1}^{N^s}$\end{small} and a query set \begin{small}$\datakq=\{(X_m^s, Y_m^s)\}=\{(x_{m,j}^q, y_{m,j}^q)\}_{i=1}^{N^q}$\end{small} are sampled from $\datak$. Representatively, in model-agnostic meta-learning (MAML)~\cite{finn2017meta}, given a predictive model $f$ with parameter \begin{small}$\theta$\end{small}, it aims to learn an initialization \begin{small}$\theta^{*}$\end{small} from meta-training tasks \begin{small}$\{\mathcal{T}_m\}_{m=1}^{|M|}$\end{small}. Specifically, at the meta-training phase, MAML obtained the task-specific parameter $\phi_m$ for each task $\task_m$ by performing a few gradient steps starting from $\theta$. Then, the corresponding query set $\datakq$ is used to evaluate the performance of the task-specific model and optimize the model initialization as:
\begin{equation}
\small
\label{eq:maml_optim}
    \theta^{*}:=\arg\min_{\theta} \frac{1}{|M|} \sum_{i=1}^{|M|} \loss(f_{\phi_m};\datakq),\;where\;\;\phi_m=\theta-\alpha \nabla_{\theta}\loss(f_{\theta};\dataks)
\end{equation}
At the meta-testing phase, for each meta-testing task $\taskt$, MAML fine-tunes the learned initialization $\theta^{*}$ on the support set $\datats$ and evaluates the performance on the corresponding query set $\datatq$.

To improve task generalization, MetaMix~\cite{yao2021improving} adapts mixup (Eqn.~\eqref{eq:mixup}) to meta-learning, which linearly interpolates the support set and query set and uses the interpolated set to replace the original query set \begin{small}$\mathcal{D}^q_m$\end{small} in Eqn.~\eqref{eq:maml_optim}. Specifically, the interpolated query set is formulated as:
\begin{equation}
\small
\label{eq:metamix}
    \Tilde{X}_m^q = \lambda X_m^s + (1-\lambda) X_m^q,\;\Tilde{Y}_m^q = \lambda Y_m^s + (1-\lambda) Y_m^q,
\end{equation}
where \begin{small}$\lambda \sim \mathrm{Beta}(\alpha, \alpha)$\end{small}.
\section{Mixup for Regression (C-Mixup)}
For continuous labels, the example in Figure~\ref{fig:iid_mixup}(b) illustrates that applying vanilla mixup to the entire distribution is likely to produce arbitrary labels. To resolve this issue, C-Mixup proposes to sample closer pairs of examples with higher probability. Specifically, given an example $(x_i, y_i)$, C-Mixup introduces a symmetric Gaussian kernel to calculate the sampling probability $P((x_j, y_j)|(x_i, y_i))$ for another $(x_j, y_j)$ example to be mixed as follows:
\begin{equation}
\label{eq:similarity}
\small
    P((x_j, y_j)|(x_i, y_i)) \propto  \exp\left(-\frac{d(i, j)}{2\sigma^2}\right)
\end{equation}
where $d(i,j)$ represents the distance  between the examples $(x_i, y_i)$ and $(x_j, y_j)$, and $\sigma$ describes the bandwidth. For the example $(x_i, y_i)$, the set $\{P((x_j, y_j)|(x_i, y_i))|\forall j \}$ is then normalized to a probability mass function that sums to one. 

\begin{algorithm}[H]
\caption{Training with C-Mixup}
\label{alg:C-Mixup}
\begin{algorithmic}[1]
\REQUIRE Learning rates $\eta$; Shape parameter $\alpha$
\REQUIRE Training data $\mathcal{D}:=\{(x_i, y_i)\}_{i=1}^N$
\STATE Randomly initialize model parameters $\theta$
\STATE Calculate pairwise distance matrix $P$ via Eqn.~\eqref{eq:similarity}
\WHILE{not converge}
\STATE Sample a batch of examples $\mathcal{B}\sim \mathcal{D}$
\FOR{each example $(x_i, y_i) \in \mathcal{B}$}
\STATE Sample ($x_j$, $y_j$) from $P(\cdot \mid (x_i, y_i))$ and $\lambda$ from $\mathrm{Beta}(\alpha, \alpha)$
\STATE Interpolate ($x_i$, $y_i$), ($x_j$, $y_j$) to get ($\Tilde{x}, \Tilde{y}$) according to Eqn.~\eqref{eq:interpolate}
\ENDFOR
\STATE Use interpolated examples to update the model via Eqn.~\eqref{eq:mixup}
\ENDWHILE
\end{algorithmic}
\end{algorithm}
One natural way to compute the distance is using the input feature $x$, i.e., $d(i,j)=d(x_i, x_j)$. However, when dealing with the high-dimensional data such as images or videos, we lack good distance metrics to capture structured feature information and the distances can be easily influenced by feature noise. Additionally, computing feature distances for high-dimensional data is time-consuming. Instead, C-Mixup leverages the labels with $d(i,j)=d(y_i, y_j)=\| y_i-y_j\|_2^2$, where $y_i$ and $y_j$ are vectors with continuous values. The dimension of label is typically much smaller than that of the input feature, therefore reducing computational costs (see more discussions about compuatational efficiency in Appendix~\ref{app:a_3}). The overall algorithm of C-Mixup is described in Alg.~\ref{alg:C-Mixup} and we detail the difference between C-Mixup and mixup in Appendix~\ref{app:mixup_detail}. According to Alg.~\ref{alg:C-Mixup}, C-Mixup assigns higher probabilities to example pairs with closer continuous labels.
In addition to its computational benefits, C-Mixup improves generalization on three distinct kinds of regression problems -- in-distribution generalization, task generalization, and out-of-distribution robustness, which is theoretically and empirically justified in the following sections.

\section{Theoretical Analysis}
In this section, we theoretically explain how C-Mixup benefits in-distribution generalization, task generalization, and out-of-distribution robustness.
\subsection{C-Mixup for Improving In-Distribution Generalization}
\label{sec:theory_iid}
In this section, we show that C-Mixup provably improves in-distribution generalization when the features are observed with noise, and the response depends on a small fraction of the features in a monotonic way. Specifically, we consider the following single index model with measurement error,
\begin{equation}
\small
    y=g(\theta^\top z)+\epsilon,
\end{equation}
where $\theta\in\mathbb R^p$ and $\epsilon$ is a sub-Gaussian random variable and $g$ is a monotonic transformation. Since images are often inaccurately observed in practice, we assume the feature $z$ is observed or measured with noise, and denote the observed value by $x$: $x=z+\xi$ with $\xi$ being a random vector with mean 0 and covariance matrix $\sigma_{\xi}^2 I$.
We assume $g$ to be monotonic to model the nearly one-to-one correspondence between causal features (e.g., the car pose in Figure~\ref{fig:iid_mixup}(a)) and labels (rotation) in the in-distribution setting. The out-of-distribution setting will be discussed in Section~\ref{sec:theorey:ood}. We would like to also comment that the single index model has been commonly used in econometrics, statistics, and deep learning theory~\cite{ge2019learning,horowitz2009semiparametric,lyu2019gradient,powell1989semiparametric,yang2017high}. 

Suppose we have $\{(x_i,y_i)\}_{i=1}^N$ i.i.d. drawn from the above model. We first follow the single index model literature (e.g. \cite{yang2017high}) and estimate $\theta$ by minimizing the square error \begin{small}$\sum_{i=1}^n (\tilde y_i-\tilde x_i^\top\theta)^2$\end{small}, where the $(\tilde x_i, \tilde y_i)'s$ are the augmented data by either vanilla mixup, mixup with input feature similarity, and C-Mixup. We denote the solution by $\theta^{*}_{mixup}$, $\theta^{*}_{feat}$, and $\theta^{*}_{C-Mixup}$ respectively. Given an estimate $\theta^*$, we estimate $g$ by $\hat g$ via the standard nonparametric kernel estimator \citep{tsybakov2004introduction} (we specify this in detail in Appendix~\ref{app:b_1} for completeness) using the augmented data. We consider the mean square error metric as \begin{small}$\mathrm{MSE}(\theta)=\E[(y-\hat g(\theta^\top x))^2]$\end{small}, and then have the following theorem (proof: Appendix~\ref{app:b_1}):
\begin{theorem}
Suppose $\theta\in\R^p$ is sparse with sparsity $s=o(\min\{p,\sigma_{\xi}^2\})$, $p=o(N)$ and $g$ is smooth with $c_0<g'<c_1$, $c_2<g''<c_3$ for some universal constants $c_0, c_1,c_2, c_3>0$. There exists a distribution on $x$ with a kernel function, such that when the sample size $N$ is sufficiently large, with probability $1-o(1)$, 
\begin{equation}
\small
    \mathrm{MSE}(\theta^{*}_{C-Mixup})<\min(\mathrm{MSE}(\theta^{*}_{feat}),\mathrm{MSE}(\theta^{*}_{mixup})).
\end{equation}
\end{theorem}

The high-level intuition of why C-Mixup helps is that the vanilla mixup imposes linearity regularization on the relationship between the feature and response. When the relationship is strongly nonlinear and one-to-one, such a regularization hurts the generalization, but could be mitigated by C-Mixup.

\subsection{C-Mixup for Improving Task Generalization}
The second benefit of C-Mixup is improving task generalization in meta-learning when the data from each task follows the model discussed in the last section. Concretely, we apply C-Mixup to MetaMix \citep{yao2021improving}. For each query example, the support example with a more similar label will have a higher probability of being mixed. The algorithm of C-Mixup on MetaMix is summarized in Appendix~\ref{app:a_2}.

Similar to in-distribution generalization analysis, we consider the following data generative model: for the $m$-th task ($m \in [M]$), we have $(x^{(m)},y^{(m)})\sim\mathcal T_m$ with
\begin{equation}
\small
y^{(m)}=g_m(\theta^\top z^{(m)})+\epsilon \quad \text{ and } \quad x^{(m)}=z^{(m)}+\epsilon^{(m)}.
\end{equation}
Here, $\theta$ denotes the globally-shared representation, and $g_m$'s are the task-specific transformations. Note that, the formulation is close to~\cite{raghu2020rapid} and wildly applied to theoretical analysis of meta-learning~\cite{yao2021improving,yao2022meta}. Following similar spirit of \cite{tripuraneni2021provable} and last section, we obtain the estimation of \begin{small}$\theta$\end{small} by
\begin{small}$\small
    \theta^*=\frac{1}{M}\sum_{m=1}^M (\E_{(\tilde x^{(m)},\tilde y^{(m)})\in \mathcal{\hat{D}}_m}[\tilde y^{(m)}-\theta^\top \tilde x^{(m)}]).
$\end{small}
Here, \begin{small}$\mathcal{\hat{D}}_m$\end{small} denotes the generic dataset augmented by different approaches, including the vanilla MetaMix, MetaMix with input feature similarity, and C-Mixup. We denote these approaches by \begin{small}$ \theta^{*}_{MetaMix}$\end{small}, \begin{small}$\theta^{*}_{Meta-feat}$\end{small}, and \begin{small}$\theta^{*}_{Meta-C-Mixup}$\end{small} respectively. For a new task \begin{small}$\mathcal{T}_t$\end{small}, we again use the standard nonparametric kernel estimator to estimate $g_{t}$ via the augmented target data. We then consider the following error metric 
\begin{small}$
\small
    \mathrm{MSE}_{\mathrm{Target}}(\theta^{*})=\E_{(x,y)\sim\mathcal T_{t}}[(y-\hat g_{t}({\theta^{*}}^\top x))^2].
$\end{small} 

Based on this metric, we get the following theorem to show the promise of C-Mixup in improving task generalization (see Appendix~\ref{app:b_2} for detailed proof). Here, C-Mixup achieves smaller $\mathrm{MSE}_{\mathrm{Target}}$ compared to vanilla MetaMix and MetaMix with input feature similarity.
\begin{theorem}
Let $N=\sum_{m=1}^M N_m$ and $N_m$ is the number of examples of $\mathcal{T}_m$. Suppose $\theta_k$ is sparse with sparsity $s=o(\min\{d,\sigma_{\xi}^2\})$, $p=o(N)$ and $g_m$'s are smooth with $0<g_m'<c_1$, $c_2<g_m''<c_3$ for some universal constants $c_1,c_2, c_3>0$ and $m\in[M]\cup\{t\}$.  There exists a distribution on $x$ with a kernel function, such that when the sample size $N$ is sufficiently large, with probability $1-o(1)$, 
\begin{equation}
\small
    \mathrm{MSE}_{\mathrm{Target}}(\theta^{*}_{Meta-C-Mixup})<\min(\mathrm{MSE}_{\mathrm{Target}}(\theta^{*}_{Meta-feat}),\mathrm{MSE}_{\mathrm{Target}}(\theta^{*}_{MetaMix})).
\end{equation}
\end{theorem}

\subsection{C-Mixup for Improving Out-of-distribution Robustness}\label{sec:theorey:ood}
Finally, we show that C-Mixup improves OOD robustness in the covariate shift setting where some unrelated features vary across different domains. In this setting, we regard the entire data distribution consisting of $\mathcal{E}=\{1,\ldots,E\}$ domains, where each domain is associated with a data distribution $P_e$ for $e\in\mathcal E$. Given a set of training domains $\mathcal{E}^{tr} \subseteq \mathcal{E}$, we aim to make the trained model generalize well to an unseen test domain  $\mathcal{E}^{ts}$ that is not necessarily in $\mathcal{E}^{tr}$. Here, we focus on covariate shift, i.e., the change of $P_e$ among domains is only caused by the change of marginal distribution $P_e(X)$, while the conditional distribution $P_e(Y|X)$ is fixed across different domains.

To overcome covariate shift, mixing examples with close labels without considering domain information can effectively average out domain-changeable correlations and make the predicted values rely on the invariant causal features. To further understand how C-Mixup improves the robustness to covariate shift, we provide the following theoretical analysis. 

We assume the training data $(x_i,y_i)_{i=1}^{n}$ follows $x_i=(z_i;a_i)\in\R^{p_1+p_2}$ and $y_i=\theta^\top x_i+\epsilon_i$, where $z_i\in\R^{p_1}$ and $a_i\in\R^{p_2}$ are regarded as invariant and domain-changeable unrelated features, respectively, and the last $p_2$ coordinates of $\theta\in\R^{p_1+p_2}$ are 0. Now we consider the case where the training data consists of a pair of domains with almost identical invariant features and opposite domain-changeable features, i.e., $x_i=(z_i,a_i),x_i'=(z_i', a_i')$, where \begin{small}$z_i \sim \mathcal{N}_{p_1}(0,\sigma_x^2I_{p_1})$, $z_i'=z_i+\epsilon_{i}'$\end{small}, \begin{small}$a_i \sim \mathcal{N}_{p_2}(0,\sigma_a^2I_{p_2})$\end{small}, $a_i'=-a_i + \epsilon_{i}''$. $\epsilon_i, \epsilon_i', \epsilon_i{''}$ are noise terms with mean 0 and sub-Gaussian norm bounded by $\sigma_\epsilon$. We use ridge estimator \begin{small}$\theta^{*}(k) = \mathop{\arg\min}_{\theta}({\sum_i\|{y_i - \theta^\top x_i}\|^2} + k\| \theta\|^2)$\end{small} to reflect the implicit regularization effect of deep neural networks~\cite{Neyshabur17regular}. The covariant shift happens when the test domains have different $a_i$ distributions compared with training domains. We then have the following theorem to show that C-Mixup can improve robustness to covariant shift (see proof in Appendix~\ref{app:b_3}):

\begin{theorem}
Supposed for some \begin{small}$\max(\exp(-n^{1-o(1)}), \exp(-\frac{p_1^2}{2n})) < \delta \ll 1$\end{small}, we have variance constraints:\begin{small} $\sigma_a = c_1 \sigma_x$ , $\sigma_x \geq c_2 \max(\frac{n^{5/2}}{\Vert{\theta}\Vert\delta}\sigma_{\epsilon}, \frac{\sqrt{p_2}\Vert{\theta}\Vert}{\sqrt{n}p_1})$\end{small} and \begin{small}
$\sigma_{\epsilon}^2  \leq \frac{c_3}{pn^{3/2}}$
\end{small}. Then for any penalty k satisfies \begin{small}$c_4\sqrt{\frac{p_2}{p_1}}n^{1/4+o(1)} < k < c_5 \min(\frac{\sigma_x}{\Vert{\theta}\Vert}\sqrt{p_1n^{1-o(1)}}, n)$\end{small} and bandwidth h satisfies \begin{small}$0 < h \leq c_6 \frac{l}{\sqrt{\log(n^2/p_1)}} $\end{small} in C-Mixup, when $n$ is sufficiently large, with probability at least $1-o(1)$, we have
\begin{equation}
\small
\mathrm{MSE}(\theta^{*}_{C-Mixup})< \min(\mathrm{MSE}(\theta^{*}_{feat}), \mathrm{MSE}(\theta^{*}_{mixup})),
\end{equation}
where $c_1 \geq 1$, $c_2$, $c_3$, $c_4$, $c_5$, $c_6$ > 0 are  universal constants, \begin{small}$l = \min_{i\neq j}\lvert y_i - y_j^{\prime}\rvert$\end{small} and 
\begin{small}$p_1 \ll n < p_1^2$\end{small}.

\end{theorem}

\section{Experiments}
\label{sec:exp}
In this section, we evaluate the performance of C-Mixup, aiming to answer the following questions: \textbf{Q1:} Compared to corresponding prior approaches, can C-Mixup improve the in-distribution, task generalization, and out-of-distribution robustness on regression? \textbf{Q2:} How does C-Mixup perform compared to using other distance metrics? \textbf{Q3:} Is C-Mixup sensitive to the choice of bandwidth $\sigma$ in the Gaussian kernel of Eqn.~\eqref{eq:similarity}? In our experiments, we apply cross-validation to tune all hyperparameters with grid search.

\subsection{In-Distribution Generalization}
\label{sec:exp_iid}
\textbf{Datasets.} We use the following five datasets to evaluate the performance of in-distribution generalization (see Appendix~\ref{app:c_1} for detailed data statistics). \textbf{(1)\&(2) Airfoil Self-Noise (Airfoil) and NO2~\cite{kooperberg1997statlib}} are both are tabular datasets, where airfoil contains aerodynamic and acoustic test results of airfoil blade sections and NO2 aims to predict the mount of air pollution at a particular location. \textbf{(3)\&(4): Exchange-Rate, and Electricity~\cite{lai2018modeling}} are two time-series datasets, where Exchange-Rate reports the collection of the daily exchange rates and Electricity is used to predict the hourly electricity consumption. \textbf{(5) Echocardiogram Videos (Echo)~\cite{ouyang2020video}} is a ejection fraction prediction dataset, which consists of a series of videos illustrating the heart from different aspects.

\textbf{Comparisons and Experimental Setups.} We compare C-Mixup with mixup and its variants (Manifold mixup~\cite{verma2019manifold}, k-Mixup~\cite{greenewald2021k} and Local Mixup~\cite{baena2022preventing}) that can be easily to adapted to regression tasks. We also compare to MixRL, a recent reinforcement learning framework to select mixup pairs in regression. Note that, for k-Mixup, Local Mixup, MixRL, and C-Mixup, we apply them to both mixup and Manifold Mixup and report the best-performing combination.

For Airfoil and NO2, we employ a three-layer fully connected network as the backbone model. We use LST-Attn~\cite{lai2018modeling} for the Exchange-Rate and Electricity, and EchoNet-Dynamic~\cite{ouyang2020video} for predicting the ejection fraction. We use Root Mean Square Error (RMSE) and Mean Averaged Percentage Error (MAPE) as evaluation metrics. Detailed experimental setups are in Appendix~\ref{app:c_2}.

\begin{wrapfigure}{r}{0.5\textwidth}
\vspace{-1.5em}
\centering
\includegraphics[width=0.24\textwidth]{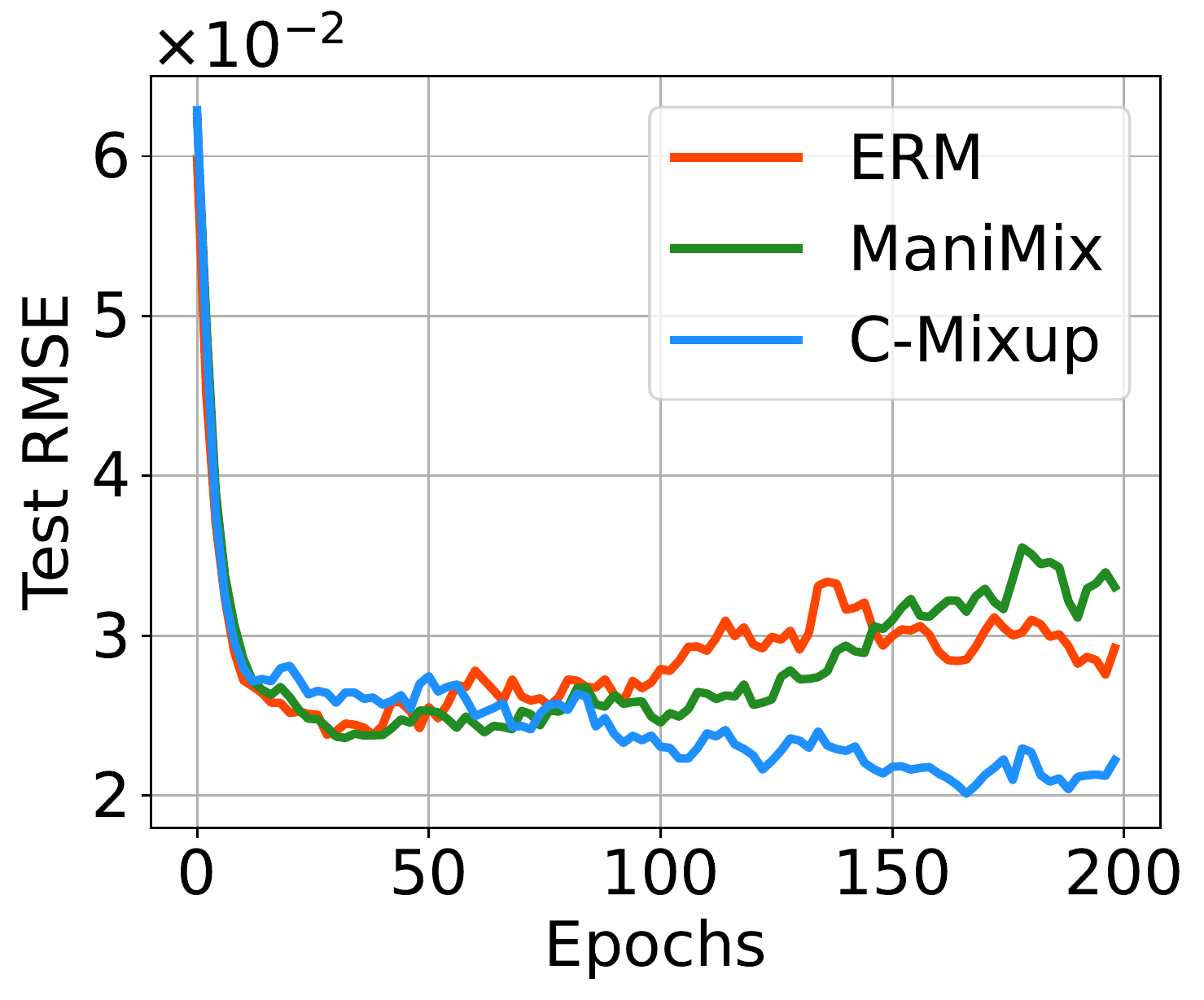}
\includegraphics[width=0.25\textwidth]{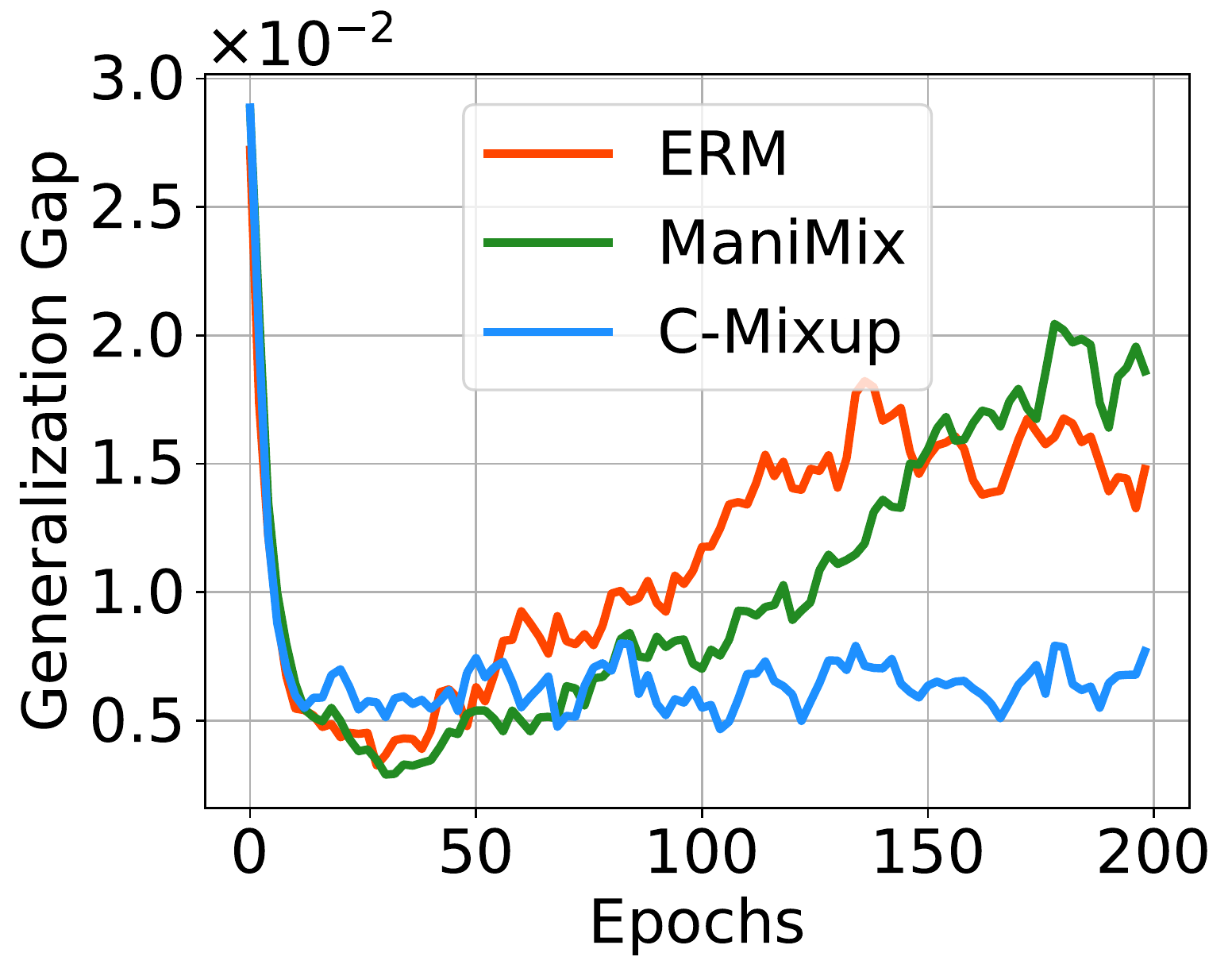}
\vspace{-1.5em}
\caption{Overfitting Analysis on Exchange-Rate. C-Mixup achieves better test performance and smaller generalization gap.}
\label{fig:overfitting}
\vspace{-1em}
\end{wrapfigure}
\textbf{Results.}
We report the results in Table~\ref{tab:results_id} and have the following observations. \emph{First}, vanilla mixup and manifold mixup are typically less performant than ERM when they are applied directly to regression tasks. These results support our hypothesis that random selection of example pairs for mixing may produce arbitrary inaccurate virtual labels. \emph{Second}, while limiting the scope of interpolation (e.g., k-Mixup, Manifold k-Mixup, Local Mixup) helps to improve generalization in most cases, the performance of these approaches is inconsistent across different datasets. As an example, Local Mixup outperforms ERM in NO2 and Exchange-Rate, but fails to benefit results in Electricity. \emph{Third}, even in more complicated datasets, such as Electricity and Echo, MixRL performs worse than Mixup and ERM, indicating that it is non-trivial to train a policy network that works well. \emph{Finally}, C-Mixup consistently outperforms mixup and its variants, ERM, and MixRL, demonstrating its capability to improve in-distribution generalization on regression problems.

\textbf{Analysis of Overfitting.} In Figure~\ref{fig:overfitting}, we visualize the test loss (RMSE) and the generalization gap between training loss and test loss of ERM, Manifold Mixup, and C-Mixup with respect to the training epoch for Exchange-Rate. More results are illustrated in Appendix~\ref{app:c_3}. Compared with ERM and C-Mixup, the better test performance and smaller generalization gap of C-Mixup further demonstrates its ability to improve in-distribution generalization and mitigate overfitting.

\begin{table}[t]
\small
\caption{Results for in-distribution generalization. We report the average RMSE and MAPE of three seeds. Full results with standard deviation are reported in Appendix~\ref{app:c_4}. The best results and second best results are \textbf{bold} and \underline{underlined}, respectively.}
\label{tab:results_id}
\begin{center}
\resizebox{\columnwidth}{!}{\setlength{\tabcolsep}{1.5mm}{
\begin{tabular}{l|c|c|c|c|c|c|c|c|c|c}
\toprule
& \multicolumn{2}{c|}{Airfoil} & \multicolumn{2}{c|}{NO2} & \multicolumn{2}{c|}{Exchange-Rate} & \multicolumn{2}{c|}{Electricity} & \multicolumn{2}{c}{Echo} \\\cmidrule{2-11}
& RMSE & MAPE & RMSE & MAPE & RMSE & MAPE & RMSE & MAPE & RMSE & MAPE \\\midrule
ERM & \underline{2.901} & \underline{1.753\%} & 0.537 & 13.615\% & 0.0236 & 2.423\% & 0.0581 & \underline{13.861}\% & 5.402 & \underline{8.700}\%\\
mixup & 3.730 & 2.327\%& 0.528 & 13.534\% & 0.0239 & 2.441\% & 0.0585 & 14.306\% & \underline{5.393} & 8.838\%\\
Mani mixup & 3.063 & 1.842\% & 0.522 & 13.382\% & 0.0242 & 2.475\%  & 0.0583 & 14.556\% & 5.482 & 8.955\%\\
k-Mixup & 2.938 & 1.769\% & 0.519 & \underline{13.173\%} & 0.0236 & 2.403\% & \underline{0.0575} & 14.134\% & 5.518 & 9.206\% \\
Local Mixup & 3.703 & 2.290\% & \underline{0.517}  & 13.202\% & \underline{0.0236} & \underline{2.341\%} & 0.0582 & 14.245\% & 5.652 & 9.313\% \\
MixRL & 3.614 & 2.163\% & 0.527 & 13.298\% & 0.0238 & 2.397\% & 0.0585 & 14.417\% & 5.618 & 9.165\% \\
\midrule
\textbf{C-Mixup (Ours)} & \textbf{2.717}& \textbf{1.610\%} & \textbf{0.509} & \textbf{12.998\%} &  \textbf{0.0203} & \textbf{2.041\%} & \textbf{0.0570} & \textbf{13.372\%} & \textbf{5.177} & \textbf{8.435\%}\\
\bottomrule
\end{tabular}}}
\end{center}
\vspace{-1.5em}
\end{table}

\vspace{-0.3em}
\subsection{Task Generalization}
\label{sec:exp_task}
\textbf{Dataset and Experimental Setups.}
To evaluate task generalization in meta-learning settings, we use
two rotation prediction datasets named as ShapeNet1D~\cite{gao2022matters} and Pascal1D~\cite{xiang2014beyond,yin2020meta}, the goal of both datasets is to predict an object's rotation relative to the canonical orientation. Each task is rotation regression for one object, where the model takes a 128×128 grey-scale image as the input, and the output is an azimuth angle normalized between $[0, 10]$. We detail the description in Appendix~\ref{app:d_1}.

Since MetaMix outperforms most other methods in PASCAL3D in~\citet{yao2021improving}, we only select two other representative approaches -- Meta-Aug~\cite{rajendran2020meta} and MR-MAML~\cite{yin2020meta} for comparison. We further apply k-Mixup and Local Mixup to MetaMix, which are called as k-MetaMix and Local MetaMix, respectively. Following~\citet{yin2020meta}, the base model consists of an encoder with three convolutional blocks and a decoder with four convolutional blocks. Hyperparameters are listed in Appendix~\ref{app:d_2}.

\begin{wraptable}{r}{0.5\textwidth}
\begin{center}
\small
\vspace{-2em}
\caption{Meta-regression performance (MSE $\pm$ 95\% confidence interval) on 6000 meta-test tasks. }
\vspace{-0.5em}
\label{tab:meta_regression_results}
\begin{tabular}{l|cc}
\toprule
Model & ShapeNet1D $\downarrow$ & PASCAL3D $\downarrow$ \\\midrule
MAML & 4.698 $\pm$ 0.079 & 2.370 $\pm$ 0.072 \\
MR-MAML & 4.433 $\pm$ 0.083 & 2.276 $\pm$ 0.075\\
Meta-Aug & 4.312 $\pm$ 0.086 & 2.298 $\pm$ 0.071\\
MetaMix & 4.275 $\pm$ 0.082 &  2.135 $\pm$ 0.070\\
k-MetaMix & 4.268 $\pm$ 0.078 & \underline{2.091 $\pm$ 0.069}\\
Local MetaMix & \underline{4.201 $\pm$ 0.087} & 2.107 $\pm$ 0.078\\
\midrule
\textbf{C-Mixup (Ours)} & \textbf{4.024 $\pm$ 0.081} & \textbf{1.995 $\pm$ 0.067} \\\bottomrule
\end{tabular}
\end{center}
\vspace{-1.5em}
\end{wraptable}
\textbf{Results.} Table~\ref{tab:meta_regression_results} shows the average MSE with a 95\% confidence interval over 6000 meta-testing tasks. Our results corroborate the findings of~\citet{yao2021improving} that MetaMix improves performance compared with non-mixup approaches (i.e., MAML, MR-MAML, Meta-Aug). Similar to our previous in-distribution generalization findings, the better performance of Local MetaMix over MetaMix demonstrates the effectiveness of interpolating nearby examples. By mixing query examples with support examples that have similar labels, C-Mixup outperforms all of the other approaches on both datasets, verifying its effectiveness of improving task generalization.

\vspace{-0.3em}
\subsection{Out-of-Distribution Robustness}

\begin{wrapfigure}{r}{0.47\textwidth}
\vspace{-1em}
\centering
\includegraphics[width=0.47\textwidth]{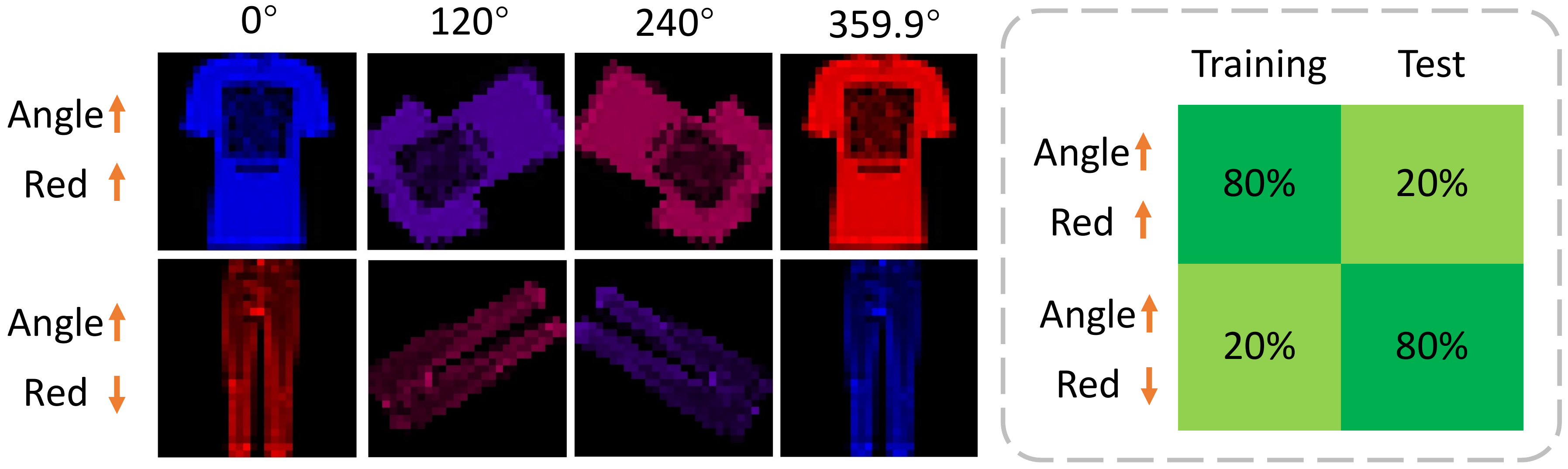}
\caption{Illustration of RCF-MNIST}
\label{fig:rcfmnist}
\vspace{-1em}
\end{wrapfigure}
\textbf{Synthetic Dataset with Subpopulation Shift.} 
Similar to synthetic datasets with subpopulation shifts in classification, e.g., ColoredMNIST~\cite{arjovsky2019invariant}, we build a synthetic regression dataset -- RCFashion-MNIST (\textbf{RCF-MNIST}), which is illustrated in Figure~\ref{fig:rcfmnist}. Built on the FashionMNIST~\cite{gao2022matters}, the goal of RCF-MNIST is to predict the angle of rotation for each object. As shown in Figure~\ref{fig:rcfmnist}, we color each image with a color between red and blue. There is a spurious correlation between angle (label) and color in training set. The larger the angle, the more red the color. In test set, we reverse spurious correlations to simulate distribution shift. We provide detailed description in Appendix~\ref{app:e_1}. 

\textbf{Real-world Datasets with Domain Shifts.} In the following, we briefly discuss the real-world datasets with domain shifts; the detailed data descriptions in Appendix~\ref{app:e_1}: \textbf{(1) PovertyMap~\cite{koh2021wilds}} is a satellite image regression dataset, aiming to estimate asset wealth in countries that are not shown in the training set. \textbf{(2) Communities and Crime (Crime)}~\cite{misc_communities_and_crime_183} is a tabular dataset, where the problem is to predict total number of violent crimes per 100K population and we aim to generalize the model to unseen states. \textbf{(3) SkillCraft1 Master Table (SkillCraft)}~\cite{misc_skillcraft1_master_table_dataset_272} is a tabular dataset, aiming to predict the mean latency from the onset of a perception action cycles to their first action in milliseconds. Here, ``LeagueIndex" is treated as domain information. \textbf{(4) Drug-target Interactions (DTI)}~\cite{tdc} is aiming to predict out-of-distribution drug-target interactions in 2019-2020 after training on 2013-2018.

\textbf{Comparisons and Experimental Setups.} To evaluate the out-of-distribution robustness of C-Mixup, we compare it with nine invariant learning approaches that can be adapted to regression tasks, including ERM, IRM~\cite{arjovsky2019invariant}, IB-IRM~\cite{ahuja2021invariance}, V-REx~\cite{krueger2021out}, CORAL~\cite{li2018domain}, DRNN~\cite{ganin2015unsupervised}, GroupDRO~\cite{sagawa2019distributionally}, Fish~\cite{shi2021gradient}, and mixup. All approaches use the same backbone model, where we adopt ResNet-50, three-layer full connected network, and DeepDTA~\cite{ozturk2018deepdta} for Poverty, Crime and SkillCraft, and DTI, respectively. Following the original papers of PovertyMap~\cite{yeh2020using} and DTI~\cite{tdc}, we use $R$ value to evaluate the performance. For RCF-MNIST, Crime and SkillCraft, we use RMSE as the evaluation metric. In datasets with domain shifts, we report both average and worst-domain (primary metric for PovertyMap~\cite{koh2021wilds}) performance. Detailed hyperparameters are listed in Appendix~\ref{app:e_2}.

\begin{table*}[t]
\small
\caption{Results for out-of-distribution robustness. We report the average and worst-domain (primary metric) performance here and the full results are listed in Appendix~\ref{app:e_3}. Sub. Shift means Subpopulation Shift. Higher $R$ or lower RMSE represent better performance. mixup and C-Mixup uses the same type of mixup variants reported in Table~\ref{app:e_2}. For PovertyMap, most results are copied from WILDS benchmark~\cite{koh2021wilds} and worst-domain performance is the primary metric. We bold the best results and underline the second best results.}
\vspace{-1em}
\label{tab:results_ood}
\begin{center}
\resizebox{\columnwidth}{!}{\setlength{\tabcolsep}{1.5mm}{\begin{tabular}{l|c|cc|cc|cc|cc}
\toprule
\multirow{3}{*}{} & Sub. Shift & \multicolumn{8}{c}{Domain Shift} \\\cmidrule{2-10}
 & RCF-MNIST & \multicolumn{2}{c|}{PovertyMap ($R$)} & \multicolumn{2}{c|}{Crime (RMSE)} & \multicolumn{2}{c|}{SkillCraft (RMSE)} & \multicolumn{2}{c}{DTI ($R$)} \\
& Avg. (RMSE) $\downarrow$ & Avg. $\uparrow$ & Worst $\uparrow$ & Avg. $\downarrow$ & Worst $\downarrow$ & Avg. $\downarrow$ & Worst $\downarrow$ & Avg. $\uparrow$ & Worst $\uparrow$\\\midrule
ERM & 0.162 & 0.80 & \underline{0.50} & 0.134 & 0.173 &  5.887 & 10.182 &  0.464  & 0.429 \\
IRM & \underline{0.153} & 0.77 & 0.43 & \underline{0.127} & 0.155 & 5.937 & 7.849 &  0.478   &   0.432  \\
IB-IRM & 0.167 & 0.78 & 0.40 & 0.127 & 0.153 & 6.055 & 7.650 &  0.479  & 0.435 \\
V-REx & 0.154 & \textbf{0.83} & 0.48 & 0.129 & 0.157 & 6.059 & \underline{7.444} &  \underline{0.485} & 0.435 \\
CORAL & 0.163 & 0.78 & 0.44 & 0.133 & 0.166 & 6.353 & 8.272 & 0.483 & 0.432\\
GroupDRO & 0.232 & 0.75 & 0.39 & 0.138 & 0.168 & 6.155 & 8.131 & 0.442  &  0.407 \\
Fish & 0.263 & 0.80 & 0.30 & 0.128 & \underline{0.152} & 6.356 & 8.676 &  0.470  & \underline{0.443} \\
mixup & 0.176 & \underline{0.81} & 0.46 & 0.128 & 0.154 & \underline{5.764} & 9.206 &  0.465 &  0.437 \\\midrule
\textbf{C-Mixup (Ours)} & \textbf{0.146}& \underline{0.81} & \textbf{0.53} & \textbf{0.123} & \textbf{0.146} & \textbf{5.201} & \textbf{7.362} & \textbf{0.498} & \textbf{0.458} \\
\bottomrule
\end{tabular}}}
\end{center}
\vspace{-2.5em}
\end{table*}

\textbf{Results.} We report both average and worst-domain performance in Table~\ref{tab:results_ood}. According to the results, we can see that the performance of prior invariant learning approaches is not stable across datasets. For example, IRM and CORAL outperform ERM on Camelyon17, but fail to improve the performance on PovertyMap. C-Mixup instead consistently shows the best performance regardless of the data types, indicating its effeacy in improving robustness to covariate shift.

\begin{wrapfigure}{r}{0.4\textwidth}
\vspace{-2em}
\centering
\includegraphics[width=0.35\textwidth]{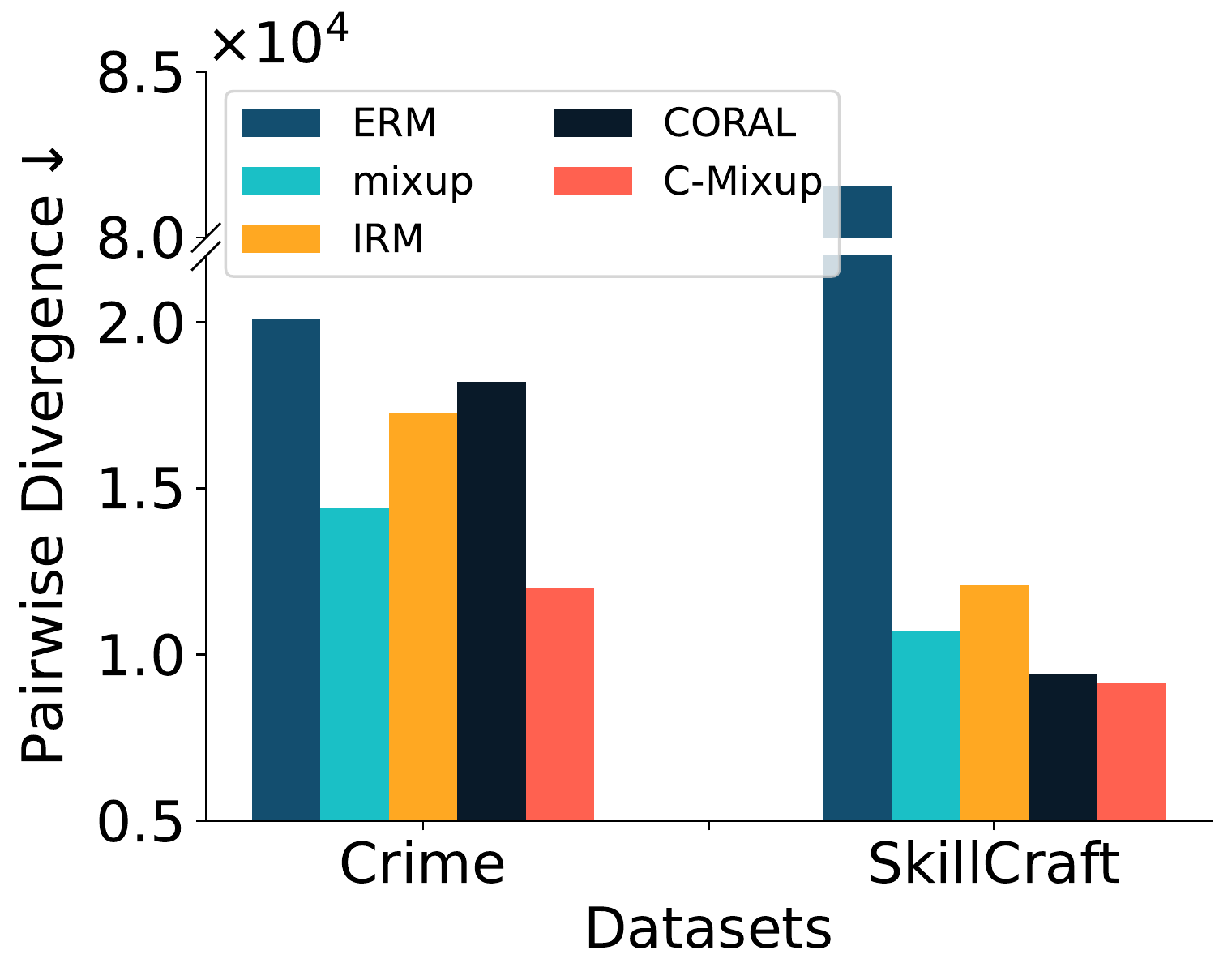}
\caption{Results of invariance analysis. Smaller pairwise divergence value represents stronger invariance.}
\label{fig:invariance}
\vspace{-2em}
\end{wrapfigure}

\textbf{Analysis of Learned Invariance.}
Following~\citet{yao2022improving}, we analyze the domain invariance of the model learned by C-Mixup. For regression, we measure domain invariance as pairwise divergence of the last hidden representation. Since the labels are continuous, we first evenly split examples into $\mathcal{C}=\{1,\ldots,C\}$ bins according to their labels, where $C=10$. After splitting, we perform kernel density estimation to estimate the probability density function $P(h_e^{c})$ of the last hidden representation. We calculate the pairwise divergence as \begin{small}$\mathrm{Inv}=\frac{1}{|\mathcal{C}||\mathcal{E}|^2}\sum_{c\in \mathcal{C}}\sum_{e',e\in \mathcal{E}}\mathrm{KL}(P(g_E^c\mid E=e)|P(g_E^c\mid E=e'))$\end{small}. The results on Crime and SkillCraft are reported in Figure~\ref{fig:invariance}, where smaller number denotes stronger invariance. We observe that C-Mixup learns better invariant representation compared to prior invariant learning approaches.

\subsection{Analysis of C-Mixup}
In this section, we conduct three analyses including the compatibility of C-Mixup, alternative distance metrics, and sensitivity of bandwidth. In Appendix~\ref{sec:app_robust_alpha} and~\ref{sec:app_robustness}, we analyze the sensitivity of hyperparameter $\alpha$ in Beta distribution and the robustness of C-Mixup to label noise, respectively.

\begin{wraptable}{r}{0.45\textwidth}
\small
\vspace{-1.5em}
\caption{Compatibility analysis. See Appendix~\ref{sec:app_compatibility} for full results.}
\vspace{-1em}
\label{tab:compatibility}
\begin{center}
\setlength{\tabcolsep}{1.5mm}{\resizebox{0.45\textwidth}{!}{
\begin{tabular}{l|c|c|c}
\toprule
\multicolumn{2}{c|}{\multirow{2}{*}{Model}} & RCF-MNIST & PovertyMap
\\\cmidrule{3-4}
\multicolumn{2}{c|}{} & RMSE $\downarrow$  & Worst $R$ $\uparrow$\\\midrule
\multirow{2}{*}{CutMix} & & 0.194 & 0.46 \\
  & +C-Mixup  & \textbf{0.186}  & \textbf{0.53}\\\midrule
\multirow{2}{*}{PuzzleMix}  &   & 0.159 & 0.47\\
  & +C-Mixup  & \textbf{0.150} & \textbf{0.50} \\\midrule
\multirow{2}{*}{AutoMix} &   & 0.152 & 0.49 \\
 & +C-Mixup  & \textbf{0.146} & \textbf{0.53} \\
\bottomrule
\end{tabular}}}
\end{center}
\vspace{-2em}
\end{wraptable}
\textbf{I. Compatibility of C-Mixup.}
C-Mixup is a complementary 
approach to vanilla mixup and its variants, where it changes the probabilities of sampling mixing pairs instead of changing the way to mixing. We further conduct compatibility analysis of C-Mixup by integrating it to three representative mixup variants -- PuzzleMix~\cite{kim2020puzzle}, CutMix~\cite{yun2019cutmix}, AutoMix~\cite{liu2021unveiling}. We evaluate the performance on two datasets (RCF-MNIST, PovertyMap). The reported results in Table~\ref{tab:compatibility} indicate the compatibility and efficacy of C-Mixup in regression.

\begin{wraptable}{r}{0.45\textwidth}
\small
\caption{Performance of different distance metrics. $x$, $y$, $h$ represent input feature, label, and hidden representation, respectively. The full results are listed in Appendix~\ref{sec:app_distance}.}
\vspace{-0.5em}
\label{tab:similarity_comparison}
\begin{center}
\setlength{\tabcolsep}{1.6mm}{\resizebox{0.45\textwidth}{!}{
\begin{tabular}{l|c|c|c}
\toprule
\multirow{2}{*}{Model}  & Ex.-Rate & Shape1D & DTI \\\cmidrule{2-4}
& RMSE $\downarrow$ & MSE $\downarrow$ & Avg. $R$ $\uparrow$
\\\midrule
ERM/MAML & 0.0236  & 4.698 & 0.464  \\
mixup/MetaMix  & 0.0239 &  4.275 & 0.465\\\midrule
$d(x_i, x_j)$  & 0.0212 & 4.539 & 0.478\\
$d(x_i\oplus y_i, x_j\oplus y_j)$  & 0.0212 & 4.395 & 0.484\\
$d(h_i, h_j)$ & 0.0213  & 4.202 & 0.483\\
$d(h_i\oplus y_i, h_j\oplus y_j)$  & 0.0208  & 4.176 & 0.487 \\

\midrule
\textbf{$d(y_i, y_j)$ (C-Mixup)}  &  \textbf{0.0203} & \textbf{4.024} & \textbf{0.498} \\
\bottomrule
\end{tabular}}}
\end{center}
\vspace{-2em}
\end{wraptable}

\textbf{II. Analysis of Distance Metrics.} 
Besides the theoretical analysis, here we empirically analyze the effectiveness of using label distance. Here, we use $d(a,b)$ to denote the distance between objects $a$ and $b$, e.g., $d(y_i, y_j)$ in our case. We consider four substitute distance metrics, including: (1) \emph{feature distance}: $d(x_i, x_j)$; (2) \emph{feature and label distance}: we concatenate the input feature $x$ and label $y$ to compute the distance, i.e., $d(x_i\oplus y_i, x_j\oplus y_j)$; (3) \emph{representation distance}: since using feature distance may fail to capture high-level feature relations, we propose to use the model's hidden representations to measure the distance $d(h_i, h_j)$, detailed in Appendix~\ref{sec:app_distance}. (4) \emph{representation and label distance}: $d(h_i\oplus y_i, h_j\oplus y_j)$.

The results are reported in Table~\ref{tab:similarity_comparison}. We find that (1) feature distance $d(x_i, x_j)$ does benefit performance compared with ERM and vanilla mixup in all cases since it selects more reasonable example pairs for mixing; (2) representation distance $d(h_i, h_j)$ outperforms feature distance $d(x_i, x_j)$ since it captures high-level features; (3) though involving label in representation distance is likely to overwhelm the effect of labels, the better performance of $d(h_i \oplus y_i, h_j \oplus y_j)$ over $d(h_i, h_j)$ indicates the effectiveness of regression labels in measuring example distance, which is further verified by the superiority of C-Mixup. (4) Our discussion of computational efficiency in Appendix~\ref{app:a_3} indicates that C-Mixup is much more computationally efficient than other distance measurements. 

\begin{wrapfigure}{r}{0.46\textwidth}
\vspace{-0.5em}
\centering
\begin{subfigure}[c]{0.225\textwidth}
		\centering
\includegraphics[width=\textwidth]{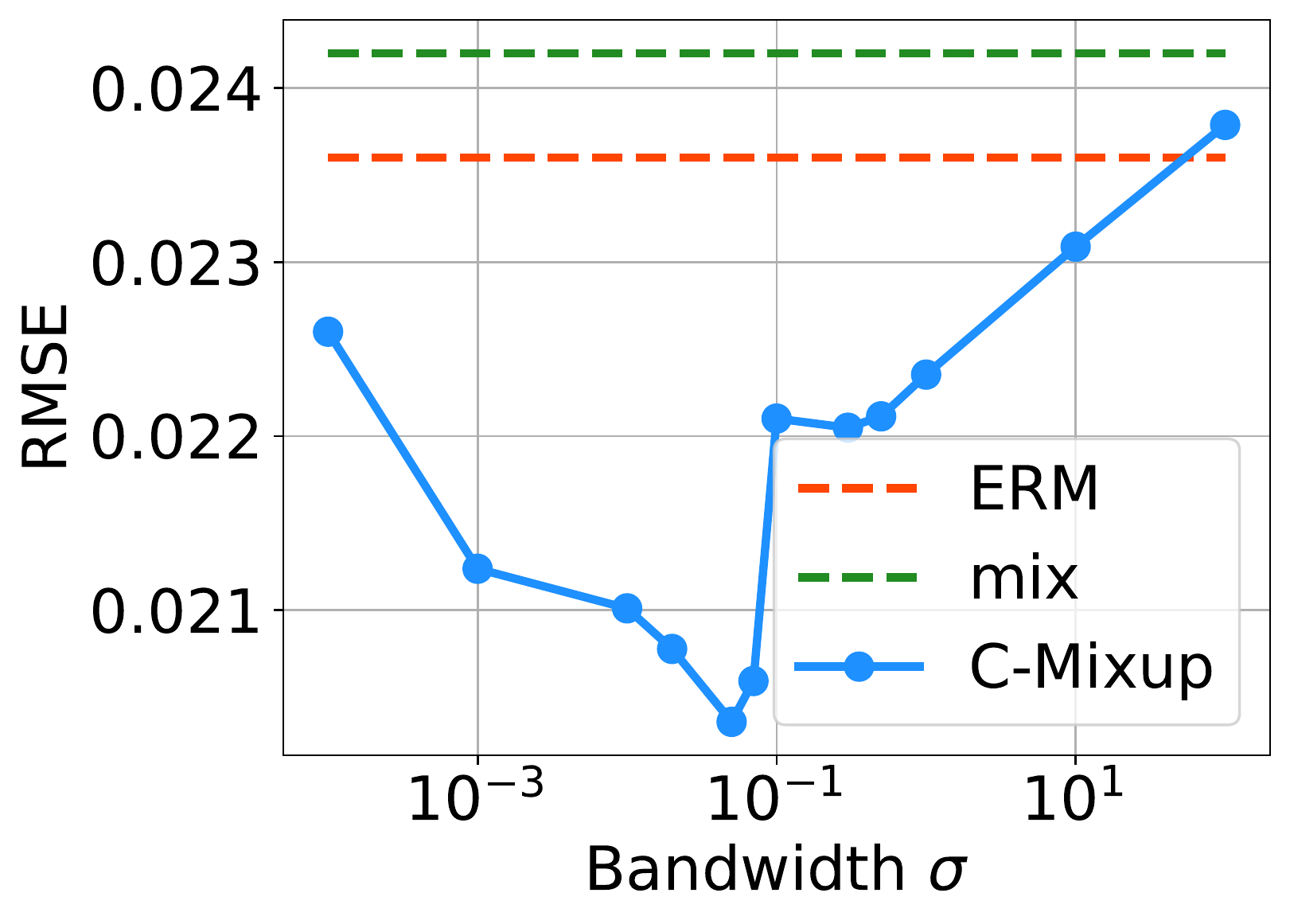}
    \caption{\label{fig:bandwidth_exchangerate}: Exchange-Rate}
\end{subfigure}
\begin{subfigure}[c]{0.225\textwidth}
		\centering
\includegraphics[width=\textwidth]{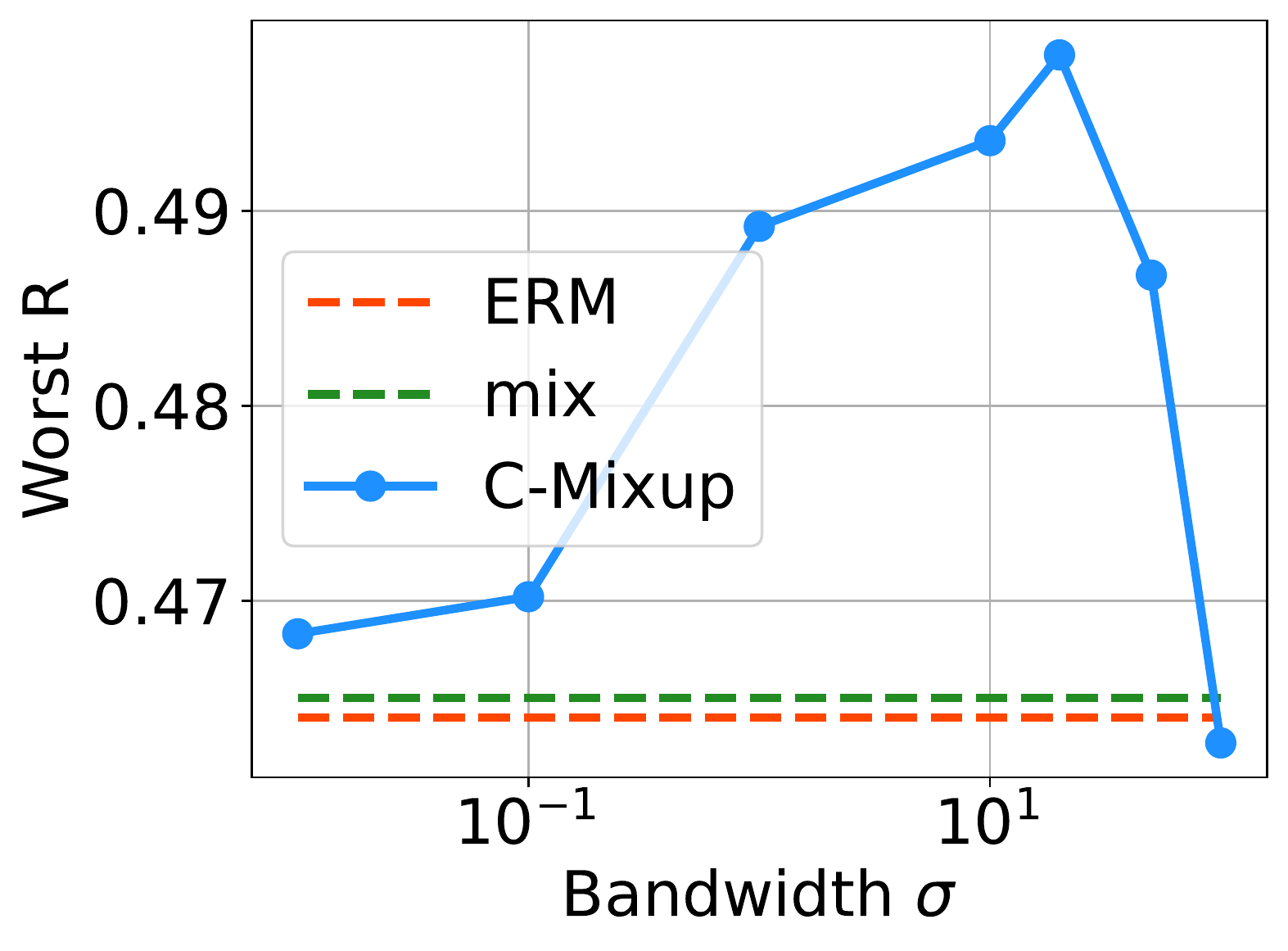}
    \caption{\label{fig:dti}: DTI}
\end{subfigure}
\vspace{-0.5em}
\caption{Sensitivity analysis of bandwidth. mix represents the better mixing approach between mixup and Manifold Mixup.}
\label{fig:bandwidth}
\vspace{-1em}
\end{wrapfigure}
\textbf{III. How does the choice of bandwidth affects the performance?} Finally, we analyze the effects of the bandwidth $\sigma$ in Eqn.~\eqref{eq:similarity}. The performance with respect to bandwidth of Exchange-Rate and DTI are visualized in Figure~\ref{fig:bandwidth}, respectively. We point out that if the bandwidth is too large, the results are close to vanilla mixup or Manifold Mixup, depending on which one is used. Otherwise, the results are close to ERM if the bandwidth is too small. According to the results, we see that the improvements of C-Mixup is somewhat stable over different bandwidths. Most importantly, C-Mixup yields a good model for a wide range of bandwidths, which reduces the efforts to tune the bandwidth for every specific dataset. Additionally, we provide empirical guidance about how to pick suitable bandwidth in Appendix~\ref{sec:app_bandwidth}.
\vspace{-0.5em}
\section{Related Work}
\vspace{-0.5em}
\textbf{Data Augmentation and Mixup.} Various data augmentation strategies have been proposed to improve the generalization of deep neural networks, including directly augmenting images with manually designed strategies (e.g., whitening, cropping)~\cite{krizhevsky2012imagenet}, generating more examples with generative models~\cite{antoniou2017data,bowles2018gan,zoph2020learning,zhou2022deep}, and automatically finding augmentation strategies~\cite{cubuk2019autoaugment,cubuk2020randaugment,lim2019fast}. Mixup~\cite{zhang2017mixup} and its variants~\cite{chen2022transmix,devries2017improved,greenewald2021k,hendrycks2019augmix,hong2021stylemix,kim2020puzzle,kim2021co,liu2022tokenmix,liu2021unveiling,park2022saliency,uddin2020saliencymix,venkataramanan2022alignmixup,verma2019manifold,yun2019cutmix,zhu2020automix} propose to improve generalization by linear interpolating input features of a pair of examples and their corresponding labels. Though mixup and its variants have demonstrated their power in classification~\cite{zhang2017mixup}, sequence labeling~\cite{zhang2020seqmix}, and reinforcement learning~\cite{wang2020improving}, systematic analysis of mixup on different regression tasks is still underexplored. The recent MixRL~\cite{hwang2021mixrl} learns a policy network to select nearby example pairs for mixing, which requires substantial computational resources and is not suitable to high-dimensional real-world data. Unlike this method, C-Mixup instead adjusts the sampling probability of mixing pairs based on label similarity, which makes it much more efficient. Furthermore, C-Mixup, which focuses on how to select mixing examples, is a complementary 
method over mixup and its representative variants (see more discussion in Appendix A.4). Empirically, our experiments show the effectiveness and compatibility of C-Mixup on multiple regression tasks in Section~\ref{sec:exp_iid}.

\textbf{Task Generalization.}
Our experiments extended C-Mixup to gradient-based meta-learning, aiming to improve task generalization. In the literature, there are two lines of related works. The first line of research directly imposes regularization on meta-learning algorithms~\cite{guiroy2019towards,jamal2019task,tseng2020regularizing,yin2020meta}. The second line of approaches introduces task augmentation to produce more tasks for meta-training, including imposing label noise~\cite{rajendran2020meta}, mixing support and query sets in the outer-loop optimization~\cite{ni2020data,yao2021improving}, and directly interpolating tasks to densify the entire task distribution~\cite{yao2022meta}. C-Mixup is complimentary to the latter mixup-based task augmentation methods. Furthermore, Section~\ref{sec:exp_task} indicates that C-Mixup empirically outperforms multiple representative prior approaches~\cite{yin2020meta,rajendran2020meta}.

\textbf{Out-of-Distribution Robustness.}
Many recent methods aim to build machine learning models that are robust to distribution shift, including learning invariant representations with domain alignment~\cite{ganin2016domain,li2018domain,li2022learning,long2015learning,tzeng2014deep,xu2020adversarial,yue2019domain,zhou2020deep} or using explicit regularizers to finding a invariant predictors that performs well over all domains~\cite{ahuja2021invariance,arjovsky2019invariant,guo2021out,khezeli2021invariance,koyama2020out,krueger2021out,yao2022wildtime}. Recently, LISA~\cite{yao2022improving} cancels out domain-associated correlations and learns invariant predictors by mixing examples either with the same label but different domains or with the same domain but different labels. While related, C-Mixup can be considered as a more general version of LISA that can be used for regression tasks. Besides, unlike LISA, C-Mixup do not use domain annotations, which are often expensive to obtain.
\vspace{-0.5em}
\section{Conclusion}
\vspace{-0.5em}
In this paper, we proposed C-Mixup, a simple yet effective variant of mixup that is well-suited to regression tasks in deep neural networks. Specifically, C-Mixup adjusts the sampling probability of mixing example pairs by assigning higher probability to pairs with closer label values. Both theoretical and empirical results demonstrate the promise of C-Mixup in improving in-distribution generalization, task generalization, and out-of-distribution robustness.

Theoretical future work may relax the assumptions in our theoretical analysis and extend the results to more complicated scenarios. We also plan to analyze more properties of C-Mixup theoretically, e.g., the relation between C-Mixup and manifold intrusion~\cite{guo2019mixup} in regression. Empirically, we plan to investigate how C-Mixup performs in more diverse application tasks such as semantic segmentation, natural language understanding, reinforcement learning.
\vspace{-0.5em}
\section*{Acknowledgement}
\vspace{-0.5em}
We thank Yoonho Lee, Pang Wei Koh, Zhen-Yu Zhang, and members of the IRIS lab for the many insightful discussions and helpful feedback. This research was funded in part by JPMorgan Chase \& Co. Any views or opinions expressed herein are
solely those of the authors listed, and may differ from the views and opinions expressed by JPMorgan Chase
\& Co. or its affiliates. This material is not a product of the Research Department of J.P. Morgan Securities
LLC. This material should not be construed as an individual recommendation for any particular client and is
not intended as a recommendation of particular securities, financial instruments or strategies for a particular
client. This material does not constitute a solicitation or offer in any jurisdiction. The research was also supported by Apple, Intel and Juniper Networks. CF is a CIFAR fellow. The research of Linjun Zhang is partially supported by NSF DMS-2015378.
\bibliographystyle{plainnat}
\bibliography{ref}

\section*{Checklist}
\begin{enumerate}

\item For all authors...
\begin{enumerate}
  \item Do the main claims made in the abstract and introduction accurately reflect the paper's contributions and scope?
    \answerYes{}
  \item Did you describe the limitations of your work?
    \answerYes{See Appendix G}
  \item Did you discuss any potential negative societal impacts of your work?
    \answerNA{
    C-Mixup is a general machine learning method and we believe that it has no potential negative societal impacts.
    }
  \item Have you read the ethics review guidelines and ensured that your paper conforms to them?
    \answerYes{}
\end{enumerate}

\item If you are including theoretical results...
\begin{enumerate}
  \item Did you state the full set of assumptions of all theoretical results?
    \answerYes{See Section 3 and Appendix B}
        \item Did you include complete proofs of all theoretical results?
    \answerYes{See Appendix B}
\end{enumerate}

\item If you ran experiments...
\begin{enumerate}
  \item Did you include the code, data, and instructions needed to reproduce the main experimental results (either in the supplemental material or as a URL)?
   \answerYes{See the code in \href{https://github.com/huaxiuyao/C-Mixup}{https://github.com/huaxiuyao/C-Mixup}}
  \item Did you specify all the training details (e.g., data splits, hyperparameters, how they were chosen)?
    \answerYes{See Appendix C.2, D.2, and E.2.}
  \item Did you report error bars (e.g., with respect to the random seed after running experiments multiple times)?
    \answerYes{See Appendix C, D, E, F.}
  \item Did you include the total amount of compute and the type of resources used (e.g., type of GPUs, internal cluster, or cloud provider)?
    \answerYes{See Appendix C.2, D.2, and E.2.}
\end{enumerate}

\item If you are using existing assets (e.g., code, data, models) or curating/releasing new assets...
\begin{enumerate}
  \item If your work uses existing assets, did you cite the creators?
    \answerYes{}
  \item Did you mention the license of the assets?
    \answerYes{}
  \item Did you include any new assets either in the supplemental material or as a URL?
    \answerYes{See the code in \href{https://github.com/huaxiuyao/C-Mixup}{https://github.com/huaxiuyao/C-Mixup}}
  \item Did you discuss whether and how consent was obtained from people whose data you're using/curating?
    \answerNA{}
  \item Did you discuss whether the data you are using/curating contains personally identifiable information or offensive content?
    \answerNA{}
\end{enumerate}

\item If you used crowdsourcing or conducted research with human subjects...
\begin{enumerate}
  \item Did you include the full text of instructions given to participants and screenshots, if applicable?
    \answerNA{}
  \item Did you describe any potential participant risks, with links to Institutional Review Board (IRB) approvals, if applicable?
    \answerNA{}
  \item Did you include the estimated hourly wage paid to participants and the total amount spent on participant compensation?
    \answerNA{}
\end{enumerate}

\end{enumerate}
\appendix

\section{Additional Information for C-Mixup}
\subsection{Illustration of How C-Mixup Improves Out-of-Distribution Robustness}
\label{app:a_1}
\begin{figure}[h]
\centering\includegraphics[width=\textwidth]{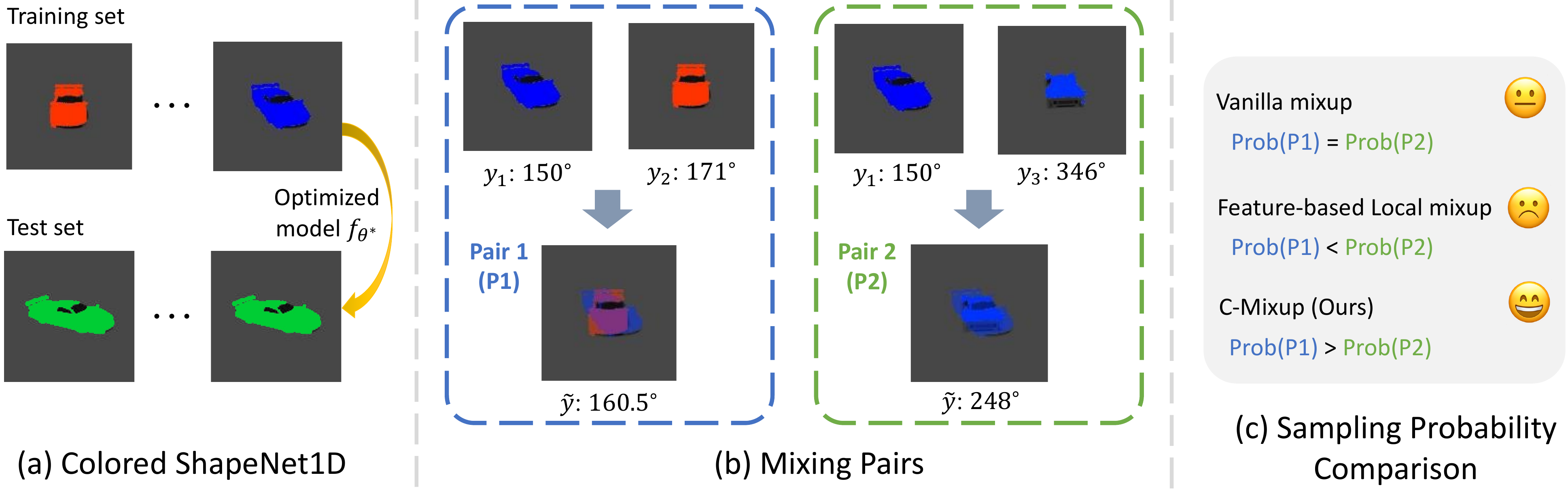}
\caption{Illustration of C-Mixup for out-of-distribution robustness. Here, we color the ShapeNet1D and regard color as the domain information. $\lambda$ represents the interpolation ratio. (a) Colored ShapeNet1D pose prediction task, aiming to generalize the model trained on red and blue domains to the green domain; (b) Two mixing pairs with interpolated images and labels; (c) Illustration of a rough comparison of sampling probabilities between two mixing pairs in (b). Here, C-Mixup is capable of assigning higher sampling probability to more reasonable pairs and eliminate the effect of domain information.}
\label{fig:ood_mixup}
\vspace{-1em}
\end{figure}

In Figure~\ref{fig:ood_mixup}, we use the ShapeNet1D to illustrate how C-Mixup improves out-of-distribution robustness. Here, we color the images to construct different domains. We train the model on red and blue domains and then generalize it to the green one. In Figure~\ref{fig:ood_mixup}, we can see that C-Mixup can recognize more reasonable mixing pairs compared with vanilla mixup. Mixup with feature similarity fails to cancel out the domain information since it may be easier to mix unreasonable example pairs within the same domain. C-Mixup instead is naturally suitable to average out domain information by mixing examples with close labels.
\subsection{Algorithm of Meta-Training with C-Mixup}
\label{app:a_2}
In this section, we summarize the algorithm of applying C-Mixup to MetaMix~\cite{yao2021improving} in Alg.~\ref{alg:C-Mixup_metamix}. Here, we adopt MetaMix with mixup version, which could be easily adapted to other mixup variants (e.g., CutMix, Manifold Mixup)
\begin{algorithm}[t]
\caption{Meta-Training Process of MetaMix with C-Mixup}
\label{alg:C-Mixup_metamix}
\begin{algorithmic}[1]
\REQUIRE Outer-loop learning rate $\eta$; Inner-loop learning rate $\xi$ (we change $\alpha$ in Eqn. (4) of the main paper to $\xi$ to avoid notation conflict); Shape parameter $\alpha$; Task distribution $p(\mathcal{T})$
\STATE Randomly initialize model parameters $\theta$
\WHILE{not converge}
\STATE Sample a batch of tasks $\{\mathcal{T}_i\}_{i=1}^{|M|}$ with the corresponding dataset $\mathcal{D}_m$
\FOR{all $\mathcal{T}_m$}
\STATE Sample a support set $\mathcal{D}_m^s$ and a query set $\mathcal{D}_m^q$ from $\mathcal{D}_m$
\STATE Calculate pairwise distance matrix $P$ between query set and support set via Eqn. (6).
\STATE Calculate the task-specific parameter $\phi_m$ via the inner-loop gradient descent, i.e., $\phi_m=\theta-\xi \nabla_{\theta}\mathcal{L}(f_{\theta};\mathcal{D}_m^s)$
\FOR{each query example ($x_{m,i}^q, y_{m,i}^q$)}
\STATE Sample MetaMix parameter $\lambda \sim \mathrm{Beta}(\alpha, \alpha)$
\STATE Sample support set example ($x_{m,j}^s$, $y_{m,j}^s$) according to the probability $P(\cdot \mid (x_{m,i}^q, y_{m,i}^q))$
\STATE Linearly interpolate ($x_{m,i}^q$, $y_{m,i}^q$) and ($x_{m,j}^s$, $y_{m,j}^s$) to get ($\Tilde{x}_{m,i}^q, \Tilde{y}_{m,i}^q$)
\STATE Replace ($x_{m,i}^q, y_{m,i}^q$) with ($\Tilde{x}_{m,i}^q, \Tilde{y}_{m,i}^q$) 
\ENDFOR
\ENDFOR
\STATE Use interpolated examples to update the model via $\theta \leftarrow \theta - \eta\frac{1}{|M|} \sum_{i=1}^{|M|} \mathcal{L}(f_{\phi_m};\Tilde{\mathcal{D}}_m^q)$
\ENDWHILE
\end{algorithmic}
\end{algorithm}
\subsection{Efficiency Discussion of C-Mixup}
\label{app:a_3}
Assume the number of examples are $n$, the dimension of features and labels are $M_f$ and $M_l$, respectively. The time complexity of calculating the pairwise distance matrix $P$ with feature distance or label distance is $O(n^2M_f)$ or $O(n^2M_l)$, respectively. Generally, since $M_l\ll M_f$, using label distance (i.e., C-Mixup) substantially reduces the cost of calculating the pairwise distance matrix.

\begin{algorithm}[t]
\caption{Training with C-Mixup-batch}
\label{alg:C-Mixup_batch}
\begin{algorithmic}[1]
\REQUIRE Learning rates $\eta$; Shape parameter $\alpha$
\REQUIRE Training data $\mathcal{D}:=\{(x_i, y_i)\}_{i=1}^N$
\STATE Randomly initialize model parameters $\theta$
\WHILE{not converge}
\STATE Sample two batches of examples $\mathcal{B}_1, \mathcal{B}_2\sim \mathcal{D}$
\STATE Calculate pairwise distance matrix $P$ between $\mathcal{B}_1$ and $\mathcal{B}_2$ via Eqn. (6)
\FOR{each example $(x_i, y_i) \in \mathcal{B}_1$}
\STATE Sample example ($x_j$, $y_j$) from $\mathcal{B}_2$ according to the probability $P(\cdot \mid (x_i, y_i))$
\STATE Sample $\lambda$ from $\mathrm{Beta}(\alpha, \alpha)$
\STATE Linearly interpolate ($x_i$, $y_i$) and ($x_j$, $y_j$) to get ($\Tilde{x}, \Tilde{y}$)
\ENDFOR
\STATE Use interpolated examples to update the model via Eqn. (3)
\ENDWHILE
\end{algorithmic}
\end{algorithm}
Furthermore, the calculation of pairwise distance matrix can be accelerated using parallelized operations, but it is still challenging if $n$ is sufficiently large, e.g., billions of examples. We thus propose an alternative solution that applies C-Mixup only to every example batch, which is named as \textbf{C-Mixup-batch}. In Alg.~\ref{alg:C-Mixup_batch}, we summarize the training process of C-Mixup-batch. We compare C-Mixup and C-Mixup-batch, and report the results of in-distribution generalization and out-of-distribution robustness in Table~\ref{tab:batch_id} and Table~\ref{tab:batch_ood}, respectively. Notice that we have used C-Mixup-batch for Echo, RCF-MNIST, ProvertyMap since calculating pair-wise distance metrics for these large datasets is time-consuming. Hence, we only report the results for other datasets. According to the results, we observe that C-Mixup-batch achieves comparable performance to C-Mixup. Nevertheless, the downside of C-Mixup-batch is that we must calculate the pairwise distance matrix for every batch. Accordingly, original C-Mixup is suitable to most datasets, while C-Mixup-batch is more appropriate to large datasets (e.g., Echo).

\begin{table}[h]
\small
\caption{Comparison between C-Mixup and C-Mixup-batch to in-distribution generalization. Since Echo is a large dataset and C-Mixup-batch is used by default, we only report the results of Airfoil, NO2, Exchange-Rate and Electricity here.}
\label{tab:batch_id}
\begin{center}
\setlength{\tabcolsep}{1.5mm}{
\begin{tabular}{l|l|c|c|c|c}
\toprule
& Dataset & Airfoil & NO2 & Exchange-Rate & Electricity \\\midrule
\multirow{2}{*}{RMSE $\downarrow$} & C-Mixup-batch & 2.792 $\pm$ 0.135 & 0.510 $\pm$ 0.007 & 0.0205 $\pm$ 0.0017 & 0.0576 $\pm$ 0.0002 \\
& C-Mixup & 2.717 $\pm$ 0.067 & 0.509 $\pm$ 0.006 & 0.0203 $\pm$ 0.0011 & 0.0570 $\pm$ 0.0006 \\\midrule\midrule

\multirow{2}{*}{MAPE $\downarrow$} & C-Mixup-batch & 1.616 $\pm$ 0.053\% & 12.894 $\pm$ 0.180\% & 2.064 $\pm$ 0.218\% & 13.697 $\pm$ 0.155\% \\
& C-Mixup & 1.610 $\pm$ 0.085\% & 12.998 $\pm$ 0.271\% & 2.041 $\pm$ 0.134\% & 13.372 $\pm$ 0.106\% \\
\bottomrule
\end{tabular}}
\end{center}
\vspace{-1em}
\end{table}

\begin{table}[h]
\small
\caption{Comparison between C-Mixup and C-Mixup-batch to out-of-distribution robustness. Since we have applied C-Mixup-batch to PovertyMap, we report the results of Crime, SkillCraft, DTI.}
\label{tab:batch_ood}
\begin{center}
\setlength{\tabcolsep}{1.5mm}{
\begin{tabular}{l|l|c|c|c}
\toprule
& Dataset & Crime (RMSE $\downarrow$) & SkillCraft (RMSE $\downarrow$) & DTI ($R \uparrow$) \\\midrule
\multirow{2}{*}{Avg} & C-Mixup-batch & 0.125 $\pm$ 0.001 & 5.619 $\pm$ 0.212 & 0.490 $\pm$ 0.005 \\
& C-Mixup & 0.123 $\pm$ 0.000 & 5.201 $\pm$ 0.059 & 0.498 $\pm$ 0.008 \\\midrule\midrule

\multirow{2}{*}{Worst} & C-Mixup-batch & 0.152 $\pm$ 0.007 & 7.665 $\pm$ 0.875 & 0.453 $\pm$ 0.006 \\
& C-Mixup & 0.146 $\pm$ 0.002 & 7.362 $\pm$ 0.244 & 0.458 $\pm$ 0.004  \\
\bottomrule
\end{tabular}}
\end{center}
\end{table}

\subsection{Discussion between C-Mixup and Mixup}
\label{app:mixup_detail}
In this paper, we regard C-Mixup is an complementary approach to mixup and its most representative variants (e.g., Manifold Mixup~\cite{verma2019manifold}, CutMix~\cite{yun2019cutmix}). Here, we use vanilla mixup as an exemplar to show the difference. According to our description of mixup in Section 2 of the main paper, the entire mixup process includes three stages:
\begin{itemize}[leftmargin=*]
    \item Stage I: sample two instances ($x_i$, $y_i$), ($x_j$, $y_j$) from the training set.
    \item Stage II: sample the interpolation factor $\lambda$ from the Beta distribution Beta($\alpha$, $\alpha$).
    \item Stage III: mixing the sampled instances with interpolation factor $\lambda$ according to the following mixing formulation:
    \begin{equation}
    \nonumber
    x_{mix}=\lambda x_i + (1-\lambda) x_j, y_{mix}=\lambda y_i + (1-\lambda) y_j, \lambda \sim \mathrm{Beta}(\alpha, \alpha).
    \end{equation}
\end{itemize}

In the original mixup, the interpolation factor $\lambda$ sampled in the stage II controls how to mix these two instances. C-Mixup instead manipulates stage I and pairs with closer labels are more likely to be sampled.

In addition to the discussion of the complementarity of C-Mixup, the original mixup paper further shows that randomly interpolating examples from the same label performs worse than completely random mixing examples in classification. Compared to classification, randomly mixing examples in regression may be easier to generate semantically wrong labels. Intuitively, linearly mixing one-hot labels in classification is easy to generate semantically meaningful artificial labels, where the mixed label represents the probabilities of mixed examples to some extent. While in regression, the mixed labels may be semantically meaningless (e.g., pairs 2 and 3 in Figure~\ref{fig:iid_mixup}) and more significantly affect the performance. By mixing examples with closer labels, C-Mixup mitigates the influence of semantically wrong labels and improves the in-distribution and task generalization in regression. Additionally, C-Mixup further shows its superiority in improving out-of-distribution robustness in regression, which is not discussed in the original mixup paper.

\section{Detailed Proofs}
In this section, we provide detailed proofs of Theorem 1, 2, 3. To avoid symbol conflict, we would like to point out that we use $h$ to denote the bandwidth of kernel in the nonparametric estimation step, which is different from the bandwidth $\sigma$ in Eqn. (6) of the main paper, which was used to measure the similarity in mixup. In Section~\ref{pf:lem}, we provide proofs for all used Lemmas.
\subsection{Proof of Theorem 1}
\label{app:b_1}
We first state the kernel estimator. $$
\hat g(t;\theta)=\frac{\sum_{i=1}^nK(\theta^\top x_i,t)y_i}{\sum_{i=1}^nK(\theta^\top x_i,t)},
$$
this kernel function can take, for example, the uniform kernel function $K(t_1,t_2)=1\{|t_1-t_2|<h\}$ or a Gaussian kernel function $K(t_1,t_2)=\exp(-|t_1-t_2|^2/h^2)$ where $h$ is the bandwidth.

To make the proof easier to follow, we restate Theorem 1 below.
\begin{lemma_theorem}
Suppose $\theta\in\R^p$ is sparse with sparsity $s=o(\min\{p,\sigma_{\xi}^2\})$, $p=o(N)$ and $g$ is smooth with $c_0<g'<c_1$, $c_2<g''<c_3$ for some universal constants $c_0, c_1,c_2, c_3>0$. 
There exists a distribution on $x$ with a kernel function, such that when the sample size $N$ is sufficiently large, with probability $1-o(1)$, 
\begin{equation}
\small
    \mathrm{MSE}(\theta^{*}_{C-Mixup})<\min(\mathrm{MSE}(\theta^{*}_{feat}),\mathrm{MSE}(\theta^{*}_{mixup})).
\end{equation}
\end{lemma_theorem}

\textbf{Proof.} For identifiability, we consider the case where $\theta$ has $\ell_2$ norm of 1.  Let us construct the distribution of $x$ as $z\sim\frac{1}{K}\sum_{k=1}^K N_p(\mu_k, \sigma_z^2 I_p)$ with $\mu_k= \frac{k}{\|\theta\|}\theta$ for a fixed positive integer $K$. We break out the entire proof into three steps.

\textbf{Step 1.} We first analyze the behavior of the three mixup methods. First, we have $$
\theta^\top z\sim\frac{1}{K}\sum_{k=1}^K N_p(\mu_k^\top\theta, \sigma_z^2\|\theta\|^2),
$$
and $$
\theta^\top \xi\sim\frac{1}{K}\sum_{k=1}^K N_p(0, \sigma_{\xi}^2\|\theta\|^2).
$$
We then have $$
|y_{k'}-y_k|=|g(\mu_k^\top\theta)-g(\mu_{k'}^\top\theta)|\pm \Theta((\sigma_{\xi}+\sigma_{x})\|\theta\|)=|g(k)-g(k')|\pm \Theta((\sigma_{\xi}+\sigma_{x})\|\theta\|)
$$
$$
\|x_{k'}-x_k\|=\|\mu_k-\mu_k'\|\pm \Theta((\sigma_{\xi}+\sigma_{x})\cdot\sqrt{d})=|k-k'|\pm \Theta((\sigma_{\xi}+\sigma_{x})\cdot\sqrt{d})
$$
As a result, if we take $(\sigma_{\xi}+\sigma_x)\|\theta\|=o(1)$, and $(\sigma_{\xi}+\sigma_{x})\cdot\sqrt{d}\to\infty$, C-Mixup only interpolates examples within the same cluster, while mixup with feature similarity and vanilla mixup interpolate examples across different clusters.  

\textbf{Step 2.} We then show that $\hat\theta$ obtained by all the three methods is consistent (the estimation error goes to 0 when the sample size goes to infinity). 
We first present a lemma showing that in expectation, the solution would recover $\theta$.

\begin{lemma}\label{lem:sigmoid}
Suppose $x_i$'s are i.i.d. sampled from $N_p(\mu,I)$, then we have
$$\E[y_ix_i]=
(\E[g'(x_i^\top\theta)]+\E[g(x_i^\top\theta)])\cdot \theta.$$
\end{lemma} 
The proof of Lemma~\ref{lem:sigmoid} is deferred to Section~\ref{pf:lem}.

This lemma implies that if $\tilde x=\lambda x_k+(1-\lambda)x_k'$, $\tilde y=\lambda y_k+(1-\lambda)y_k'$, we have $$
\E[\tilde y\tilde x]=c_{k,k'}\theta,
$$
for some constant $c_{k,k'}$. Additionally, we have $\E[\tilde x\tilde x^\top]=cI+c'_{k,k'}\theta\theta^\top$. Therefore, $\E[\tilde x\tilde x^\top]^{-1}\E[\tilde y\tilde x]=\tilde c_{k,k'}\theta$ (via the Sherman–Morrison formula), for some constant $\tilde c_{k,k'}$.

Since we assume $g$ is  $c_1$-Lipschitz for some universal constant $c_1$, which also implies $\tilde c_{k,k'}=O(1)$. We then analyze the convergence of $\hat \theta$. By definition, we have $$
\hat \theta=(\frac{1}{N}\sum_{i=1}^{N}\tilde x_i\tilde x_i^\top)^{-1}(\frac{1}{N}\sum_{i=1}^{N}\tilde x_i\tilde y_i).
$$
Using Bernstein inequality, we have with probability at least $1-p^{-2}$, $$
\|\frac{1}{N}\sum_{i=1}^{N}\tilde x_i \tilde x_i^\top-\E[\tilde x\tilde x^\top]\|=O(\sqrt\frac{p}{N}),
$$
and $$
\|\frac{1}{N}\sum_{i=1}^{N} \tilde x_i\tilde y_i-\E[\tilde x\tilde y]\|=O(\sqrt\frac{p}{N}).
$$
Then using Lemma 1, since $\lambda_{\min}(\E[\tilde x\tilde x^\top])\gtrsim c_1$, when $n$ is sufficiently large, we then have $$
\|\hat{\theta}-\theta\|_2\lesssim\sqrt{\frac{p}{N}}=o(1).
$$

\textbf{Step 3.} We finally proceed to the nonparametric estimation step. 

For C-Mixup, since we only interpolates the samples within the same Gaussian cluster, using the fact that $\sigma_\xi=o(1)$, $K$ being Lipschitz, and $\|\hat{\theta}-\theta\|_2=o(1)$, we have that
 \begin{equation}\label{eq1}
\|\hat g(t;\hat\theta)-\frac{\sum_{i=1}^NK(z_i^\top\theta,t)y_i}{\sum_{i=1}^NK(z_i^\top\theta,t)}\|=\|\frac{\sum_{i=1}^NK(x_i^\top\hat\theta,t)y_i}{\sum_{i=1}^NK(x_i^\top\hat\theta,t)}-\frac{\sum_{i=1}^NK(z_i^\top\theta,t)y_i}{\sum_{i=1}^NK(z_i^\top\theta,t)}\|=o(1),
\end{equation}
here the function norm of $h$ is defined as $\|h\|=\sqrt{\E[h^2(x)]}.$

Using the standard nonparametric regression results (e.g., see \citet{tsybakov2004introduction}), when the feature is observed without noise, the kernel estimator is consistent: 
 \begin{equation}\label{eq2}
\|g(t)-\frac{\sum_{i=1}^NK(z_i^\top\theta,t)y_i}{\sum_{i=1}^NK(z_i^\top\theta,t)}\|=o(1).
\end{equation}

Combining the two inequalities \eqref{eq1} and \eqref{eq2}, we find that the $\hat g$ obtained by C-Mixup satisfies $$
\|\hat g-g\|=o(1).
$$

For vanilla mixup and mixup with feature similarity, we have show that with a nontrivial positive probability, the samples are from two different clusters. Therefore using the assumption on $g'$ and $g''$, we have $\tilde y-y_i\ge c$ for some constant $c>0$ with Jensen's inequality. As a result, 
 $$
|\frac{\sum_{i=1}^NK(x_i^\top\hat\theta,t)y_i}{\sum_{i=1}^NK(x_i^\top\hat\theta,t)}-\frac{\sum_{i=1}^NK(x_i^\top\hat\theta,t)\tilde y_i}{\sum_{i=1}^NK(x_i^\top\hat\theta,t)}|\ge c.
$$
Combining with the two inequalities \eqref{eq1} and \eqref{eq2}, we have the $\hat g$ obtained by vanilla mixup or mixup with feature similarity satisfies $$
\|\hat g-g\|>c.
$$
Since  $\mathrm{MSE}(\theta)\asymp\|\hat g(\cdot;\theta)-g(\cdot)\|$, we then have
\begin{equation*}
\small
    \mathrm{MSE}(\theta^{*}_{C-Mixup})<\min(\mathrm{MSE}(\theta^{*}_{feat}),\mathrm{MSE}(\theta^{*}_{mixup})).
\end{equation*}
\subsection{Proof of Theorem 2}
\label{app:b_2}
We first restate Theorem 2.
\begin{lemma_theorem}
Let $N=\sum_{m=1}^M N_m$ and $N_m$ is the number of examples of $\mathcal{T}_m$. Suppose $\theta_k$ is sparse with sparsity $s=o(\min\{d,\sigma_{\xi}^2\})$, $p=o(N)$ and $g_m$'s are smooth with $0<g_m'<c_1$, $c_2<g_m''<c_3$ for some universal constants $c_1,c_2, c_3>0$ and $m\in[M]\cup\{t\}$.  There exists a distribution on $x$ with a kernel function, such that when the sample size $N$ is sufficiently large, with probability $1-o(1)$, 
\begin{equation}
\small
    \mathrm{MSE}_{\mathrm{Target}}(\theta^{*}_{Meta-C-Mixup})<\min(\mathrm{MSE}_{\mathrm{Target}}(\theta^{*}_{Meta-feat}),\mathrm{MSE}_{\mathrm{Target}}(\theta^{*}_{MetaMix})).
\end{equation}
\end{lemma_theorem}

Again, we consider the distribution of $x$ to be $z\sim\frac{1}{K}\sum_{k=1}^K N_p(\mu_k, \sigma_z^2 I_p)$ with $\mu_k= \frac{k}{\|\theta\|}\theta$ for a fixed positive integer $K$. The proof of Theorem 2 largely follows Theorem 1, with the only difference in the step 2. We prove the step 2 for Theorem 2 in the following.

By Lemma~\ref{lem:sigmoid},  
 for augmented data in the $m$-th task, $\tilde x^{(m)}=\lambda x_k^{(m)}+(1-\lambda)x_k^{(m)'}$,
 $\tilde y^{(m)}=\lambda y_k^{(m)}+(1-\lambda)y^{(m)'}_k$, we have $$
\E[\tilde y^{(m)}\tilde x^{(m)}]=c^{(m)}_{k,k'}\theta,
$$
for some constant $c^{(m)}_{k,k'}$.

Additionally, we have $\E[\tilde x^{(m)}\tilde x^{(m)\top}]=c^{(m)}I+c^{(m)'}_{k,k'}\theta\theta^\top$. Therefore, $(\sum_{m=1}^T\E[\tilde x^{(m)}\tilde x^{(m)\top}])^{-1}(\sum_{m=1}^T\E[\tilde y^{(m)}\tilde x^{(m)}])=\tilde c_{k,k'}\theta,$ for some constant $\tilde c_{k,k'}$.


We then analyze the convergence of $\hat \theta$. By definition, we have $$
\hat \theta=(\frac{1}{N}\sum_{m=1}^T\frac{1}{n_m}\sum_{i=1}^{n_t}\tilde x^{(m)}_i\tilde x_i^{(m)\top})^{-1}(\frac{1}{N}\sum_{m=1}^T\frac{1}{n_m}\sum_{i=1}^{n_t}\tilde x^{(m)}_i\tilde y^{(m)}_i).
$$
Using Bernstein inequality, we have with probability at least $1-p^{-2}$, $$
\|\frac{1}{N}\sum_{m=1}^T\frac{1}{n_m}\sum_{i=1}^{n_m}x_i^{(m)}x_i^{(m)\top}-\frac{1}{N}\sum_{m=1}^T\E[\tilde x^{(m)} \tilde x^{(m)\top}]\|=O(\sqrt\frac{p}{N}),
$$
and $$
\|\frac{1}{T}\sum_{m=1}^T\frac{1}{n_m}\sum_{i=1}^{n_m}x_i^{(m)}y_i^{(m)}-\E[\frac{1}{T}\sum_{t=1}^T\E[x^{(m)}y^{(m)}]]\|=O(\sqrt\frac{p}{N }).
$$
Then using Lemma 1, when $N$ is sufficiently large, we then have $$
\|\hat{\theta}-\theta\|_2\lesssim\sqrt{\frac{p}{ N}}=o(1).
$$

\subsection{Proof of Theorem 3}
\label{app:b_3}
Similarly we restate Theorem 3.
\begin{lemma_theorem}
\label{app:theorem3_4}

Supposed for some \begin{small}$\max(\exp(-n^{1-o(1)}), \exp(-\frac{p_1^2}{2n})) < \delta \ll 1$\end{small}, we have variance constraints:\begin{small} $\sigma_a = c_1 \sigma_x$ , $\sigma_x \geq c_2 \max(\frac{n^{5/2}}{\Vert{\theta}\Vert\delta}\sigma_{\epsilon}, \frac{\sqrt{p_2}\Vert{\theta}\Vert}{\sqrt{n}p_1})$\end{small} and \begin{small}
$\sigma_{\epsilon}^2  \leq \frac{c_3}{pn^{3/2}}$
\end{small}. Then for any penalty k satisfies \begin{small}$c_4\sqrt{\frac{p_2}{p_1}}n^{1/4+o(1)} < k < c_5 \min(\frac{\sigma_x}{\Vert{\theta}\Vert}\sqrt{p_1n^{1-o(1)}}, n)$\end{small} and bandwidth h satisfies \begin{small}$0 < h \leq c_6 \frac{l}{\sqrt{\log(n^2/p_1)}} $\end{small} in C-Mixup, when $n$ is sufficiently large, with probability at least $1-o(1)$, we have
\begin{equation}
\small
\mathrm{MSE}(\theta^{*}_{C-Mixup})< \min(\mathrm{MSE}(\theta^{*}_{feat}), \mathrm{MSE}(\theta^{*}_{mixup})),
\end{equation}
where $c_1 \geq 1$, $c_2$, $c_3$, $c_4$, $c_5$, $c_6$ > 0 are  universal constants, \begin{small}$l = \min_{i\neq j}\lvert y_i - y_j^{\prime}\rvert$\end{small} and 
\begin{small}$p_1 \ll n < p_1^2$\end{small}.
\end{lemma_theorem}


Let $p = p_1 + p_2$, and $\theta_1$ represents the subvectors that contain the first $p_1$ coordinates of $\theta$. Furthermore, let $\hat X\in\mathbb{R}^{n \times p}$ be an arbitrary noise-less data matrix and $\hat\lambda_0 \geq \hat\lambda_1 \geq ... \geq \hat\lambda_p$ be the singular values of $\hat X$. Similarly, let $E\in\mathbb{R}^{n \times p}$ be the noise matrix which contains $iid.$ sub-Gaussian entries with variance proxy $\sigma_{\epsilon}^2$, and $\lambda_0 \geq \lambda_1 \geq ... \geq \lambda_p$ be the singular values of input matrix $X = \hat X+E$. 

Firstly we show that the noise matrix only makes a small difference between the singular values of noise-less data matrix $\hat X$ and that of input matrix $X$. 
\begin{lemma}
\label{B3_4_lemmaweyl}
If $e^{-n} < \delta_1 \ll 1$ and $\sigma_{\epsilon}^2  \leq \frac{c}{pn^{3/2}}$, then with probability at least $ 1 - \delta_1$ we have:
\begin{equation}
    \label{eq:weyl}
    \lvert \hat\lambda_u - \lambda_u \rvert \leq 16c n^{-1/2},
\end{equation}
for every u satisfies $1 \leq u \leq  p $.
\end{lemma}

Before analyzing the effectiveness of C-Mixup, we first present the following lemma to analyze the C-Mixup with truncated label distance measurements. Specifically, we only apply C-Mixup to examples within a label distance threshold.

\begin{lemma}
\label{threshold_mix}
Assume \begin{small}$\exp(-n^{1-o(1)}) < \delta_1 \ll 1$\end{small}, \begin{small}$\sigma_x \geq c_2 \frac{n^{5/2}}{\Vert{\theta}\Vert\delta}\sigma_{\epsilon}$\end{small} for some $c_2$ that satisfies \begin{small}$ c_{gap} := \frac{c_2\sqrt{\pi}}{4\sqrt{2+\lvert \theta_1\rvert^2}} > 1$\end{small}. Here $c_{gap}$ is the ratio of \begin{small}$\min_{i\neq j}\lvert y_i - y_j^{\prime}\rvert$\end{small} to \begin{small}$\max\limits_{i}\lvert y_i - y_i^{\prime}\rvert$\end{small}. Then if we use C-Mixup, there exists some thresholds such that with probability at least $1-\delta_1$, the training data $(x_i, y_i)$ will only be mixed with $(x^{\prime}_i, y^{\prime}_i)$. Here, we point out that $x_i$ and $x_i^{\prime}$ are defined in Line 206-207 in the main paper.
\end{lemma}

Then we consider replacing the truncated kernel with the gaussian kernel, which applies C-Mixup to all the examples with a smoother probability distribution. And we claim that there exist some bandwidths such that the data pairs with almost identical invariant features and opposite
domain-changeable features will be mixed up together with high probability.
\begin{lemma}
\label{lemma_bandwidth}
Assume \begin{small}$\exp(-\frac{p_1^2}{2n}) < \delta_1 \ll 1$\end{small} and for \begin{small}$l = \min_{i\neq j}|y_i - y_j^{\prime}|$\end{small}, we have \begin{small}$0 < h \leq c_6 \frac{l}{\sqrt{\log(n^2/p_1)}}$\end{small} for some \begin{small}$c_6: 0 < c_6 \leq c_{gap}^{-1}\sqrt{(c_{gap}^2-1)/4}$\end{small}, where $c_{gap}$ follows the definition in Lemma~\ref{threshold_mix}. We define the mixed input as $\Tilde{X}$, and let $S_i$ be a random variable that denotes if $(x_i, y_i)$ is mixed with $(x_i^{\prime}, y_i^{\prime})$. Then with probability at least $1-\delta_1$, we have:
$$
    n - \frac{p_1}{2} \leq \sum_{i=1}^n S_i \leq n.
$$

\end{lemma}

Next, we find that the input matrix $X$ corresponding to C-Mixup will have just $p_1$, rather than $p$, singular values that are much bigger than zero.
\begin{lemma}
\label{B3_4_lemma_lambda}
Assume the conditions of Lemma~\ref{B3_4_lemmaweyl} and Lemma~\ref{lemma_bandwidth} still hold and the mixup ratio $\lambda = 0.5$. If noise-less data matrix $\hat X$ is obtained by C-Mixup, then the singular values of input matrix $X$ satisfy:
\begin{equation}
    \label{singular_y}
    \begin{split}
    (1-b(n))^2n \leq &\lambda_i \leq (1+b(n))^2n, \qquad 1 \leq i \leq p_1 ,\\
    \lvert &\lambda_i \rvert \leq 16\frac{c}{\sqrt{n}}, \ \qquad\qquad p_1 \leq i \leq p.
    \end{split}
\end{equation}
\end{lemma}

Finally, we can complete the proof of Theorem~\ref{app:theorem3_4} according to the lemmas above.

\textbf{Proof of Theorem 3.}
Denote the input matrix as $X\in\mathbb{R}^{n \times p}$ and its singular values as $\lambda_0 \geq \lambda_1 \geq ... \geq \lambda_p$. Then, for ridge estimator with penalty $k$, we have~\citet{Hoerl2017ridge} :

\begin{equation}
\begin{split}
\nonumber
    \mathbb{E}[\lvert \theta^{*}(k) - \theta \rvert^2] &= \mathbb{E}[\lvert (X^TX+kI)^{-1}X^TY - \theta \rvert ^2] \\
    &= \sigma_{x}^2 \sum_{i=1}^p \frac{\lambda_i}{(\lambda_i+k)^2} + k^2\theta^T(X^TX+kI)^{-2}\theta\\
    &= \sigma_{x}^2 \sum_{i=1}^{p_1} \frac{\lambda_i}{(\lambda_i+k)^2} + \sigma_{x}^2 \sum_{i=p_1+1}^{p_2} \frac{\lambda_i}{(\lambda_i+k)^2} + k^2\theta^T(X^TX+kI)^{-2}\theta\\
    &= \gamma_1(k) + \gamma_2(k) + \gamma_3(k)
\end{split}
\end{equation}

For the first term, we have
\begin{equation}
\begin{split}
\nonumber
\gamma_1^{C-Mixup}(k) &= \sigma_{x}^2 \sum_{i=1}^{p_1} \frac{\lambda_i}{(\lambda_i+k)^2}\\
&\leq \sigma_{x}^2 \sum_{i=1}^{p_1} \frac{1}{(\min_{1\leq i \leq p_1}\lambda_i)}\\
&\leq \frac{\sigma_{x}^2p_1}{(1-b(n))^2n}, \qquad (\mathrm{Lemma}~\ref{B3_4_lemma_lambda})
\end{split}
\end{equation}

and 

$$
\gamma_1^{mixup}(k)\wedge \gamma_1^{feat}(k)\wedge \gamma_1^{C-Mixup}(k)
\geq \sigma_{x}^2 \sum_{i=1}^{p_1} \frac{1}{4k} \geq \frac{\sigma_x^2p_1}{4c_5n}. \qquad(k < c_5n, c_5 \geq \frac{1}{4})
$$

To bound the last term, we perform orthogonal decompose on $X^TX$, i.e., $X^TX = P^T\Lambda P$. $P$ is orthogonal transformation and we denote $\alpha = P\theta$. Since the last $p_2$ coordinates of $\theta$ are $0$, with probability at least $1-\delta_1$, for C-Mixup we have:

\begin{align*}
    \gamma_3^{C-Mixup}(k) &= k^2\theta^T(X^TX+kI)^{-2}\theta\\
    &= k^2\sum_{i=1}^{p}\frac{\alpha_i^2}{(\lambda_i+k)^2}\\
    &= \sum_{i=1}^{p_1}\frac{\alpha_i^2}{(\lambda_i/k+1)^2}\\
    & \leq \frac{k^2 \lvert \theta \rvert^2}{(\min_{1\leq i \leq p_1}\lambda_i)^2} \qquad (\sum_{i=1}^{p_1}\alpha_i^2 = \lvert \theta \rvert^2)\\
    & \leq \frac{c_5^2}{ (1-b(n))^4n^{o(1)}}\frac{\sigma_x^2p_1}{n} \qquad (k < c_5 \frac{\sigma_x}{\Vert{\theta}\Vert}\sqrt{p_1n^{1-o(1)}}, \ \mathrm{Lemma}~\ref{B3_4_lemma_lambda}).\\
\end{align*}
For the second term, we bound C-Mixup as
\begin{equation}
\label{eq_gamma2_y}
\begin{split}
\gamma_2^{C-Mixup}(k) &= \sigma_{x}^2 \sum_{i=p_1+1}^{p_2} \frac{\lambda_i}{(\lambda_i+k)^2}\\
&\leq \sigma_{x}^2 \sum_{i=p_1+1}^{p_2} \frac{\lambda_i}{k^2}\\
&\leq \frac{16c}{c_4^2n^{o(1)}}\frac{\sigma_x^2p_1}{n}. \qquad(k > c_4\sqrt{\frac{p_2}{p_1}}n^{1/4+o(1)}, \ \mathrm{Lemma}~\ref{B3_4_lemma_lambda})
\end{split}
\end{equation}

For mixup and mixup with feature similarity, we bound the second term as:

\begin{equation}
\begin{split}
\nonumber
\gamma_2^{feat}(k)\wedge \gamma_2^{mixup}(k)
&\geq \sigma_{x}^2 \sum_{i=1}^{p_1} \frac{1}{4k} \geq \frac{\sigma_x^2p_2}{4c_5n}. \qquad(k < c_5n)
\end{split}
\end{equation}

Thus there exists some constants $c$, $c_4$, $c_5$ (for example, $c_4 = 4\sqrt{c}$ and $c_5 = 1/4$), such that when n is sufficiently large, with probability at least $1-o(1)$:

\begin{equation}
\nonumber
    \mathbb{E}\lvert \theta_{C-Mixup}^{*}(k) - \theta \rvert^2 < \min(\mathbb{E}\lvert \theta_{feat}^{*}(k) - \theta \rvert^2, \mathbb{E}\lvert \theta_{mixup}^{*}(k) - \theta \rvert^2).
\end{equation}
which can reduce to the results immediately.
\subsection{Proofs of Lemmas}\label{pf:lem}
\noindent\textbf{Proof of Lemma~\ref{lem:sigmoid}.}
In order to prove Lemma~\ref{lem:sigmoid}, let us invoke the First-order Stein's Identity \citep{stein2004use}. 
\begin{lemma}
Let $X \in \R^d$ be a real-valued random vector with density $p$. Assume that $p$: $\R^d \to R$ is differentiable. In addition, let $g : \R^d \to \R$ be a continuous function such that $\E[\nabla g(X)]$ exists. Then it holds that$$
\E[g(X)\cdot S(X)]=\E[\nabla g(X)],
$$
where $S(X)=-\nabla p(x)/p(x)$ is the score function of $p$.
\end{lemma}
Now, let us plug in the density of $N_{d_1}(0,\Sigma_X)$, $p(x)=ce^{x^\top\Sigma_X^{-1}x/2}$ for some constant $c$. We then have $\nabla p(x)=ce^{x^\top\Sigma_X^{-1}x/2}\cdot \Sigma_X^{-1}x$ and $\nabla p(x)/p(x)= \Sigma_X^{-1} x$.

As a result, we have $$
\E[p^*(x) \Sigma_{X}^{-1}x]=\E[\nabla p^*(x)],
$$
implying $$
\E[p^*(x)x]= \Sigma_{X}\E[\nabla p^*(x)].
$$
Then recall that $p^*(x)=g(\theta^\top x)$, so we have $\nabla p^*(x)=g'(\theta^\top x)\theta$. Combining all the piece, we obtain
$$
\E[p^*(x)x]= \Sigma_{X}\E[g'(\theta^\top x)]\theta.
$$
Now plugging in $x\sim N(\theta,I)$, we have
$$\E[y_ix_i]=
(\E[g'(x_i^\top\theta)]+\E[g(x_i^\top\theta)])\cdot \theta.$$

\noindent\textbf{Proof of Lemma~\ref{B3_4_lemmaweyl}.}
Denote the entry of $E$ as $\epsilon_{ij}$, $\epsilon_{ij} \sim subG(\sigma_{\epsilon}^2)$, i.e. sub-gaussian distribution with variance proxy $\sigma_{\epsilon}^2$. From~\citet{rigollet17hds}, we find $\epsilon_{ij}^2 \sim subE(16\sigma_{\epsilon}^2)$, i.e. sub-exponential distribution with variance proxy $16\sigma_{\epsilon}^2$. Thus we choose $\delta_1$ from the Bernstein's inequality:
\begin{equation}
    \label{lemm1bernstein}
    \mathbb{P}(\frac{1}{np}\sum_{ij}{\epsilon_{i, j}^2} > t) \leq \exp[-\frac{np}{2}\min(\frac{t^2}{(16\sigma_{\epsilon}^2)^2}, \frac{t}{16\sigma_{\epsilon}^2})] = \delta_1
\end{equation}
Since $p_1, p_2 \geq 1$ we get $\delta_1 > e^{-n} \geq e^{-np/2}$, then:
\begin{align*}
    t &= 16\sigma_{\epsilon}^2\max(\sqrt{\frac{2}{np}\log{\frac{1}{\delta_1}}}, \frac{2}{np}\log{\frac{1}{\delta_1}})\\
    & = 16\sigma_{\epsilon}^2\sqrt{\frac{2}{np}\log{\frac{1}{\delta_1}}}\\
    & \leq 16\sigma_{\epsilon}^2 
\end{align*}

Thus, based on $\sigma_{\epsilon}^2  \leq \frac{c_3}{pn^{3/2}}$, with probability at least $1-\delta_1$, we have:
$$
    \Vert E \Vert_F = \sum_{i, j}\epsilon_{ij}^2 \leq npt \leq 16\frac{c}{\sqrt{n}},
$$
where $\| \cdot \|_F$ represents Frobenius norm. Then, with Hoffman-Weilandt's inequality~\cite{rigollet17hds} we prove that:

$$
    \max_{u} \lvert \hat\lambda_u - \lambda_u \rvert \leq \Vert E \Vert_F
$$


\noindent\textbf{Proof of Lemma~\ref{threshold_mix}.}
Since $z_i \sim \mathcal{N}_{p_1}(0,\sigma_x^2I_{p_1})$, we have $(z_i - z_j)^T\theta_1 \sim \mathcal{N}(0, 2\sigma_x^2\lvert \theta_1\rvert^2)$ for every $i \neq j$. Then for $t$ satisfies $ 0 < t \ll 1$, we have:
\begin{equation}
\label{normal_equ}
\begin{split}
p_{ij} &:= \mathbb{P}(\lvert (z_i - z_j)^T\theta_1\rvert < t)\\
& \leq 2 \Phi(\frac{t}{\sqrt{2}\sigma_x\lvert \theta\rvert}) - 1\\
& = \frac{t}{\sqrt{\pi}\sigma_x\lvert \theta\rvert} + o(t^2),
\end{split}
\end{equation}
where $\Phi(u) = \frac{1}{\sqrt{2\pi}}\int_{-\infty}^{u}e^{-q^2/2} dq$ is the distribution function of standard normal distribution. Thus if $t = \frac{\sqrt{\pi}\sigma_x\lvert\theta\rvert}{n(n-1)}\delta_1$, we have:

\begin{equation}
\label{union_equ}
\begin{split}
    \mathbb{P}(\min_{i\neq j}\lvert (z_i - z_j^{\prime})^T\theta_1 \rvert \geq t) 
    &= 1 - \mathbb{P}(\min_{i\neq j}\lvert (z_i - z_j^{\prime})^T\theta_1\rvert \leq t)\\
    &= 1 - \mathbb{P}(\bigcup\limits_{i\neq j}\{\lvert (z_i - z_j^{\prime})^T\theta_1\rvert \leq t\})\\
    &\geq 1 - \sum_{j\neq i} \sum_{i=1}^n p_{ij}\\
    &\geq 1 - \frac{n(n-1)}{2}(\frac{2t}{\sqrt{\pi}\sigma_x\lvert \theta\rvert}) \qquad (\mathrm{Eqn.}~\eqref{normal_equ})\\
    &= 1 - \delta_1.
\end{split}
\end{equation}

On the other hand, $\max\limits_{i}\lvert y_i - y_i^{\prime}\rvert = \max\limits_{i}\lvert \theta_1^T\epsilon_i^{\prime} + \epsilon_i - \epsilon_{i^{\prime}}\rvert$, and for every $i$,  $(\theta_1^T\epsilon_i^{\prime} + \epsilon_i - \epsilon_{i^{\prime}}) \sim subG((2+\lvert\theta_1\rvert^2)\sigma_{\epsilon}^2)$, then by maximum inequality~\cite{rigollet17hds}, when $n$ is sufficiently large, with probability at least $1-\delta_1$:
\begin{equation}
\begin{split}
\max_{i}\lvert y_i - y_i^{\prime}\rvert &\leq \sqrt{2(2+\lvert\theta_1\rvert^2)\log(n/\delta_1)}\cdot \sigma_{\epsilon}\\
&\leq 2\sigma_{\epsilon}\sqrt{(2+\lvert\theta_1\rvert^2)n} \qquad(e^{-n} < \delta_1 \ll 1)\\
&\leq \frac{2\sqrt{2+\lvert\theta_1\rvert^2}}{c_2\sqrt{\pi}}t \qquad(\sigma_x \geq c_2 \frac{n^{5/2}}{\Vert{\theta}\Vert\delta}\sigma_{\epsilon})\\
&\leq \frac{2\sqrt{2+\lvert\theta_1\rvert^2}}{c_2\sqrt{\pi}}\min_{i\neq j}\lvert (z_i - z_j^{\prime})^T\theta_1 \rvert \qquad (\mathrm{Eqn}.~\eqref{union_equ})\\
&\leq \frac{2\sqrt{2+\lvert\theta_1\rvert^2}}{c_2\sqrt{\pi}}(\min_{i\neq j}\lvert y_i - y_j^{\prime}\rvert + \sqrt{2\log n}\sigma_{\epsilon})\\
&\leq \frac{4\sqrt{2+\lvert\theta_1\rvert^2}}{c_2\sqrt{\pi}}\min_{i\neq j}\lvert y_i - y_j^{\prime}\rvert \qquad(t \geq c_2\sqrt{\pi n}\sigma_{\epsilon} \geq 2\sqrt{2\log n}\sigma_{\epsilon})\\
\end{split}
\end{equation}

Choose $c_2$ such that $ c_{gap} := \frac{c_2\sqrt{\pi}}{4\sqrt{2+\lvert \theta_1\rvert^2}} > 1$ and denote $l =\min_{i\neq j}\lvert y_i - y_j^{\prime}\rvert$, then we finish the proof with feasible threshold range $\frac{l}{c_{gap}} < h < l$.

\noindent\textbf{Proof of Lemma~\ref{lemma_bandwidth}.}
We define $Q_i=1-S_i$. From C-Mixup, we obtain:
$$
    P((x_j^{\prime}, y_j^{\prime})|(x_i, y_i)) \propto  \exp\left(- \frac{(y_i - y_j^{\prime})^2}{2 h^2}\right)
$$
Then, by $\max\limits_{i}\lvert y_i - y_i^{\prime}\rvert \leq \frac{l}{c_{gap}}$, we have
\begin{equation*}
    \begin{split}
    \mathbb{E}(S) &\geq \frac{1}{K}\exp(-\frac{l^2}{2h^2c_{gap}^2})\\
    \mathbb{E}(Q) &\leq \frac{1}{K}(n-1)\exp(-\frac{l^2}{2h^2})\\
    \end{split}
\end{equation*}
where $K = \sum_{j=1}^{n}(\exp\left(- (y_j^{\prime} - y_i)^2/(2 h^2)\right))$ is used for normalization. The upper bound of bandwidth $h$ is:
\begin{equation}
\nonumber
    \begin{split}
    h \ < \ c_6 \frac{l}{\sqrt{\log(n^2/p_1)}} \ < \ l\cdot\sqrt{\frac{c_{gap}^2-1}{2c_{gap}^2}}\log^{-\frac{1}{2}}(\frac{(n-1)(n-p_1/4)}{p_1/4}).
    \end{split}
\end{equation}
Since $S_i + Q_i = 1$, we obtain
$$
    \mathbb{E}(S) \geq \frac{n-p_1/4}{n}, \qquad \mathbb{E}(Q) \leq \frac{p_1/4}{n}.
$$

Finally, since $S_i, Q_i \in [0,1]$, we apply Hoeffding's inequality and obtain: 
$$
    \mathbb{P}(\frac{1}{n}\sum_{i=1}^{n}S_i - \mathbb{E}(S) < -t) \leq \exp(-2nt^2) = \delta_1
$$
Then with probability at least $1-\delta_1$, we have:
\begin{equation}
\nonumber
    \begin{split}
        \sum_{i=1}^{n}S_i &\geq n(\mathbb{E}(S) - \sqrt{\frac{1}{2n}\log(\frac{1}{\delta_1})})\\
        &\geq n - \frac{p_1}{2}\\
    \end{split}
\end{equation}

\noindent\textbf{Proof of Lemma~\ref{B3_4_lemma_lambda}} For C-Mixup, according to Lemma~\ref{lemma_bandwidth}, the corresponding noise-less matrix $\hat X$ is close to $(Z, O)$ when $n$ is sufficient large. Here, $Z \in \mathbb{R}^{n\times p_1}$ has rows $z_i$ and $O \in \mathbb{R}^{n\times p_2}$ is a matrix with at most $rank(\max(\frac{1}{2}p_1, p_2))$. We now simplify $O$ to be a zero matrix. In fact, Eqn.~\ref{eq_gamma2_y} in the following proof just scale at most $3/2$ when $O$ is $rank(\max(\frac{1}{2}p_1, p_2))$, which does not affect the final results. From Theorem 4.6.1 in~\citet{vershynin20hdp}, we find that there exists some positive absolute constants, such that with probability at least $1-\delta_1$, 
$$
    \sqrt{n} - C(\sqrt{p_1}+\sqrt{\log(2/\delta_1)}) \leq \hat\lambda_{p_1} \leq \hat\lambda_{1} \leq \sqrt{n} + C(\sqrt{p_1}+\sqrt{\log(2/\delta_1)})
$$

And $\hat\lambda_i = 0$ for $p_1 < i \leq p$. From~Eqn.\eqref{eq:weyl} and $\log(1/\delta_1) < n^{1-o(1)}$, we get Eqn.~\eqref{singular_y}. 

\section{Additional Experiments of In-Distribution Generalization}

\subsection{Detailed Dataset Description}
\label{app:c_1}

In this section, we provide detailed descriptions of datasets used in the experiments of in-distribution generalization.

\textbf{Airfoil Self-Noise~\cite{misc_airfoil_self-noise_291}} contains aerodynamic and acoustic test results for different sizes NACA 0012 airfoils at various wind tunnel speeds and angles of attack. Specifically, each input have 5 features, including frequency, angle of attack, chord length, free-stream velocity, and suction side displacement thickness. The label is one-dimensional scaled sound pressure level. Min-max normalization is used to normalize input features. Follow~\cite{hwang2021mixrl}, the number of examples in training, validation, and test sets are 1003, 300, and 200, respectively.

\textbf{NO2.} The NO2 emission dataset ~\cite{magne2004statlib} originated from the study where air pollution at a road is related to traffic volume and meteorological variables. Each input contains 7 features, including logarithm of the number of cars per hour, temperature 2 meter above ground, wind speed, temperature difference between 25 and 2 meters above ground, wind direction, hour of day and day number from October 1st. 2001. The hourly values of the logarithm of the concentration of NO2, which was measured at Alnabru in Oslo between October 2001 and August 2003, are used as the response variable, i.e., the label. Follow~\cite{hwang2021mixrl}, the number of training, validation and test sets are 200, 200 and 100, respectively.

\textbf{Exchange-Rate} is a time-series dataset that contains the collection of the daily exchange rates of 8 countries, including Australia, British, Canada, Switzerland, China, Japan, New Zealand and Singapore ranging from 1990 to 2016. The length of the entire time series is 7,588, and they adopt daily sample frequency. The slide window size is 168 days. The input dimension is $168 \times 8$ and the label dimension is $1 \times 8$ data. The dataset has been split into training ($60\%$), validation set ($20\%$) and test set ($20\%$) in chronological order as used in~\citet{lai2018modeling}. 

\textbf{Electricity}~\cite{Dua2019uci} is also a time-series dataset collected from 321 clients, which covers the electricity consumption in kWh every 15 minutes from 2012 to 2014. The length of the entire time-series is 26,304 and we use the hourly sample rate. Similar to Exchange-Rate data, the window size is set to 168, thus the input dimension is $168 \times 321$ the corresponding label dimension is $1 \times 321$. The dataset is also split as~\citet{lai2018modeling}.

\textbf{Echocardiogram Videos (Echo)}~\cite{ouyang2020video} includes 10,030 apical-4-chamber labeled echocardiogram videos from different aspects and human expert annotations to study cardiac motion and chamber sizes. These videos were collected from individuals who underwent imaging at Stanford University Hospital between 2016 and 2018. To identify the area of the left ventricle, we first preprocess the videos with frame-by-frame semantic segmentation. This method outputs video clips that contain 32 frames of $112 \times 112$ RGB images, which are be used to predict ejection fraction. The entire dataset are split into training, validation and test sets with size 7,460, 1,288, and 1,276, respectively.

\subsection{Hyperparameters}
\label{app:c_2}
We list the hyperparameters for every dataset in Table~\ref{tab:hyperameter_id}. Here, as we mentioned in Line 235-236 in the main paper, we apply k-Mixup, Local Mixup, MixRL, and C-Mixup to both mixup and Manifold Mixup, and report the best-performing one. Thus, we also treat the mixup type as another hyperparameter. All hyperparameters are selected by cross-validation. In addition, in Section F.3.1 of Appendix, we provide some guidance about how to tune and pick bandwidth $\sigma$. The guidance is also suitable to tasks beyond in-distribution generalization, i.e., task generalization and out-of-distribution robustness.

\begin{table}[h]
    \centering
    \small
    \caption{Hyperparameter settings for the experiments of in-distribution generalization. Here, FCN3 means 3-layer fully connected network and ManiMix means Manifold Mixup.}
    \vspace{0.5em}
    \label{tab:hyperameter_id}
    \begin{tabular}{l|ccccc}
    \toprule
        Dataset & Airfoil & NO2 & Exchange-Rate & Electricity & Echo  \\
        \midrule
        Learning rate & 1e-2 & 1e-2 & 1e-3 & 1e-3 & 1e-4\\
        Weight decay & 0 & 0 & 0 & 0 & 1e-4 \\
        Scheduler & n/a & n/a & n/a & n/a & StepLR\\ 
        Batch size & 16 & 32 & 128 & 128 & 10 \\
        Type of mixup & ManiMix & mixup & ManiMix & mixup & mixup \\
        Architecture & FCN3 & FCN3 & LST-Attn & LST-Attn & EchoNet-Dynamic \\
        Horizon & n/a & n/a & 12 & 24 & n/a \\
        Optimizer & Adam & Adam & Adam & Adam & Adam \\
        Maximum Epoch & 100 & 100 & 100 & 100 & 20 \\
        Bandwidth $\sigma$ & 1.75 & 1.2 & 5e-2 & 0.5 & 100.0 \\
        $\alpha$ in Beta Dist.  & 0.5 & 2.0 & 1.5 & 2.0 & 2.0\\
        \bottomrule
    \end{tabular}
\end{table}

\subsection{Overfitting}
\label{app:c_3}
\begin{figure}[h]
\vspace{-1em}
\centering
\begin{subfigure}[c]{0.40\textwidth}
		\centering
\includegraphics[width=\textwidth]{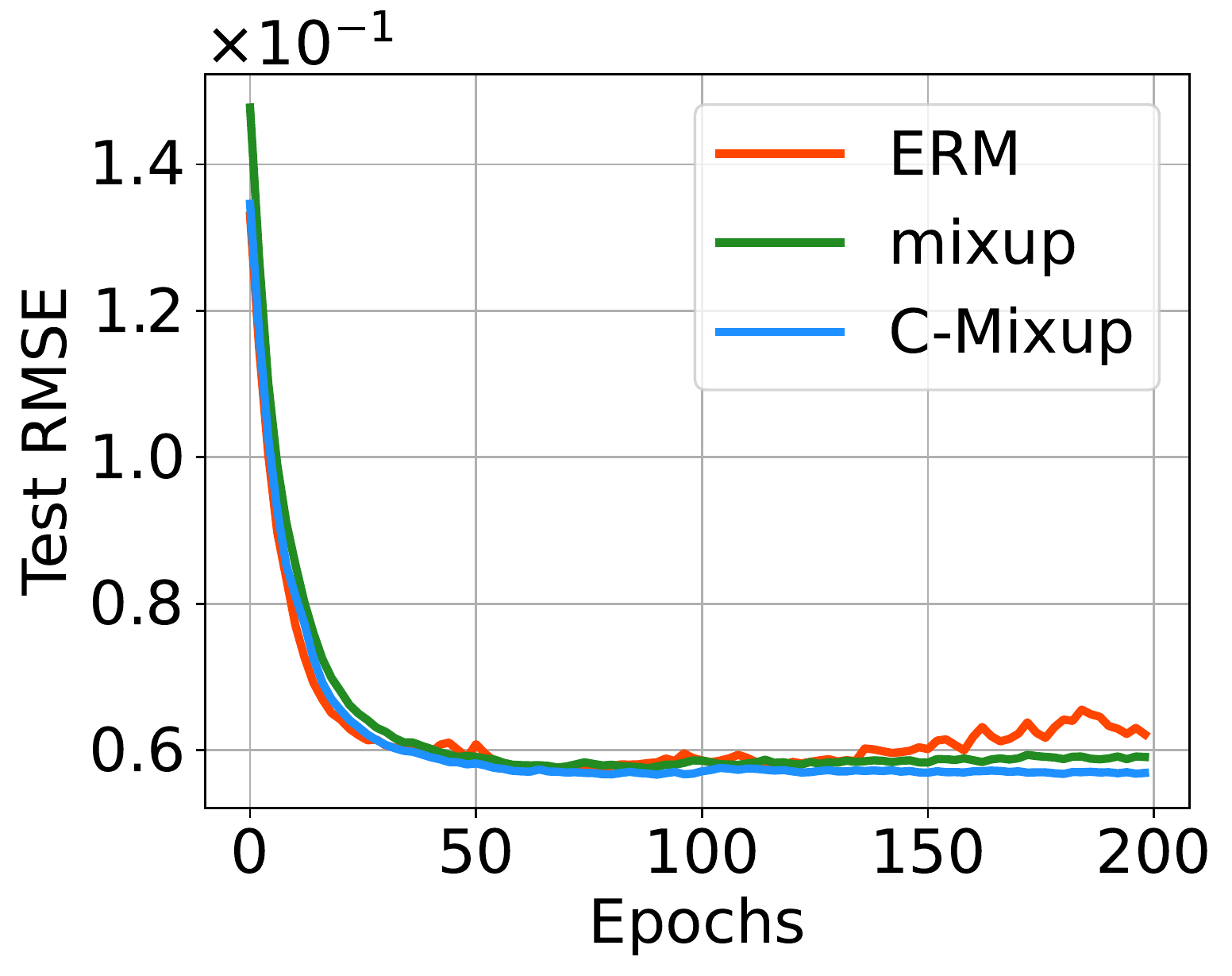}
    \caption{\label{fig:overfit_ele_loss}: Test Loss}
\end{subfigure}
\begin{subfigure}[c]{0.40\textwidth}
		\centering
\includegraphics[width=\textwidth]{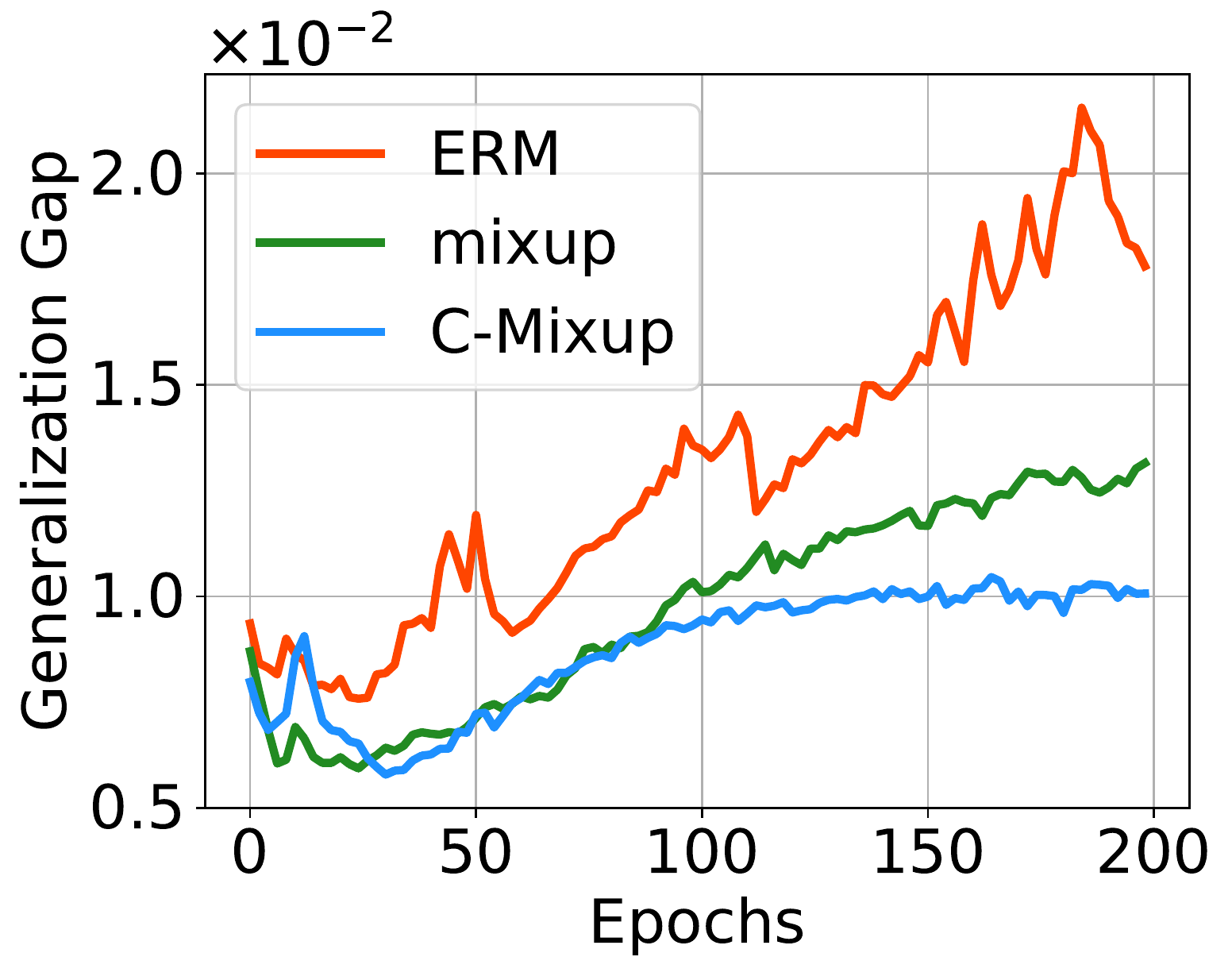}
    \caption{\label{fig:overfit_ele_gap}: Generalization Gap}
\end{subfigure}
\caption{Additional overfitting analysis of electricity}
\vspace{-1em}
\label{fig:app_overfitting}
\end{figure}

In Figure~\ref{fig:app_overfitting}, we visualize additional overfitting analysis on Electricity and the results corroborate our findings in the main paper, where C-Mixup reduces the generalization gap and achieves better test performance.

\subsection{Full Results}
\label{app:c_4}
In Table~\ref{tab:app_id_full}, we report the full results of in-distribution generalization.

\begin{table*}[h]
\small
\centering
\caption{Full results of in-distribution generalization. We compute the mean and standard deviation for results of three seeds.}
\label{tab:app_id_full}
\resizebox{\columnwidth}{!}{\setlength{\tabcolsep}{1.2mm}{
\begin{tabular}{l|l|c|c|c|c|c}
\toprule

& & Airfoil  & NO2  & Exchange-Rate  & Electricity & Echo \\\midrule
\multirow{7}{*}{RMSE} & ERM & \underline{2.901 $\pm$ 0.067} & 0.537 $\pm$ 0.005  & 0.0236 $\pm$ 0.0031 & 0.0581 $\pm$ 0.0011 & 5.402 $\pm$ 0.024\\
& mixup & 3.730 $\pm$ 0.190 & 0.528 $\pm$ 0.005  & 0.0239 $\pm$ 0.0027 & 0.0585 $\pm$ 0.0004 & \underline{5.393 $\pm$ 0.040}\\
& Mani mixup & 3.063 $\pm$ 0.113 & 0.522 $\pm$ 0.008  & 0.0242 $\pm$ 0.0043 &  0.0583 $\pm$ 0.0004 & 5.482 $\pm$ 0.066\\
& k-Mixup &  2.938 $\pm$ 0.150  & 0.519 $\pm$ 0.005  & 0.0236 $\pm$ 0.0029 & \underline{0.0575 $\pm$ 0.0002} & 5.518 $\pm$ 0.034\\
& Local Mixup & 3.703 $\pm$ 0.151  & \underline{0.517 $\pm$ 0.004}  & \underline{0.0236 $\pm$ 0.0024}  & 0.0582 $\pm$ 0.0004 & 5.652 $\pm$ 0.043\\
& MixRL & 3.614 $\pm$ 0.293 & 0.527 $\pm$ 0.003 & 0.0238 $\pm$ 0.0037 & 0.0585 $\pm$ 0.0006 &  5.618 $\pm$ 0.071\\
\cmidrule{2-7}
& \textbf{C-Mixup (Ours)} & \textbf{2.717} $\pm$ \textbf{0.067} & \textbf{0.509} $\pm$ \textbf{0.006}  & \textbf{0.0203} $\pm$ \textbf{ 0.0011} & \textbf{0.0570} $\pm$ \textbf{0.0006} & \textbf{5.177} $\pm$ \textbf{0.036}\\
\midrule\midrule
\multirow{7}{*}{MAPE} & ERM & \underline{1.753 $\pm$ 0.078\%}  & 13.615 $\pm$ 0.165\%  & 2.423 $\pm$ 0.365\%  & \underline{13.861 $\pm$ 0.152\%} & \underline{8.700 $\pm$ 0.015\%}\\
& mixup  & 2.327 $\pm$ 0.159\% & 13.534 $\pm$ 0.125\%  & 2.441 $\pm$ 0.286\%  & 14.306 $\pm$ 0.048\% & 8.838 $\pm$ 0.108\%\\
& Mani mixup  & 1.842 $\pm$ 0.114\%  & 13.382 $\pm$ 0.360\%  & 2.475 $\pm$ 0.346\%   & 14.556 $\pm$ 0.057\% & 8.955 $\pm$ 0.082\%\\
& k-Mixup  & 1.769 $\pm$ 0.035\% & \underline{13.173 $\pm$ 0.139\%} & 2.403 $\pm$ 0.311\% & 14.134 $\pm$ 0.134\% & 9.206 $\pm$ 0.117\%\\
& Local Mixup & 2.290 $\pm$ 0.101\%  & 13.202 $\pm$ 0.176\% & \underline{2.341 $\pm$ 0.229\%}  & 14.245 $\pm$ 0.152\% & 9.313 $\pm$ 0.115\%\\
& MixRL & 2.163 $\pm$ 0.219\% & 13.298 $\pm$ 0.182\% & 2.397 $\pm$ 0.296\% & 14.417 $\pm$ 0.203\% & 9.165 $\pm$ 0.134\%\\
\cmidrule{2-7}
& \textbf{C-Mixup (Ours)} & \textbf{1.610} $\pm$ \textbf{0.085\%}  & \textbf{12.998} $\pm$ \textbf{0.271\%} & \textbf{2.041} $\pm$ \textbf{0.134\%}  & \textbf{13.372} $\pm$ \textbf{0.106\%} & \textbf{8.435} $\pm$ \textbf{0.089\%}\\\bottomrule
\end{tabular}}}
\vspace{-1em}
\end{table*}

\section{Additional Experiments of Task Generalization}
\subsection{Detailed Dataset Description}
\label{app:d_1}

\textbf{ShapeNet1D.} We adopt the same preprocessing strategy to preprocess the ShapeNet1D dataset~\cite{gao2022matters}. The ShapeNet1D dataset contains 27 categories with 60 objects per category. For each category, we randomly select 50 objects for meta-training and the rest ones are used for meta-testing. The model takes a $128 \times 128$ grey-scale image as the input, and the label is normalized to $[0, 10]$.

\textbf{PASCAL3D.} In PASCAL3D, we follow~\cite{yin2020meta} to preprocess the dataset, where 50 and 15 categories are used for meta-training and meta-testing, respectively. The input image size and output label scale are same as ShapeNet1D.

\subsection{Hyperparameters.}
\label{app:d_2}
We list the hyperparameters used in the experiments of ShapeNet1D and PASCAL3D in Table~\ref{tab:hyperpara_meta}.

\begin{table*}[h]
\small
\caption{Hyperparameters of task generalization experiments.}
\vspace{-1em}
\label{tab:hyperpara_meta}
\begin{center}
\begin{tabular}{l|c|c}
\toprule
Hyperparameters &  ShapeNet1D & PASCAL3D \\\midrule
outer-loop learning rate& 0.0005 & 0.001 \\
inner-loop learning rate & 0.002 & 0.01\\
\# of inner-loop updates & 5 & 5\\
$\alpha$ in Beta Dist. & 0.5 & 0.5\\
batch size & 10 & 10 \\
support/query shot & 15 & 15 \\
max. iterations & 15,000 & 15,000 \\
\bottomrule
\end{tabular}
\end{center}
\vspace{-1em}
\end{table*}

\section{Additional Experiments of Out-of-Distribution Robustness}

\subsection{Detailed Dataset Description}
\label{app:e_1}

We provide detailed descriptions for datasets that are used in the experiments of out-of-distribution robustness.

\vspace{-0.5em}
\begin{wraptable}{r}{0.3\textwidth}
\small
\vspace{-2.5em}
\caption{Spurious correlation analysis in RCF-MNIST. We list the test performance w/ and w/o distribution shifts.}
\vspace{-0.5em}
\label{tab:spu_strength}
\begin{center}
\setlength{\tabcolsep}{1.2mm}{
\begin{tabular}{l|c|c}
\toprule
 & w/o shift  & w/ shift
\\\midrule
RMSE $\downarrow$ & 0.111 & 0.162 \\
\bottomrule
\end{tabular}}
\end{center}
\vspace{-1.5em}
\end{wraptable}

\textbf{RCF-MNIST. } The prefix "RCF" of RCF-MNIST means "Rotated-Colored-Fashion". To construct RCF-MNIST, assume the normalized RGB vector of red and blue is $[1,0,0]$ and $[0,0,1]$ and the normalized angle of rotation (i.e., label) for one image is $g \in [0,1]$. In training set, we color 80\% images with RGB value $[g,0,1-g]$ and the rest images are colored with $[1-g,0,g]$. Hence, the color information is strongly spuriously correlated with the label in the training set. In test set, we reverse spurious correlations to simulate distribution shift, where $80\%$ and $20\%$ images are colored with RGB values $[1-g,0,g]$ and $[g,0,1-g]$, respectively. We further verify that the spurious correlation between color and label affects the performance. Here, we compare the performance of same test set with or without distribution shift. The results are reported in Table~\ref{tab:spu_strength}, where we observe that the subpopulation shift caused by spurious correlation does 
hurt the performance as expected.

\textbf{PovertyMap} is included in the WILDS benchmark~\cite{koh2021wilds}, which contains satellite images from 23 African countries that can be used to predict the village-level real-valued asset wealth index. The input is a $224 \times 224$ multispectral LandSat satellite image with 8 channels, and the label is the real-valued asset wealth index. The domains of the images consist the country, urban and rural area information. This dataset includes 5 different cross validation folds, and all countries in these splits are disjoint to support the out-of-distribution setting. All experimental settings follow~\citet{koh2021wilds}. 

\textbf{Communities And Crimes (Crime)} is a tabular dataset combining socio-economic data from the 1990 US Census, law enforcement data from the 1990 US LEMAS survey, and crime data from the 1995 FBI UCR. The input features include 122 attributes that have some plausible connection to crime, such as the median family income and percent of officers assigned to drug units. The label attribute to be predicted is per capita violent crimes, which covers violent crimes including murder, rape, robbery, assault and so on. All numeric features are normalized into the decimal range $0.00 \sim 1.00$ by equal-interval binning method, and the missing values are filled with the average values of the corresponding attributes. State identifications are used as the domain information, resulting to 46 domains in total. We split the dataset into training, validation and test sets with size 1,390, 231 and 373, while they contain 31, 6 and 9 disjoint domains, respectively.


\textbf{SkillCraft.} SkillCraft is a UCI tabular dataset~\cite{misc_skillcraft1_master_table_dataset_272} originated from a study that used video game telemetry data from real-time strategy (RTS) games to explore the development of expertise. Input x contains 17 player-related parameters in the game, such as the Cognition-Action-cycle variables and the Hotkey Usage variables. And the action latency in the game was considered as the label y. Missing data are filled by mean padding on each attribute. We use "League Index", which correspond to different levels of competitors, to be the identifier of domain. The dataset is split into training, validation and test sets with size $1878$, $806$, $711$ and disjoint domain number $4$, $1$, $3$, respectively.

\textbf{Drug-target Interactions (DTI).} Drug-target Interactions dataset ~\cite{tdc} originated aims to predict the binding activity score between each small molecule and the corresponding target protein. The input features contain both drug and target protein information, which are represented by one-hot vectors. The output label is the binding activity score. The training and validation set are selected from 2013-2018, and test set is drawn from 2019-2020. We regard "Year" as the domain information. 

\subsection{Hyperparameters}
\label{app:e_2}
We list the hyperparameters for the experiments of out-of-distribution robustness in Table~\ref{tab:hyperameter_ood}.

\begin{table}[h]
    \centering
    \small
    \caption{Hyperparameter settings for the experiments of out-of-distribution robustness.}
    \label{tab:hyperameter_ood}
    \begin{tabular}{l|ccccc}
    \toprule
        Dataset & RCF-MNIST & PovertyMap & Crime & SkillCraft & DTI  \\
        \midrule
        Learning rate & 7e-5 & 1e-3 & 1e-3 & 1e-2 & 5e-5 \\
        Weight decay & 0 & 0 & 0 & 0 &0 \\
        Scheduler & n/a & StepLR & n/a & n/a & n/a\\ 
        Batch size & 64 & 64 & 16 & 32 & 64\\
        Type of mixup & ManiMix & CutMix & ManiMix & mixup & ManiMix\\
        Architecture & ResNet-18 & ResNet-50 & FCN3 & FCN3 & DeepDTA\\
        Optimizer & Adam & Adam & Adam & Adam & Adam\\
        Maximum Epoch  & 30 & 50 & 200 & 100 & 20\\
        Bandwidth $\sigma$ & 0.2 & 0.5 & 1.0 & 5e-4 & 21.0\\
        $\alpha$ in Beta Dist.  & 2.0 & 0.5 & 2.0 & 2.0 & 2.0\\
        \bottomrule
    \end{tabular}
\end{table}

\subsection{Full Results}
\label{app:e_3}
In Table~\ref{tab:app_results_ood}, we report the full results of out-of-distribution robustness.

\begin{table*}[h]
\small
\caption{Full results of out-of-distribution robustness. The standard deviations are calculated by 5-fold data split in PovertyMap~\cite{koh2021wilds}, or over 3 seeds in other datasets.}
\vspace{-0.5em}
\label{tab:app_results_ood}
\begin{center}
\resizebox{\columnwidth}{!}{\begin{tabular}{l|c|cc|cc}
\toprule
\multirow{2}{*}{} & RCF-MNIST (RMSE) & \multicolumn{2}{c|}{PovertyMap ($R$)} & \multicolumn{2}{c}{Crime (RMSE)} \\
& Avg. $\downarrow$ & Avg. $\uparrow$ & Worst $\uparrow$ & Avg. $\downarrow$ & Worst $\downarrow$ \\\midrule
ERM & 0.162 $\pm$ 0.003 & 0.80 $\pm$ 0.04 & \underline{0.50 $\pm$ 0.07} & 0.134 $\pm$ 0.003 & 0.173 $\pm$ 0.009 \\
IRM & \underline{0.153 $\pm$ 0.003} & 0.77 $\pm$ 0.05 & 0.43 $\pm$ 0.07 & \underline{0.127 $\pm$ 0.001} & 0.155 $\pm$ 0.003  \\
IB-IRM & 0.167 $\pm$ 0.003 &  0.78 $\pm$ 0.05 & 0.40 $\pm$ 0.05 & 0.127 $\pm$ 0.002 & 0.153 $\pm$ 0.004 \\
V-REx & 0.154 $\pm$ 0.011 & \textbf{0.83 $\pm$ 0.02} & 0.48 $\pm$ 0.03 & 0.129 $\pm$ 0.005 & 0.157 $\pm$ 0.007  \\
CORAL & 0.163 $\pm$ 0.016 & 0.78 $\pm$ 0.05 & 0.44 $\pm$ 0.06  & 0.133 $\pm$ 0.007 & 0.166 $\pm$ 0.015  \\
GroupDRO & 0.232 $\pm$ 0.016 & 0.75 $\pm$ 0.07 & 0.39 $\pm$ 0.06 & 0.138 $\pm$ 0.005 & 0.168 $\pm$ 0.009  \\
Fish & 0.263 $\pm$ 0.017 & 0.80 $\pm$ 0.02 & 0.30 $\pm$ 0.01 & 0.128 $\pm$ 0.000 &  \underline{0.152 $\pm$ 0.001}  \\
mixup & 0.176 $\pm$ 0.003 & \underline{0.81 $\pm$ 0.04} & 0.46 $\pm$ 0.03 & 0.128 $\pm$ 0.002 & 0.154 $\pm$ 0.001 \\
\midrule
\textbf{Ours} & \textbf{0.146 $\pm$ 0.005} & \underline{0.81 $\pm$ 0.03} & \textbf{0.53 $\pm$ 0.07} & \textbf{0.123} $\pm$ \textbf{0.000} & \textbf{0.146} $\pm$ \textbf{0.002}  \\
\midrule\midrule
\multirow{2}{*}{} & n/a & \multicolumn{2}{c|}{SkillCraft (RMSE)} & \multicolumn{2}{c}{DTI ($R$)} \\
& n/a &   Avg. $\downarrow$ & Worst $\downarrow$ & Avg. $\uparrow$ & Worst $\uparrow$ \\\midrule
ERM & n/a &  5.887 $\pm$ 0.362 & 10.182 $\pm$ 1.745 & 0.464 $\pm$ 0.014 & 0.429 $\pm$ 0.004 \\
IRM & n/a & 5.937 $\pm$ 0.254 & 7.849 $\pm$ 0.371  & 0.478 $\pm$ 0.007 &  0.432 $\pm$ 0.003 \\
IB-IRM & n/a &  6.055 $\pm$ 0.503 &  7.650 $\pm$ 0.653 & 0.479 $\pm$ 0.009 & 0.435 $\pm$ 0.007\\
V-REx & n/a &  6.059 $\pm$ 0.429 & \underline{7.444 $\pm$ 0.494} & \underline{0.485 $\pm$ 0.009} & 0.435 $\pm$ 0.004 \\
CORAL & n/a &  6.353 $\pm$ 0.102 & 8.272 $\pm$ 0.436 & 0.483 $\pm$ 0.010 & 0.432 $\pm$ 0.005 \\
GroupDRO & n/a &  6.155 $\pm$ 0.537 &  8.131 $\pm$ 0.608 & 0.442 $\pm$ 0.043 &  0.407 $\pm$ 0.039 \\
Fish & n/a &   6.356 $\pm$ 0.201 & 8.676 $\pm$ 1.159 & 0.470 $\pm$ 0.022 & \underline{0.443 $\pm$ 0.010} \\
mixup & n/a &  \underline{5.764 $\pm$ 0.618} & 9.206 $\pm$ 0.878 & 0.465 $\pm$ 0.004 &  0.437 $\pm$ 0.016 \\
\midrule
\textbf{Ours} & n/a &  \textbf{5.201} $\pm$ \textbf{0.059} & \textbf{7.362 $\pm$ 0.244} &  \textbf{ 0.498} $\pm$ \textbf{0.008}  & \textbf{0.458} $\pm$ \textbf{0.004} \\
\bottomrule

\end{tabular}}
\end{center}
\vspace{-1em}
\end{table*}

\section{Additional Analysis of C-Mixup}
\subsection{Additional Compatibility Analysis}
\label{sec:app_compatibility}
In Table~\ref{si_tab:results_comparison}, we report the full results of compatibility analysis. Here, the performances on ERM, mixup, mixup+C-Mixup are also reported for comparison. In addition to the compatibility of C-Mixup, we also observe that some powerful inter-class mixup policies (e.g., PuzzleMix) improve the performance on part of regression tasks, e.g., RCF-MNIST. However, these approaches may also yield worse performances than ERM in other datasets, e.g., PovertyMap. Nevertheless, integrating C-Mixup on these mixup-based variants performs better than their vanilla versions, showing the compatibility and complementarity of C-Mixup to the existing mixup-based approaches in regression.

\begin{table*}[h]
\small
\caption{Full results (performance with standard deviation) of compatibility analysis.}
\label{si_tab:results_comparison}
\vspace{-1em}
\begin{center}
\setlength{\tabcolsep}{2mm}{{
\begin{tabular}{l|c|c|c}
\toprule
\multicolumn{2}{c|}{\multirow{2}{*}{Model}} & RCF-MNIST  & PovertyMap 
\\\cmidrule{3-4}
\multicolumn{2}{c|}{} & RMSE $\downarrow$ & Worst $R$ $\uparrow$ \\\midrule
\multicolumn{2}{c|}{ERM} & 0.162 $\pm$ 0.003 & 0.50 $\pm$ 0.07\\\midrule 
\multirow{2}{*}{mixup} &    & 0.176 $\pm$ 0.003 & n/a \\
 & +C-Mixup  & \textbf{0.146 $\pm$ 0.005} & 
 n/a
 \\\midrule
\multirow{2}{*}{CutMix} &     & 0.194 $\pm$ 0.010 & 0.46 $\pm$ 0.03  \\
  & +C-Mixup  & \textbf{0.186 $\pm$ 0.013}  & \textbf{0.53 $\pm$ 0.07}\\\midrule
\multirow{2}{*}{PuzzleMix} &  & 0.159 $\pm$ 0.004 & 0.47 $\pm$ 0.03\\
  & +C-Mixup  & \textbf{0.150 $\pm$ 0.012} & \textbf{0.50 $\pm$ 0.04} \\\midrule
\multirow{2}{*}{AutoMix} &    & 0.152 $\pm$ 0.021 & 0.49 $\pm$ 0.07\\
 & +C-Mixup  & \textbf{0.146 $\pm$ 0.009} & \textbf{0.53 $\pm$ 0.07} \\
\bottomrule
\end{tabular}}}
\end{center}
\vspace{-2em}
\end{table*}

\subsection{Distance Metrics}
\label{sec:app_distance}
In this section, we first discuss how to calculate the representation distance $d(h_i, h_j)$. Then, we provide complete analysis of distance metrics. 

\textbf{Measuring Representation Distance.} In this paper, we adopt a two-stage training process for each iteration. In the first stage, we feed the data into the current backbone and get hidden representations $h$, which is used to calculate the example distance, i.e., $d(h_i, h_j)$. In the second stage, we apply C-Mixup with representation distance.

\textbf{Complete Analysis of Distance Metrics}
We report full results of the analysis of distance metrics in Table~\ref{tab:app_similarity_comparison_full}. In addition to the existing analysis, we conduct one analysis by changing how to calculate the distance between low-dimensional hidden representations, where we compare the Euclidean distance and the cosine distance. We observe that C-Mixup performs better than using both Euclidean and cosine distances to measure the similarity between low-dimensional representations, corroborating the effectiveness of C-Mixup.

\begin{table}[h]
\small

\caption{Full results (performance with standard deviation) of different distance metrics.}

\label{tab:app_similarity_comparison_full}
\begin{center}
\setlength{\tabcolsep}{1.5mm}{
\begin{tabular}{l|c|c|c}
\toprule
\multirow{2}{*}{Model} & Exchange-Rate  & ShapeNet1D & DTI
\\\cmidrule{2-4}
& RMSE $\downarrow$ & MSE $\downarrow$ & Avg. $R$ $\uparrow$
\\\midrule
ERM/MAML & 0.0236 $\pm$ 0.0031  & 4.698 $\pm$ 0.079 & 0.464 $\pm$ 0.014  \\
mixup/MetaMix  & 0.0239 $\pm$ 0.0027 &  4.275 $\pm$ 0.082 & 0.465 $\pm$ 0.004 \\\midrule
$d(x_i, x_j)$  & 0.0212 $\pm$ 0.0014 & 4.539 $\pm$ 0.082 & 0.478 $\pm$ 0.003\\
$d(x_i\oplus y_i, x_j\oplus y_j)$  & 0.0212 $\pm$ 0.0009 & 4.395 $\pm$ 0.085 & 0.484 $\pm$ 0.002\\
$d(h_i, h_j)$ (Euclidean distance) & 0.0213 $\pm$ 0.0006 & 4.202 $\pm$ 0.078 & 0.483 $\pm$ 0.001\\
$d(h_i, h_j)$ (Cosine distance) & 0.0209 $\pm$ 0.0012 & 4.411 $\pm$ 0.081 & 0.477 $\pm$ 0.004\\
$d(h_i\oplus y_i, h_j\oplus y_j)$  & 0.0208 $\pm$ 0.0016 & 4.176 $\pm$ 0.077 & 0.487 $\pm$ 0.001 \\

\midrule
\textbf{$d(y_i, y_j)$ (C-Mixup)}  &  \textbf{0.0203} $\pm$ \textbf{0.0011} & \textbf{4.024 $\pm$ 0.081} & \textbf{0.498} $\pm$ \textbf{0.008} \\
\bottomrule
\end{tabular}}
\end{center}

\end{table}

\subsection{Additional Hyperparameter Sensitivity}
In this section, we first provide more experiments for bandwidth analysis. Then, we conduct experiments to show the effect of hyperparameter $\alpha$ in Beta distribution, i.e., $\mathrm{Beta}(\alpha, \alpha)$.
\subsubsection{Additional Bandwidth Analysis}
\label{sec:app_bandwidth}
\begin{figure}[h]
\centering
\begin{subfigure}[c]{0.22\textwidth}
		\centering
\includegraphics[width=\textwidth]{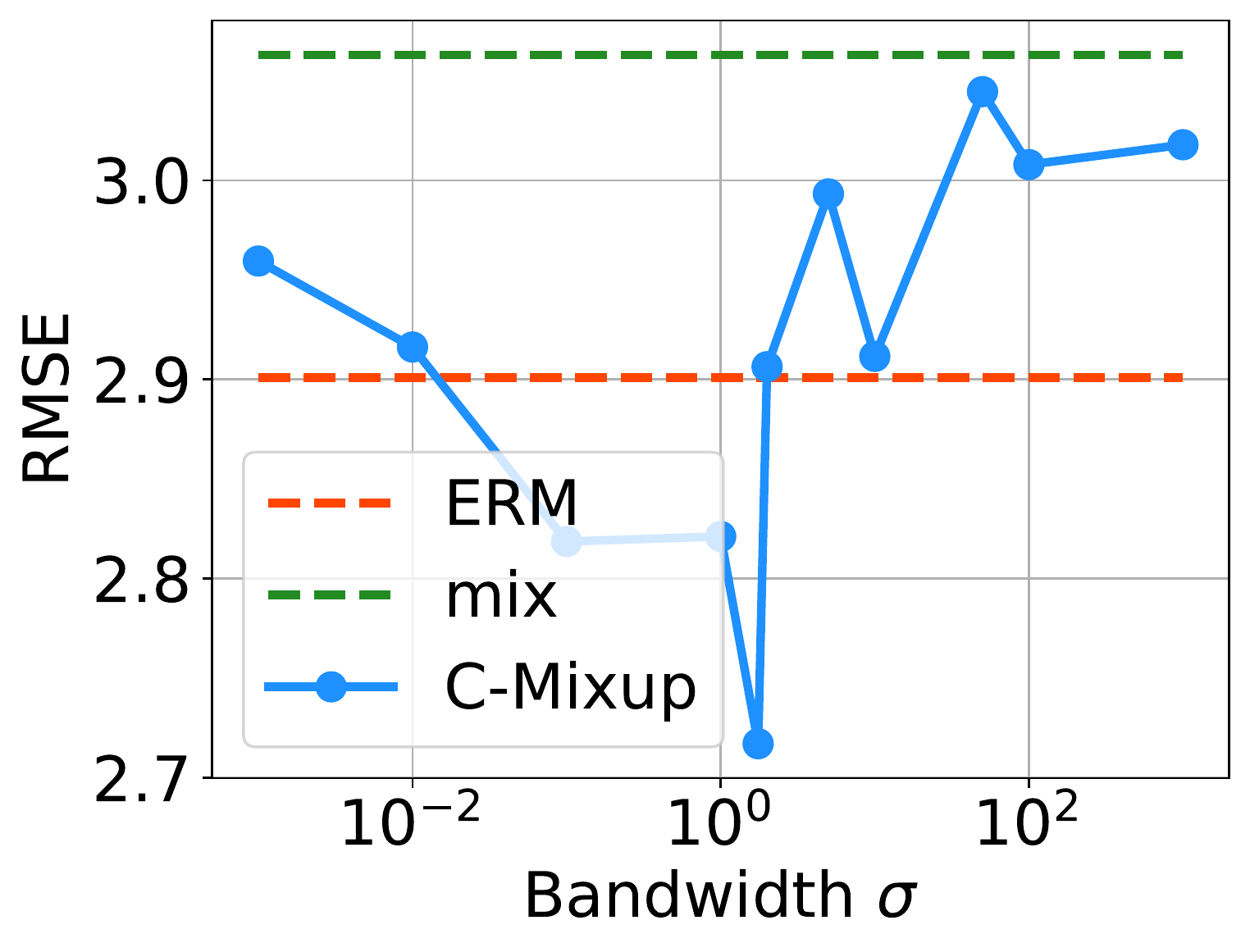}
    \caption{\label{fig:bandwidth_airfoil}: Airfoil}
\end{subfigure}
\begin{subfigure}[c]{0.22\textwidth}
		\centering
\includegraphics[width=\textwidth]{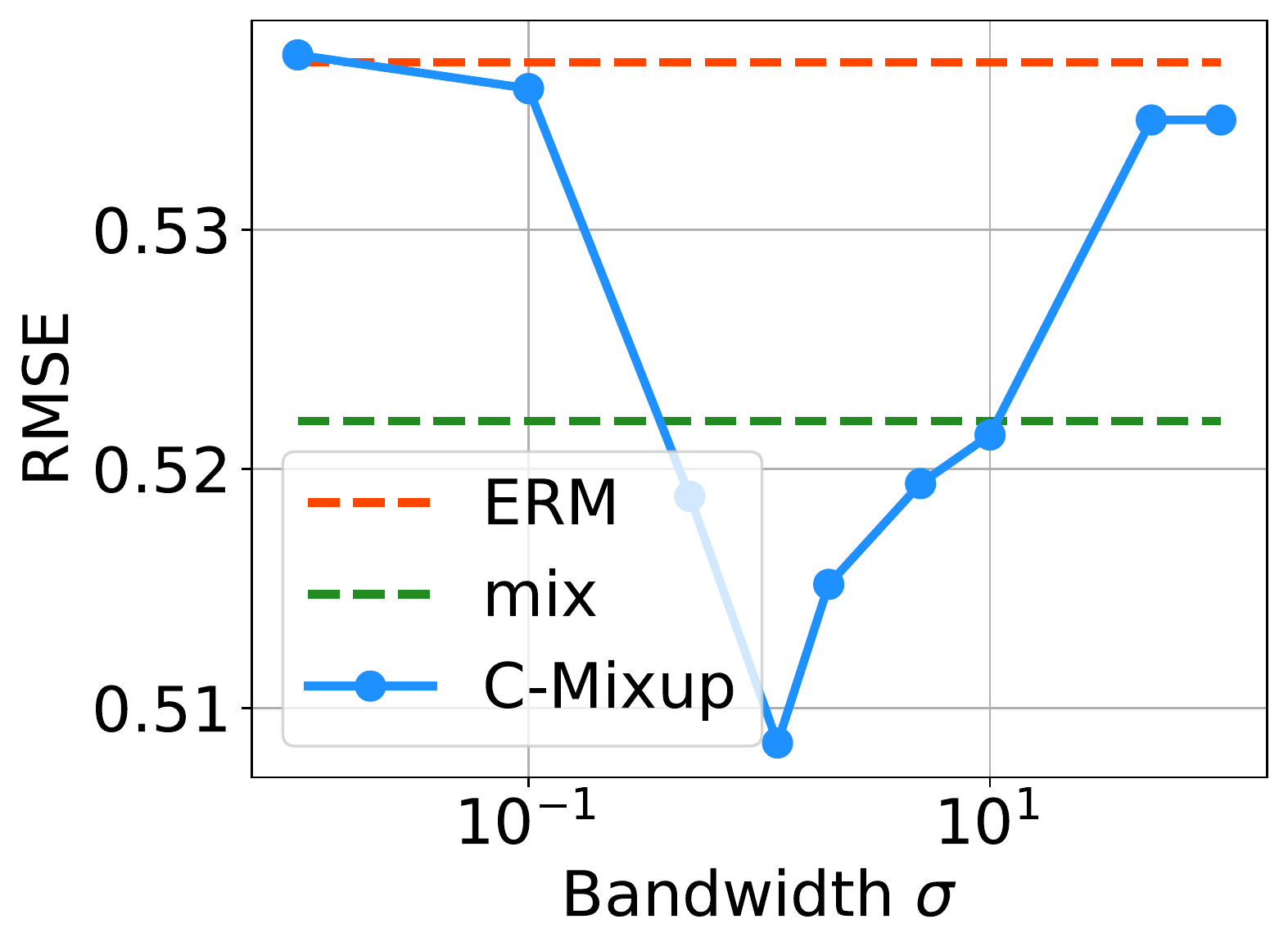}
    \caption{\label{fig:bandwidth_no2}: NO2}
\end{subfigure}
\begin{subfigure}[c]{0.22\textwidth}
		\centering
\includegraphics[width=\textwidth]{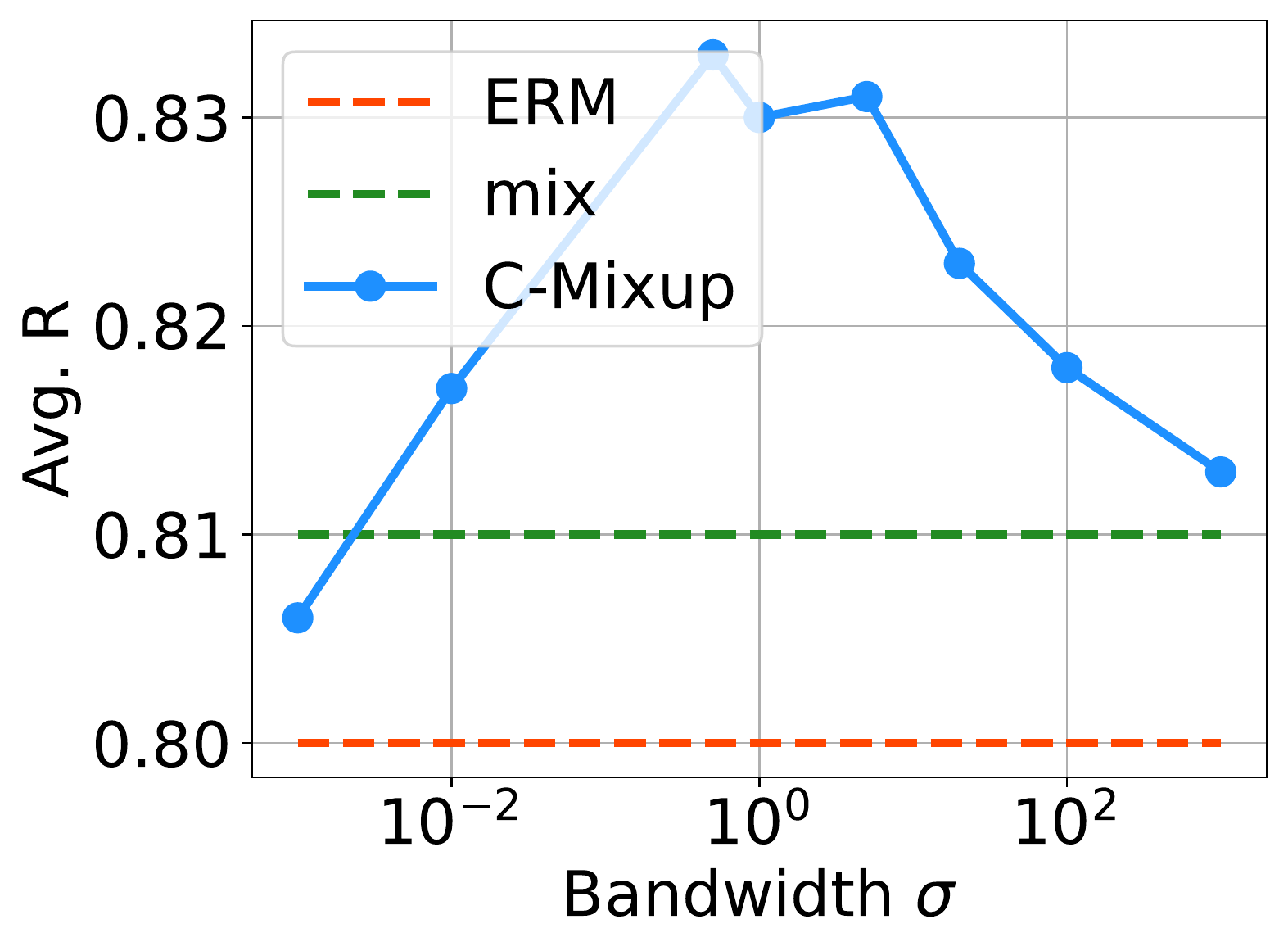}
    \caption{\label{fig:bandwidth_povertymap}: PovertyMap}
\end{subfigure}
\begin{subfigure}[c]{0.22\textwidth}
		\centering
\includegraphics[width=\textwidth]{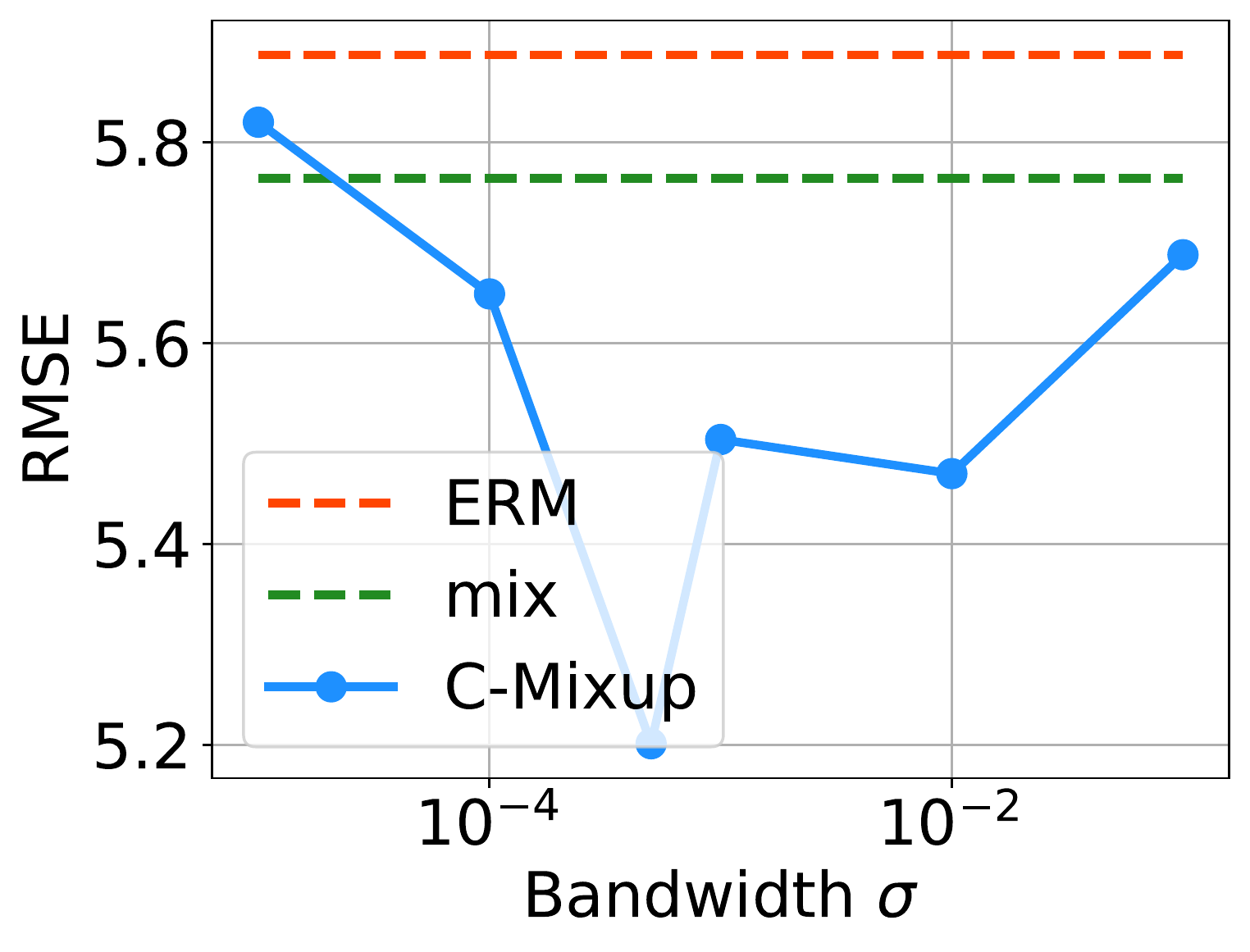}
    \caption{\label{fig:bandwidth_skillcraft}: SkillCraft}
\end{subfigure}
\caption{Additional robustness analysis of bandwidth}
\label{fig:app_bandwidth}
\end{figure}

We illustrate the bandwidth analysis for additional four datasets in Figure~\ref{fig:app_bandwidth}, including Airfoil, NO2, PovertyMap, SkillCraft. The results corroborate our finding in the main paper that C-Mixup yields a good model in a relative wide range of bandwidth, reducing the efforts to tune the bandwidth $\sigma$ for every specific dataset.

According to our empirical results, we conclude that roughly tuning the bandwidth in the range [0.01, 0.1, 1, 10, 100] is sufficient to get a relatively satisfying performance. To get the optimal bandwidth, we suggest to perform grid search.

\subsubsection{Effect of Shape Parameter $\alpha$ in Beta Distribution}
\label{sec:app_robust_alpha}

\begin{figure}[h]
\centering
\begin{subfigure}[c]{0.19\textwidth}
		\centering
\includegraphics[width=\textwidth]{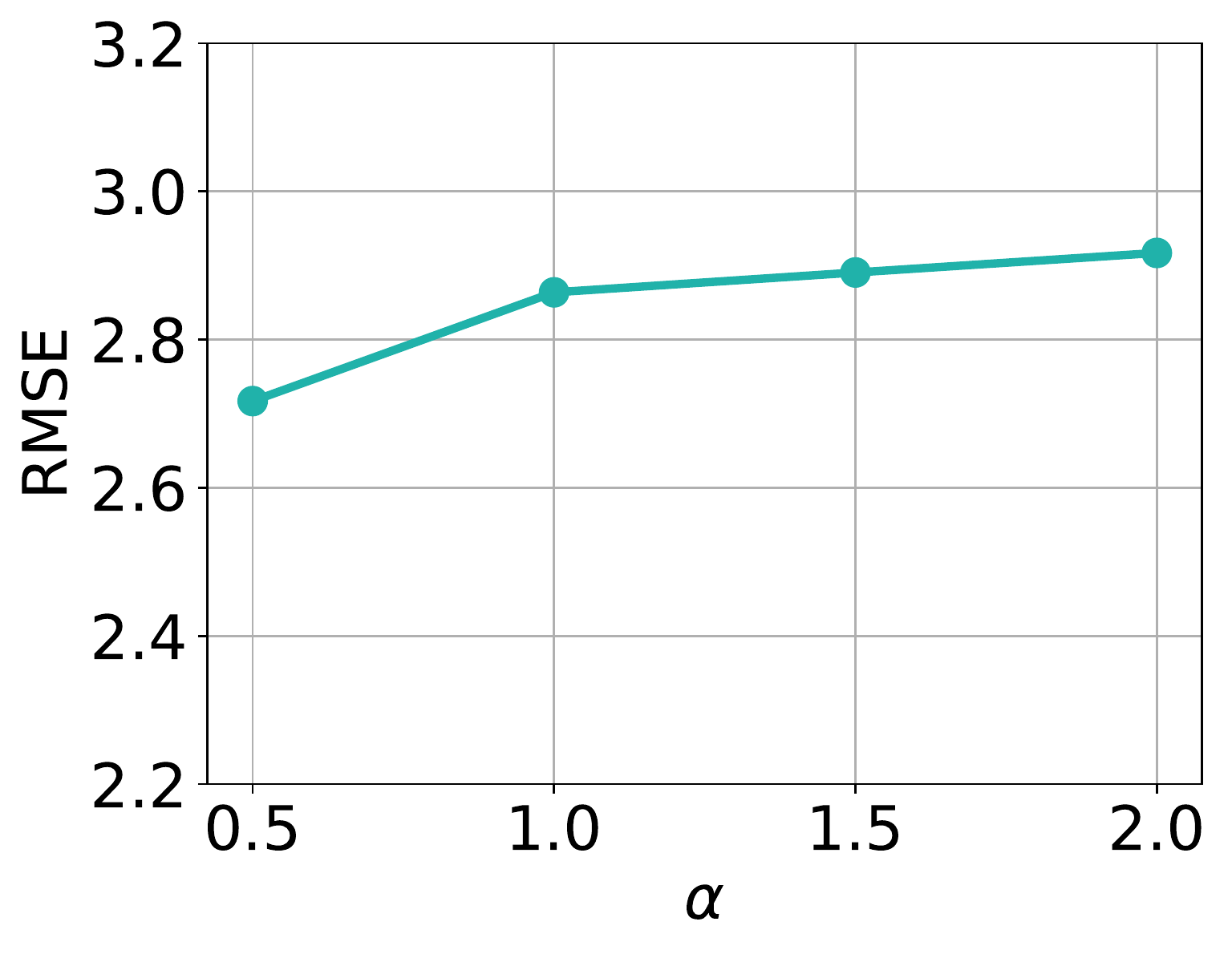}
    \caption{\label{fig:alpha_airfoil}: Airfoil}
\end{subfigure}
\begin{subfigure}[c]{0.19\textwidth}
		\centering
\includegraphics[width=\textwidth]{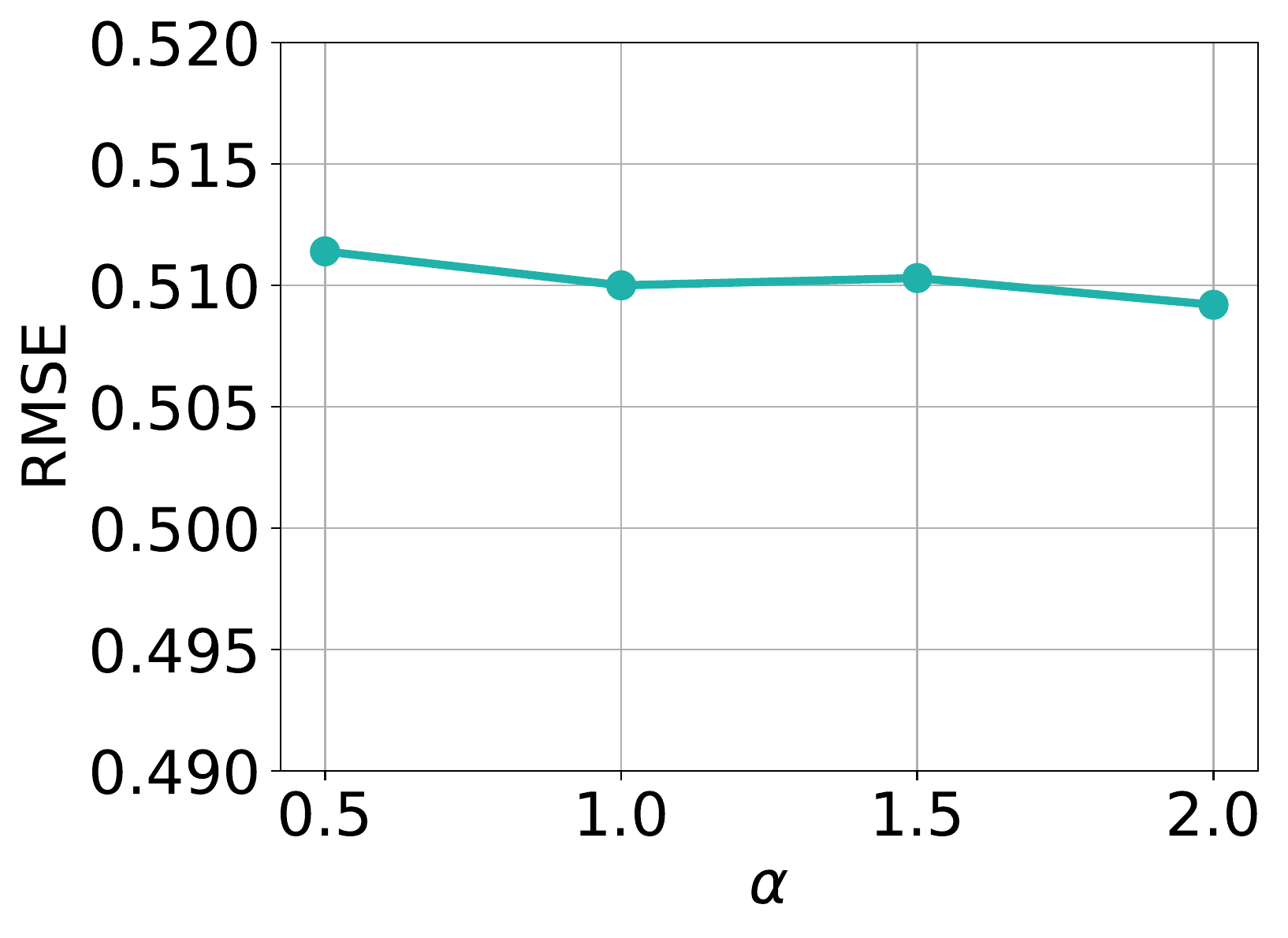}
    \caption{\label{fig:alpha_no2}: NO2}
\end{subfigure}
\begin{subfigure}[c]{0.19\textwidth}
		\centering
\includegraphics[width=\textwidth]{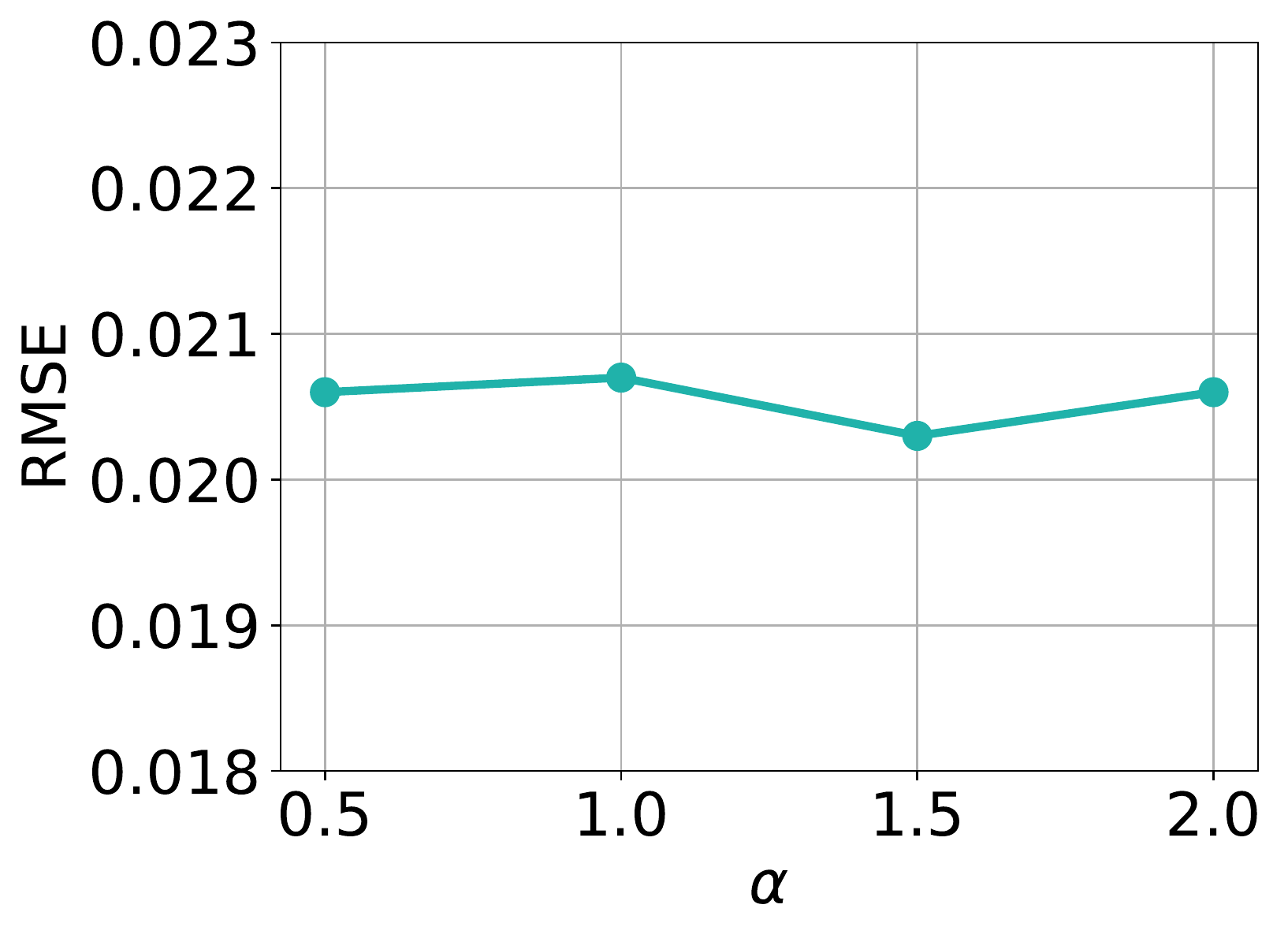}
    \caption{\label{fig:alpha_stock}: Exchange-Rate}
\end{subfigure}
\begin{subfigure}[c]{0.19\textwidth}
		\centering
\includegraphics[width=\textwidth]{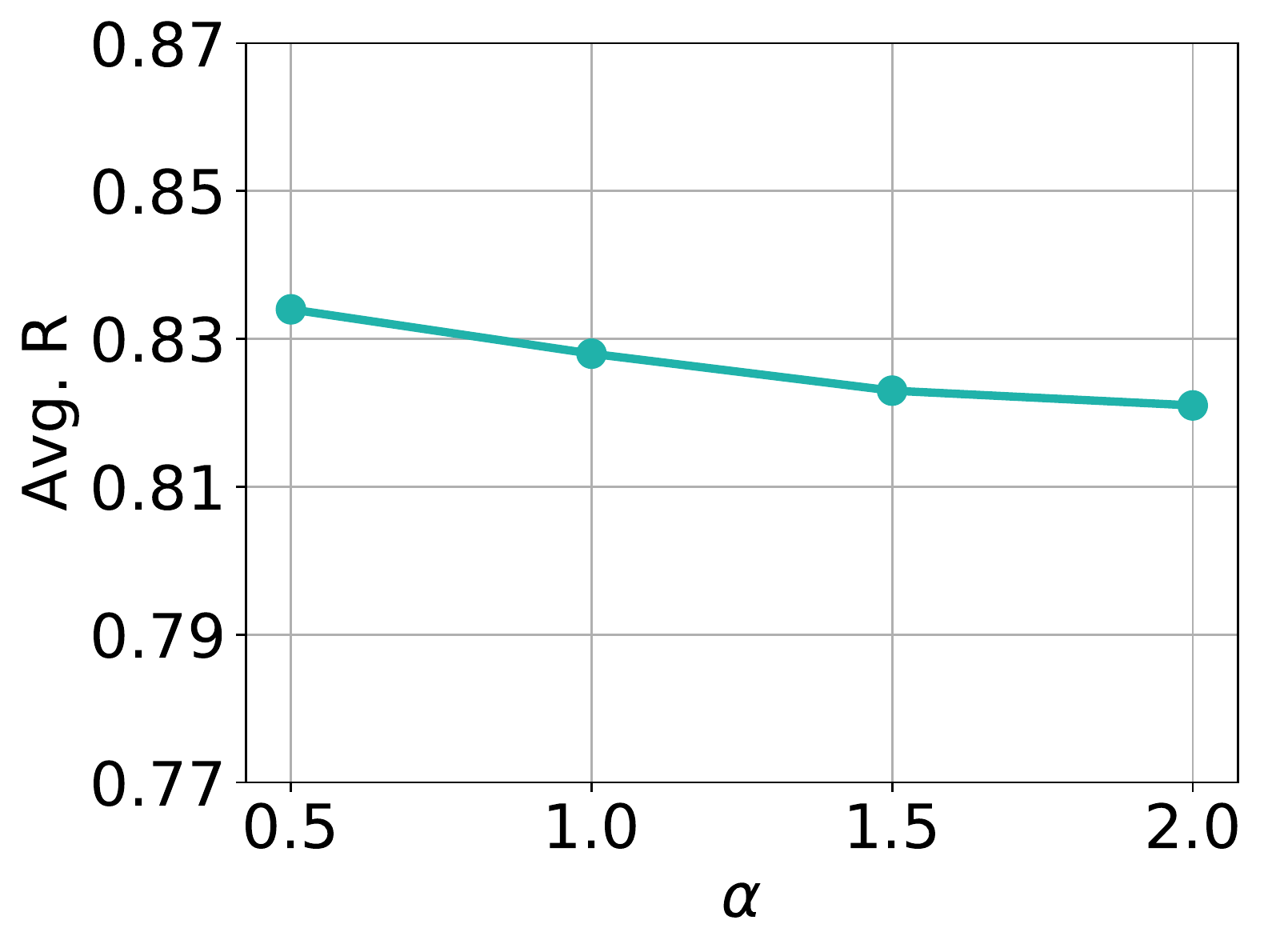}
    \caption{\label{fig:alpha_povertymap}: PovertyMap}
\end{subfigure}
\begin{subfigure}[c]{0.19\textwidth}
		\centering
\includegraphics[width=\textwidth]{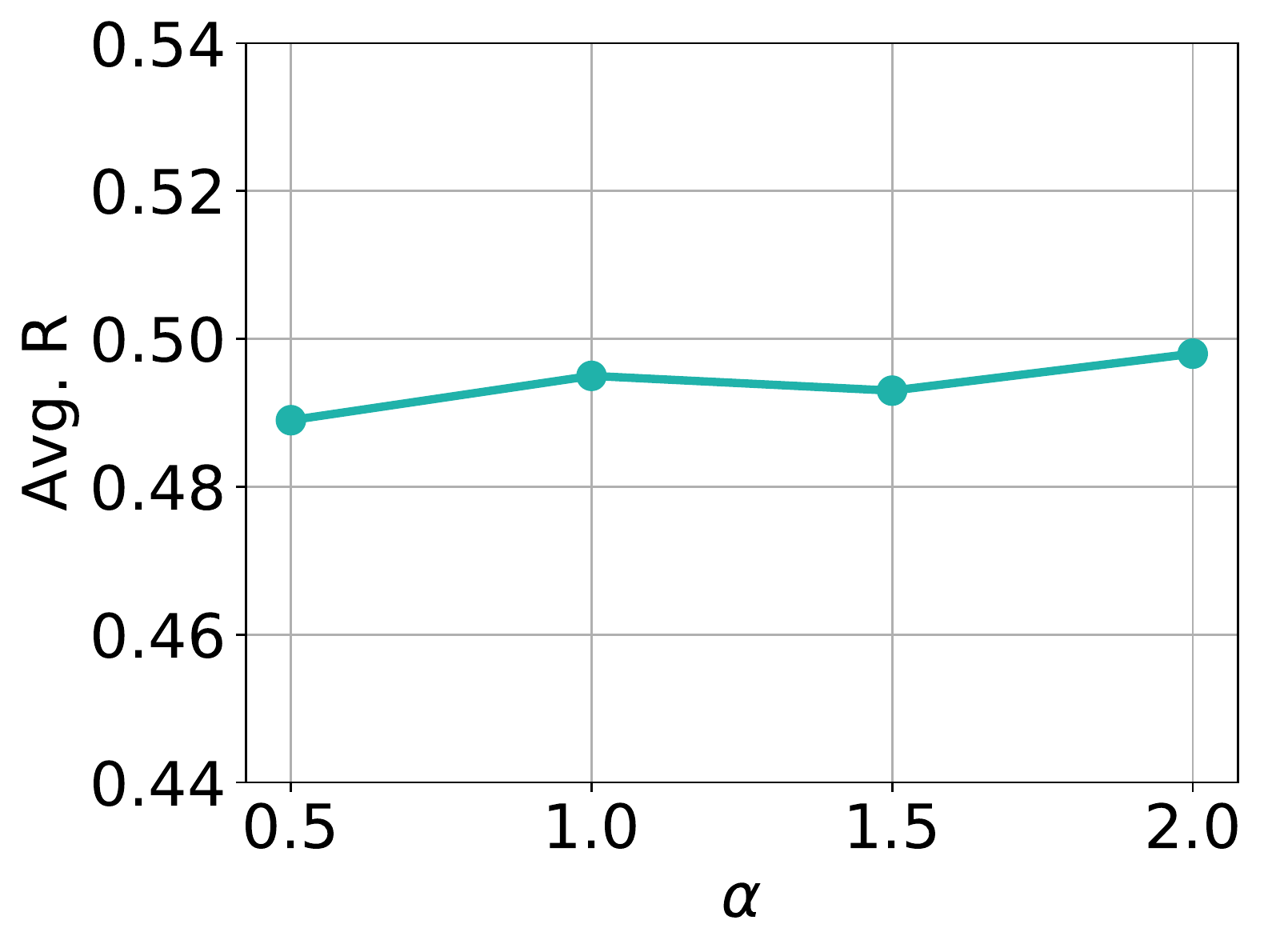}
    \caption{\label{fig:alpha_dti}: DTI}
\end{subfigure}
\caption{Robustness analysis of $\alpha$ in Beta distribution}
\label{fig:alpha}
\end{figure}

Finally, we analyze the effect of shape parameter $\alpha$ in the Beta distribution. The results of five datasets are illustrated in Figure~\ref{fig:alpha}, including Airfoil, NO2, Exchange-Rate, PovertyMap, and DTI. We observe that the performance is relatively stable with the change of $\alpha$, indicating the robustness of C-Mixup to the shape of Beta distribution.

\subsection{Robustness Analysis to Label Noise}
\label{sec:app_robustness}
We conduct experiments to investigate the robustness of C-Mixup to label noise. Specifically, we inject Gaussian noises into labels for all training examples. For each dataset, the noise is set as 30\% of the standard deviation of the corresponding original labels, where adding noise significantly degrades the performance compared to that with clean data. The results and the corresponding noise distributions on three datasets -- Exchange-Rate, ShapeNet1D, DTI are reported in Table~\ref{tab:app_noise_full}. According to Table~\ref{tab:app_noise_full}, C-Mixup still improves the performance over ERM and vanilla mixup, showing its robustness to label noise.

\begin{table}[h]
\small

\caption{Robustness analysis to label noise.}

\label{tab:app_noise_full}
\begin{center}
\setlength{\tabcolsep}{1.5mm}{
\begin{tabular}{l|c|c|c}
\toprule
\multirow{2}{*}{Model}  & Exchange-Rate & ShapeNet1D & DTI\\\cmidrule{2-4}
& RMSE $\downarrow$  & MSE $\downarrow$  & Avg. $R$ $\uparrow$
\\\midrule
Noise Type & $\mathcal{N}(0, 1.18\times10^{-3})$  & $\mathcal{N}(0, 0.874)$ & $\mathcal{N}(0, 7.59\times10^{-3})$
\\\midrule
ERM/MAML &  0.0381 $\pm$ 0.0014  & 5.553 $\pm$ 0.098  & 0.334 $\pm$ 0.018\\
mixup/MetaMix &  0.0375 $\pm$ 0.0017 & 5.329 $\pm$ 0.101 & 0.307 $\pm$ 0.021 \\

\textbf{\textbf{C-Mixup}}   & \textbf{0.0360} $\pm$ \textbf{0.0013} & \textbf{5.185 $\pm$ 0.096} & \textbf{0.356} $\pm$ \textbf{0.013} \\
\bottomrule
\end{tabular}}
\end{center}

\end{table}

\end{document}